# Why Autonomous Vehicles Are Not Ready Yet: A Multi-Disciplinary Review of Problems, Attempted Solutions, and Future Directions

Xingshuai Dong[1], Max Cappuccio[2], Hamad Al Jassmi[3,4], Fady Alnajjar[3,4], Essam Debie[5],
Milad Ghasrikhouzani[2], Alessandro Lanteri[6], Ali Luqman[3,4], Tate McGregor[7],
Oleksandra Molloy[8], Alice Plebe[9], Michael Regan[10] and Dongmo Zhang[11]

[1]School of Information and Communication Technology, Griffith University
[2]School of Engineering and Technology, University of New South Wales, Canberra
[3]Emirates Centre for Mobility Research Center, United Arab Emirates University
[4]College of Information Technology, United Arab Emirates University
[5]University of Canberra
[6]ESCP Business School
[7]School of Humanities and Social Sciences, University of New South Wales, Sydney
[8]School of Science, University of New South Wales, Canberra
[9]University College London
[10]School of Civil and Environmental Engineering, University of New South Wales, Sydney
[11]School of Computer, Data and Mathematical Sciences, Western Sydney University

*Abstract*— **Personal autonomous vehicles are cars, trucks and bikes capable of sensing their surrounding environment, planning their route, and driving with little or no involvement of human drivers. Despite the impressive technological achievements made by the industry in recent times and the hopeful announcements made by leading entrepreneurs, to date no personal vehicle is approved for road circulation in a 'fully' or 'semi' autonomous mode (autonomy levels 4 and 5) and it is still unclear when such vehicles will eventually be mature enough to receive this kind of approval. The present review adopts an integrative and multidisciplinary approach to investigate the major challenges faced by the automative sector, with the aim to identify the problems that still trouble and delay the commercialization of autonomous vehicles. The review examines the limitations and risks associated with current technologies and the most promising solutions devised by the researchers. This negative assessment methodology is not motivated by pessimism, but by the aspiration to raise critical awareness about the technology's state-of-the-art, the industry's quality standards, and the society's demands and expectations. While the survey primarily focuses on the applications of artificial intelligence for perception and navigation, it also aims to offer an enlarged picture that links the purely technological aspects with the relevant human-centric aspects, including cultural attitudes, conceptual assumptions, and normative (ethico-legal) frameworks. Examining the broader context serves to highlight problems that have a cross-disciplinary scope and identify solutions that may benefit from a holistic consideration.**

*Keywords* **Computer vision, Autonomous vehicles, Self-driving vehicles, Advanced Driver Assistance Systems, Automated Vehicles, Human Factors, Driverless Vehicles, Cultural Factors, Attitudes towards Autonomy, Technology Adoption, Trust, Acceptance, Ethical Dilemmas, National Policies.**

## 1. Introduction

### 1.1. Context and Goals of the Survey

In recent years, autonomous vehicles (also known as "intelligent", "self-driving", or "driverless" vehicles) and their underlying technologies underwent remarkable development [1]–[3], which was not, however, sufficient to make them road-ready. This review specifically focuses on *personal autonomous vehicles* (PAVs), i.e. autonomous cars, trucks, and bikes (as opposed to uncrewed cargos and collective passenger transports like autonomous coaches, trains, and trams) that rely on *artificial intelligence* (AI) to sense their surrounding environment, plan their route, and transport their passengers to destination with little or no supervision by human drivers. The goal is to highlight the limitations and the shortcomings that still delay the large-scale implementation and the commercialization of PAVs. This investigation is motivated by the growing tension between the expectations bestowed on PAVs and the difficulty to meet them.

To begin with, PAVs are expected to alleviate human driver's burden and assist people with disabilities or driving constraints (e.g., people legally prevented from driving) through performing some or all the intelligent operations required for driving, such as adaptive cruise control, lane keep assist, pre-collision avoidance, and traffic sign recognition. Thanks to the superior precision, accuracy, consistency, and velocity of their responses, PAVs are, in principle, likely to commit on average fewer errors than humans (especially the humans susceptible to fatigue, distraction, poor driving behaviors, substance abuse, adverse weather, etc.), thus providing increased safety [4]. That is why, once sufficiently mature, PAVs are expected to improve driving safety by reducing road accidents and human error injuries, which in turn could alleviate traffic congestion, improve viability, minimize fuel consumption and reduce air pollution [166]. While the desiderata are clear, assessing the current state of PAVs' development is not easy, as their readiness for commercialization and large-scale adoption does not depend on a single technological variable, but on a multiplicity of deeply intertwined, complex, sometimes intangible, factors, ranging from implementational to economical, socio-technical, human, and normative factors [480].

Large-scale programs for testing PAVs in urban environments and on motorways have been running in the US [167], China [168], the UAE [169], Europe [170], and Japan [171], among others. These programs aim to evaluate the safety, efficiency, and adaptability of PAVs under diverse traffic conditions and regulatory frameworks. In addition, robot-taxi services are currently being tested in cities like Beijing and Wuhan (China) [172], Dubai (UAE) [173], San Francisco (California) [174], and Tokyo (Japan) [175], where PAVs navigate complex urban layouts and interact with real traffic. These tests are critical for refining autonomous technology and preparing for widespread deployment, as they provide valuable data on vehicle performance, public safety, and regulatory compliance in a range of real-world settings. However, results are mixed and show diverse critical areas: on the one hand, self-driving cars seem able to navigate traffic and deal appropriately with standard driving scenarios while preserving a satisfactory level of safety [477]; on the other hand, several accidents have been caused by the unexpected failure of autonomous vehicles on American roads, which involved severe damage to property and casualties [483, 484, 485, 486]. In the cities where robot-taxis were tested, frequent incidents and malfunctions (often involving delays and traffic congestion) have been reported [478], [479].

Responses by users and bystanders and perceptions by prospective users and investors are also heterogenous [176], with reported levels of trust and acceptance fluctuating significantly [176]. Remarkably, some developers, enthused by the successes of the newest technological solutions, have indicated that PAVs equipped with 'full' or 'high' driving automation were almost ready for commercialization [177], at a time when some of the major car makers decided to temporally divest from 'full' or 'semi' autonomy (automation levels 5 and 4, respectively, see Section 2.1) mode to focus on simpler assistive technologies (automation levels 1-3) to manage risk and optimize the allocation of R&D resources.

Despite some impressive technological achievements and the optimistic announcements made by some entrepreneurs, to date only a very few models of private cars equipped with automation level 3 ('conditional driving automation', see Section 2.1) are legally allowed to circulate, and anyway with limitations (e.g., only in restricted areas, with speed limited to 95km/h) [481, 482], while no private vehicle is licensed to roam public roads in 'full' or 'semi' autonomous mode (automation levels 4 and 5), and it is still unclear when privately-owned PAV equipped with automation levels 3, 4 and 5 will be mature enough to operate universally and without restrictions [178]. Even the extensive experimentation with company-owned robotaxis conducted in San Francisco has offered mixed results so far, with licenses initially granted and then temporarily revoked by local transport authorities after attracting strong negative responses by the city's inhabitants: even though robot-taxis prevalently operate in fully autonomous mode, the problems they tend to generate (frequent traffic congestions, occasional accidents), their standing restrictions (speed limitations and functioning limited to specific urban areas, without an option for highways), and their persisting need for human supervision (onboard safety drivers or remote drivers intervening when required) cast doubts on their overall maturity. Experts debate whether the development of the most advanced forms of autonomous drive constitutes a long-term, complex challenge requiring profound and capillary transformations of the general transportation infrastructure. The unreadiness of PAVs technology is apparent but its causes are not entirely obvious and there is no consensus about the nature of the critical obstacles or which of them being prioritized.

## 1.2. Research Questions and Scope

The present review mainly tackles five research questions: (Q1) What is still needed to make PAVs real and accessible to anybody? (Q2) What are the most persistent obstacles faced by developers and makers? (Q3) Are these problems structural in nature (i.e., universal and permanent) or contingent upon specific technological approaches? (Q4) Are they entirely distinct problems or are they characterized by a communal root and similar patterns? (Q5) Should they be addressed using short-termed, local fixes or long-termed, global strategies? Our review adopts an integrative and multidisciplinary approach to investigate the major challenges faced by the automative sector, with the aim to identify the problems that still trouble and delay the commercialization of autonomous vehicles.

Our review is primarily meant to examine the limitations and risks of current technologies and the most promising solutions devised by the researchers: our negative methodological approach is not motivated by pessimism toward the ambitions of autonomous drive (in fact, we are eager to recognize that many problems have already been or are on the way to being permanently solved) but, on the contrary, by the desire to support research & development through a greater critical awareness of his historical successes and failures. Our review particularly focuses on the applications of AI for perception and navigation, as the main shortcomings have been imputed to core areas concerning environment scene understanding and decision-making. However, it also aims to offer an enlarged picture that links the purely technological aspects with the relevant human-centric, conceptual, and normative aspects. Examining the broader context serves to highlight problems that have a cross-disciplinary scope and identify solutions that may benefit from a holistic approach.

This survey is designed to connect ten different domains in which PAV's research still struggles. It is organized as follows: Section 2 introduces the main conceptual challenges in PAV's research and defines key notions adopted in this area (the subsections cover: 2.1 the problem of quantifying and assessing autonomy; 2.2 the challenges of advanced driver assistance systems); Section 3 explores the challenges concerning environment perception (3.2 challenges related to sensors; 3.3 challenges related to data sets and neural networks; 3.4 challenges related to environment perception); Section 4 deepens the problems concerning AI for path planning (4.2 path planning challenges; 4.3 sensor uncertainty; 4.4 balancing safety and efficiency; 4.5 real time localization and mapping challenges, 4.6 large scale map updates challenges); Section 5 addresses commonsense reasoning and its limitations (5.1 grounding problem; 5.2 semantic integration problem; 5.3 defining the research agenda; 5.4 corner cases; 5.5 safety, reliability and feasibility; 5.6 vulnerable road users; 5.7 narrow and general AI theoretical challenge); Section 6 discusses the uncertain role of road infrastructure (6.2 inconsistency and unclarity challenge; 6.3 unpredictable conditions challenge; 6.4 digital infrastructure fragility challenge; 6.5 road design challenge); Section 7 is about connected autonomous vehicles and the challenges associated with traffic management (7.2 adaptivity challenge; 7.3 intelligent intersection management challenge; 7.4 traffic control challenge; 7.5 automated negotiation challenge); Section 8 is about human factors and interfaces (8.2 the cognitive workload challenge; 8.3 the situational awareness challenge; 8.4 the human-machine interface challenge; 8.5 autonomy level 4 and 5 human factors challenges); Section 9 discusses how users' attitudes and perceptions can negatively impact the adoption of PAVs (9.2 the trust challenge; 9.3 scarce familiarity; 9.4 technology resistance; 9.5 cultural prejudices); Section 10 faces the ethical dilemmas (10.2 the trolley problem; 10.3 the value alignment problem; 10.4 the accountability problem); Section 11 reviews public policy and governance issues (11.2 Safety and licensing; 11.3 Liability and insurance; 11.4 Data privacy and cybersecurity).

## 1.3. General Methodological Notes

The number of literature reviews related to autonomous driving is breathtaking. Several surveys concerning autonomous vehicles have been published over the last ten years (for example, Table II summarizes the main surveys on computer vision). Most of these surveys focus only on one or two applications of a specific technology (e.g., computer vision or deep learning) for autonomous driving: for example, Arnold et al. [8] reviewed the application of 3D object detection in autonomous driving; in, Zhang et al. reviewed deep learning-based lanes marking detection methods [9]; several studies reviewed vehicle detection methods for autonomous driving [10]–[12]; and Ranft et al. investigated the role of machine vision in intelligent vehicles [13].

Also, the majority of the available surveys address the technical problems most frequently faced by the autonomous driving systems, while only some of them address the human-centric and societal problems. However, we have never encountered a survey that tries to identify the communal root of the merely technological, the conceptual/theoretical, and the socio-technical problems, combining the understanding of problems as different as computer vision, mapping, road infrastructure, connectivity, situational awareness, trust, and moral dilemmas to identify communal patterns, unobvious links, and analogous trajectories.

Our review adopts a cross-disciplinary approach to achieve this integrative goal. Also, it offers a map of the efforts made by the major automotive manufacturers and by several countries to promote the development of autonomous driving systems, with a view to distinguish their distinctive approaches. To identify the root causes of the problems that still afflict autonomous driving systems we will replicate the following schema through all the sections of the survey: we first identify the major challenges faced by autonomous driving systems in each area; then we review how these challenges have been addressed; finally, we attempt to recognize, when possible, the connections and similarities between different areas. In this work, instead of covering every state-of-the-art paper, we apply discipline-specific sets of selection criteria, with each thematic section adopting different criteria. Independently of this specificity, all sections prioritize the latest papers (from 2013 to 2024) and those published in prestigious journals (with impact factors greater than 3.5) or reputable international conferences.

## 2. Fundamental Notions and Concept-Level Challenges

### 2.1. Levels of Driving Automation: the Problem of Quantifying Autonomy

Conceptually differentiating between increasing levels of autonomy represents a theoretical and taxonomical challenge in itself. Establishing a universal method to measure and assess an agent's level of autonomy is challenging for at least three reasons: first, unlike *automation* (which is a merely objectual property), *autonomy* is an inherently relational property (its nature requires to specify both what makes an agent independent and from what). Secondly, the application of this notion in the context of AI does not necessarily coincide with the way it is typically applied in the context of human agency: unlike traditional automated technologies (which involve only mechanisms capable to blindly-repeat predefined routines in ad hoc synthetic environments), in order to be considered autonomous artificial agents must demonstrate *intent*, *capability*, and *awareness* somehow comparable to the analogous qualities of human agents. Each of these qualities is a complex notion in its own respect, which constitute a third challenge [179].

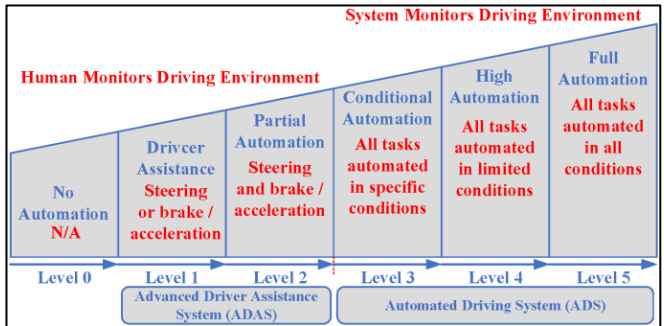

Fig. 1. SAE categories of autonomous driving system.

Despite these complexities, the subject matter experts conventionally refer to the standard taxonomy established by the Society of Automotive Engineers (SAE), which categorizes autonomous driving systems based on their amount of automation, distinguishing between six levels that range from "level 0" (no driving automation) to "level 5" (fully driving automation) [7]:

• Level 0 (*no driving automation*): warnings and momentary assistance. The vehicle has no driving automation technology, the human driver entirely operates the vehicle's movement, such as steering, accelerating, braking, etc.

• Level 1 (driver assistance): steering or brake/acceleration support. This is the lowest level of automation, where one aspect of the driving process is operated using data from sensors and cameras, while the driver retains entire control of the vehicle. The main challenge at this level is to inform the driving with one or more driving automation systems (see next section) that are transparent, simple to implement, useful to improve safety and driving performing, and not causing distraction or interfering with the normal driving routine. The information from sensors must be processed in a reliable and fast manner to provide useful input to control the vehicle.

• Level 2 (partial driving automation): steering and brake/acceleration support. At this level, *advanced driver assistance systems* (ADAS) undertake a significant portion of the driver's responsibilities. ADAS controls speed and steering simultaneously by relying on multiple data sources such as cameras, radar, LiDAR, and GPS, while drivers must keep their eyes on the driving environment. The driver is responsible for monitoring all operations and always must be ready to take over control of the vehicle. The system collects information on the driving environment and provides assistance such as acceleration, deceleration or steering to the driver. Compared to level 1, the main challenge at level 2 consists in augmenting the driving experience with more structured assistance that combines different sources and input and offering different services in a way that reduces the cognitive burden for the driver, instead of increasing. The information from different types of sensors must be integrated consistently and allow for accurate recognition of environmental features.

• Level 3 (conditional driving automation): the automated driving system performs the entire driving task without driver supervision, but only in limited conditions. Compared to level 2, the driver is no longer obliged to constantly monitor the driving environment but must be always present when an intervention request is made. The system undertakes most of the operations and monitors surrounding conditions with onboard sensors to make informed decisions in particular conditions. Drivers can take their hands off the steering wheel and eyes off the road but have to take control of the vehicle when an intervention request is made. As level 3 is between assisted and automated drive, its distinctive challenge is to implement a fluid transition of between human control and AI-based drive, ensuring both that the human is immediately available when needed and that control can be safely handed over to the AI.

• Level 4 (high driving automation): the vehicle can drive without driver supervision under all conditions and takeover requests can be ignored by the driver. The vehicle is capable of driving fully autonomously in proper settings and does not require any human interaction. The vehicle is competent at dealing with most problems on its own, so it asks the driver to take over only in some particular cases. At this stage, the distinctive challenge is to train several narrow AIs to be sufficiently developed to promptly initiate and reliably completely different goal-specific tasks (e.g., overtaking, parking, negotiating a roundabout, etc.).

Level 5 (full driving automation): like level 4, but the vehicle does not issue any takeover request. Vehicles equipped with this level of autonomy are driverless vehicles in a true sense. They are capable to reach the selected destination in any road conditions so that the intervention or even the attention of the driver are superfluous. At this stage, the distinctive challenge is to have a general AI system sufficiently developed to integrate all the tasks controlled at level 4 into a consistent plan of action while comprehending the overall context (e.g., distinguishing between waiting in line in the traffic vs being stuck in a procession) so to effectively adapt to complex, dynamic, and largely unpredictable circumstances. This level of autonomy are driverless vehicles in a true sense. They are capable to reach the selected destination in any road conditions so that the intervention or even the attention of the driver are superfluous. At this stage, the distinctive challenge is to have a general AI system sufficiently developed to integrate all the tasks controlled at level 4 into a consistent plan of action while comprehending the overall context (e.g., distinguishing between waiting in line in the traffic vs being stuck in a procession) so to effectively adapt to complex, dynamic, and largely unpredictable circumstances.

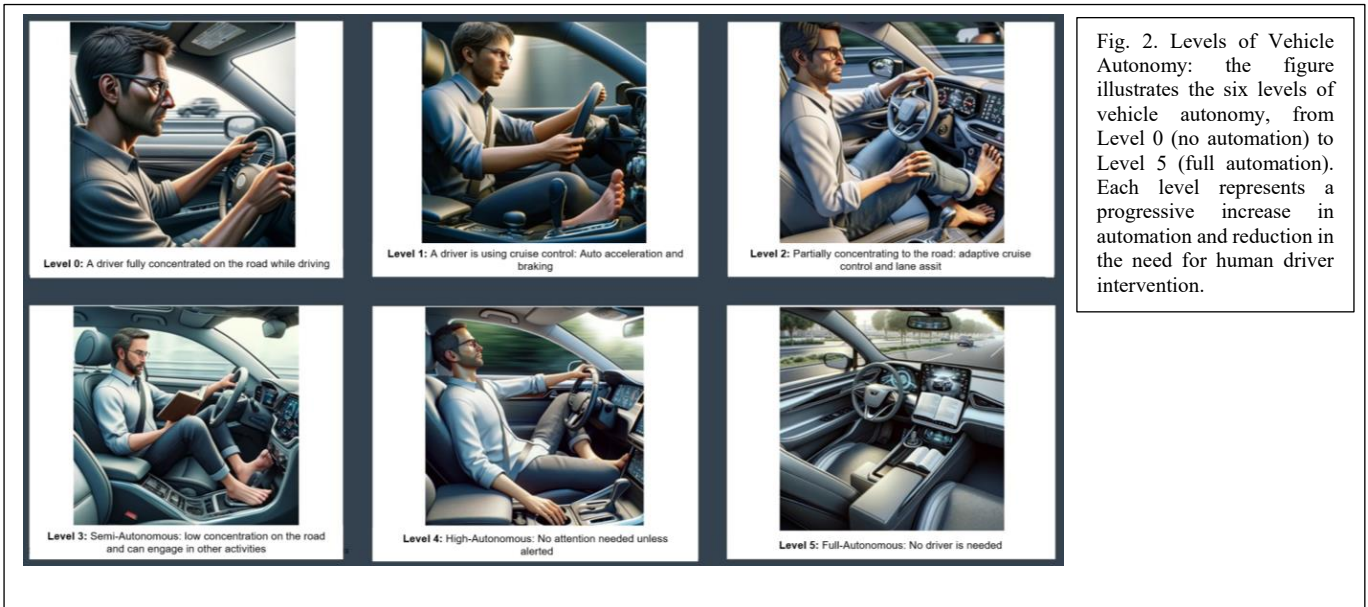

Fig. 2. Levels of Vehicle Autonomy: the figure illustrates the six levels of vehicle autonomy, from Level 0 (no automation) to Level 5 (full automation). Each level represents a progressive increase in automation and reduction in the need for human driver intervention.

This categorization allows us to conceptually distinguish between *automated* vehicles (SAE levels 1-3), which use electronic or mechanical devices to replace *some* human driving functions, and *autonomous* vehicles (SAE levels 4-5), which replace *all* human driving functions [355, p. 190]. Autonomous vehicles can drive themselves and do not strictly need to communicate or cooperate with other vehicles or infrastructure in order to complete their basic functions autonomously. To date, most of the existing driving systems advertised as "autonomous" are in fact equipped with level 2 or level 3 driving automation features that allow these systems to deal with some specific driving tasks in a minimally autonomous way. Both level 2 and level 3 systems depend on multiple sensors and computer vision algorithms to understand the driving environment. The levels of automation and the challenges related to autonomy will be further discussed in Section 8, concerning human factors and interfaces.

## 2.2. Development of Automated Driving Systems (ADSs) and Related Problems

In recent years, more and more vehicles equipped with technologies that assist human drivers or operate the vehicle under the supervision of human drivers are produced and delivered to the market. These "driving automation systems" are a set of technologies that provide drivers with assistance or warning in the process of driving. They are intended to enhance driving and road safety through efficacious and easy-to-adopt human-machine interfaces. Different levels of driving automation rely on different driving automation systems, which is why we distinguish between ADAS, that levels 1 and 2 of driving automation primarily rely on, and *automated driving systems* (ADSs), that are additionally involved in levels 3 to 5. Compared to ADAS, ADS may ultimately be able to operate a vehicle without the intervention of a human driver [5].

ADAS uses technical elements such as sensors, cameras, and computer vision algorithms to detect nearby obstacles or driver errors and respond accordingly. As a key underlying technology for autonomous vehicles, ADAS are designed to automate, adapt, and enhance vehicle technology for safety and better driving [5]. ADAS technologies are usually classified into two types: passive ADAS technologies alert the driver to a dangerous situation (so the driver must take actions to avoid an accident caused by this situation); instead, active ADAS technologies enable the vehicle to actively decide how to avoid the worst-case scenarios. For instance, if the pre-collision avoidance system detects an impending collision and the driver has failed to take evasive action, brakes can be applied automatically without the driver's interaction.

The origin of ADAS begun in 1948 when Ralph Teetor invented the modern cruise control system. In 1971, Daniel Wisner designed the electronic cruise control system that uses electric pulses to enable a vehicle to move at a constant speed. In 1984, Carnegie Mellon University (CMU) started the NavLab project that aims to use computer vision to achieve autonomous navigation [33]. The NavLab project developed the first modern autonomous vehicle that was featured with level 1 autonomy. In 1987, Mercedes-Benz developed the first level 2 autonomous vehicle that was able to simultaneously control steering and acceleration under the supervision of human driver [34].

In 1990, William Chundrlik and Pamela Labuhn invented the adaptive cruise control (ACC) system, which enables a vehicle to maintain a pre-set speed in the absence of a detected there is a preceding vehicle and maintains a preset following distance. Motivated by the advancement of the modern era and demand for new technology, more advanced system was invented. In 1995, the OnStar company introduced the collision avoidance system which utilizes a computer-operated system consisting of radar, laser, and/or vision technology to detect whether or not the vehicle has collision risk. In 2008, Volvo invented the Automatic Emergency Braking (AEB) system, and its XC60 was the first vehicle to be launched with AEB system. Two years later, Volvo introduced pedestrian detection with full auto brake, which applies radar and cameras to warn a driver if pedestrians appear in front of the vehicle and then brakes automatically if the driver fails to stop. This is a milestone in the automotive industry, acknowledging computer vision as central components of autonomous driving.

Table I: main characteristics of the autonomous driving systems developed by 18 automotive manufacturers. We compare these systems in terms of the types of sensors, the functions, and SAE levels of autonomy. Information sources: Audi [41], BMW [42], Fiat [43], Ford [44], Honda [45], Hyundai [46], Kia [47], Land Rover [48], Lexus [49], Mazda [50], Mercedes-Benz [51], [52], Mitsubishi [53], Nissan [54], [55], Subaru [40], [56], Tesla [57], TOYOTA [58], Volkswagen [59], and Volvo [37].

| Company | Country | System | Sensors | Functions | | | | | | | | Level |
|---------|---------|--------|---------|-----|-----|-----|-----|-----|-----|-----|-----|-------|
| | | | | ACC | DAM | HBA | LDW | LKA | PCW | TJA | TSR | |
| **Audi** | Germany | Pre-Sense | Camera, Radar, Ultrasonic Sensor | √ | – | √ | √ | √ | √ | √ | √ | Level 2 |
| **BMW** | Germany | Driving Assistant plus | Camera, Radar Ultrasonic Sensor | √ | – | – | √ | √ | √ | – | √ | Level 2 |
| **Fiat** | Italy | Ducato | Camera, Radar | √ | √ | √ | √ | √ | √ | √ | √ | Level 2 |
| **Ford** | USA | Co-Pilot360 | Camera, Radar | √ | √ | √ | – | √ | √ | – | √ | Level 2 |
| **Honda** | Japan | SENSING | Camera, Radar | √ | – | √ | √ | √ | √ | √ | √ | Level 2 |
| **Hyundai** | South Korea | Smart Sense | Camera, Radar | √ | √ | √ | – | √ | √ | – | √ | Level 2 |
| **Kia** | South Korea | Kia Drive Wise | Camera, Radar | √ | √ | √ | √ | √ | √ | – | – | Level 2 |
| **Land Rover** | UK | InControl | Camera | √ | √ | – | √ | √ | √ | – | √ | Level 2 |
| **Lexus** | Japan | Lexus Safety System+ | Camera, Radar | √ | – | √ | √ | √ | √ | √ | √ | Level 2 |
| **Mazda** | Japan | i-ACTIVSENSE | Camera, Radar, Ultrasonic Sensor | √ | √ | √ | √ | √ | √ | √ | √ | Level 2 |
| **Mercedes-Benz** | Germany | Drive Pilot* | Camera, LiDAR, Radar, Ultrasonic Sensor | √ | √ | √ | √ | √ | √ | √ | √ | Level 3 |
| **Mitsubishi** | Japan | MiTEC | Camera, Ultrasonic Sensor | √ | √ | √ | √ | √ | √ | √ | √ | Level 2 |
| **Nissan** | Japan | ProPILOT | Camera, Radar | √ | – | √ | √ | √ | √ | – | √ | Level 2 |
| **Subaru** | Japan | Preventative Safety** | Camera | √ | √ | √ | √ | √ | √ | √ | √ | Level 2 |
| **Tesla** | USA | Autopilot | Camera, Radar, Ultrasonic Sensor | √ | √ | √ | √ | √ | √ | √ | √ | Level 2 |
| **TOYOTA** | Japan | Toyota Safety Sense | Camera, Radar | √ | √ | √ | √ | √ | √ | – | √ | Level 2 |
| **Volkswagen** | Germany | IQ.DRIVE | Camera, Radar | √ | √ | – | √ | √ | √ | √ | √ | Level 2 |
| **Volvo** | Sweden | Ride Pilot* | Camera, LiDAR, Radar, Ultrasonic Sensor | √ | – | – | √ | √ | √ | – | – | Level 3 |

In 2014, Tesla became the first company to release commercial autonomous vehicles. These vehicles were equipped with Autopilot system [35], which has lane keep assistance, adaptive cruise control, and traffic sign recognition functions. The Autopilot system is classified as level 2, as it requires human drivers to be paying attention and ready to resume control at all times. Since October 2016, vehicles manufactured by Tesla were equipped with eight cameras, twelve ultrasonic sensors, and a radar for environment perception to enable autonomous driving.

### 2.2.1. ADAS and Their Challenges

Until now, the popular ADAS features that have been delivered to the market include:

**Adaptive Cruise Control (ACC)** – A feature that automatically adjusts the vehicle's speed to maintain a safe distance from vehicles ahead. However, ACC can struggle with poor weather conditions (e.g., rain, fog, snow) that impact sensor accuracy. Moreover, in heavy traffic, the system may react too slowly or fail to response to sudden changes, resulting in jerky accelerations or braking. Drivers might also become over-reliant on ACC, leading to reduced attentiveness, which is risky if manual intervention is required.

**High Beam Assist (HBA)** – A feature that automatically adjusts the headlamp range (switches between high beam and low beam) depending on the brightness of detected vehicles and certain road conditions. However, the system may occasionally misinterpret reflective surfaces, street signs, or distant vehicles, resulting in inappropriate headlight adjustments. This can cause temporary blindness for oncoming drivers or insufficient illumination in certain conditions, potentially compromising safety.

**Lane Departure Warning (LDW)** – A feature that uses one or more cameras to detect lane markings ahead of the vehicle and alerts the driver with visual, audible, and/or vibration warnings if the vehicle begins to drift out of its lane. The system may struggle with faded or poorly marked roads, making it less reliable in areas with poor road infrastructure. False warnings can also occur if the lane markings are unclear or during intentional lane changes without signaling, which can frustrate drivers and reduce trust in the system.

**Lane Keep Assist (LKA)** – A feature that uses one or more cameras to detect the lane markings ahead of the vehicle and monitor the vehicle's position within the lane. If the vehicle leaves its lane and the driver fails to take corrective action, the system can automatically provide steering adjustments to help keep the car securely within the detected lane. The system depends on clear lane markings, which may be absent or faded in certain areas. When lane markings are unclear, the system may not function correctly and fail to provide necessary steering corrections. In addition, excessive reliance on LKA could lead to reduced driver attentiveness.

**Pre-Collision Warning (PCW)** – A feature that uses one or more cameras or radar to detect a potential collision with a vehicle or pedestrian in front of the vehicle. If the system determines the driver has failed to take evasive action, the brakes can be applied automatically. The key issue of PCW is false positives, where the system detects a threat that does not exist, resulting in unnecessary braking that might confuse the driver. Furthermore, PCW may struggle with detecting smaller objects or fast-moving targets like motorcycles.

**Traffic Sign Recognition (TSR)** – A feature that recognizes and relays traffic sign information to drivers via the instrument panel. It may not work effectively in areas where traffic signs are obscured, damaged, or placed in non-standard positions. False recognition can cause the system to misreport speed limits or other critical signs, potentially leading to confusion or non-compliance with traffic rules.

**Driver Attention Monitor (DAM)** – A camera-based technology that tracks driver alertness. One major concern is the possibility of false alarms, where the system incorrectly interprets normal driver actions, like checking mirrors or adjusting controls, as signs of distraction. On the other hand, the system may fail to detect actual drowsiness if the driver's face is not clearly visible to the camera.

**Traffic Jam Assist (TJA)** – A feature that uses one or more camera(s) to monitor lane markings and the vehicle ahead. TJA can automatically brake and steer if the driver does not react in time. The system's main limitation is its performance in rapidly changing traffic conditions or complex environments with multiple lane changes. Poor weather conditions or unclear lane markings can also reduce its accuracy, leading to delayed or incorrect responses.

It can be observed that most of the autonomous driving systems are level 2, except for the *Drive Pilot* and the *Ride Pilot* developed by Mercedes-Benz and Volvo, respectively. The level 3 systems apply multiple sensors, including cameras, radar, LiDAR, and ultrasonic sensors, to acquire real-time data from the surrounding environment. Moreover, these systems rely on high-definition (HD) maps, which provide detailed information on road geometry, route profiles, traffic signs, and unexpected traffic events. A key feature of level 3 systems is the integration of high-accuracy LiDAR with HD maps. In these systems, LiDAR scans are continuously matched to the HD map, allowing for precise vehicle positioning. It is worth noting that the *Drive Pilot* system has been approved for use on specific freeway sections in Nevada, US, but only at speeds up to 40 mph (64.37 km/h). This system is expected to feature in Mercedes's high-end S-Class and EQS sedan vehicles, and it costs 5320 euro for the S-Class and 7448 euro for the EQS in Germany [36].

The Ride Pilot system includes five radar sensors, eight cameras, 16 ultrasonic sensors, and a LiDAR for comprehensive data collection [37]. However, this system faces several challenges in real-world applications. While the sensor array aims to provide robust environmental perception, each sensor type has limitations under varying conditions, such as poor visibility or adverse weather, which can impact detection accuracy. Besides, the integration of data from multiple sensors requires complex data fusion algorithms to handle calibration, synchronization, and the resolution of conflicting data. This complexity can impact the reliability of object detection and classification. Moreover, as the Ride Pilot system is still undergoing road tests in Sweden, regulatory approval and safety validation remain necessary steps, potentially delaying deployment.

### 2.2.2. Persisting Challenges

Despite these advancements, significant challenges remain, especially as systems progress toward level 4 and level 5 autonomy. Compared to level 3 systems, level 2 systems primarily depend on cameras, radar, and ultrasonic sensors rather than LiDAR and HD maps. In April 2023, Audi abandoned its plan to introduce level 3 autonomy in its A8 sedan, reverting to its *Pre Sense* system, which supports level 2 autonomy [38]. The *Pre Sense* system includes additional features, such as a night vision assistant that uses a long-range infrared camera to detect heat-emitting objects, providing visual information to the driver in low-light conditions. In addition to the features listed in Figure 1, the *Pre Sense* system has the night vision assistant function which uses a long-range infrared camera to sense the thermal energy emitted by objects. The *EyeSight* system, developed by Subaru, relies solely on stereo RGB cameras (i.e., capturing light in Red, Green, and Blue wavelengths) to detect pedestrians, cyclists, and other vehicles, assessing their distance, shape, and speed to help avoid potential collisions [40]. In addition, *EyeSight* can detect the sudden activation of brake lights in the vehicle ahead, helping to avoid potential collisions. However, both systems exemplify the limitations of level 2 autonomy, as they depend on human oversight and cannot operate independently in all driving conditions.

To sum up, most of the current autonomous driving systems provide assistive functions to support human drivers to drive safely. Features such as ACC, LDW, LKA, and PCW have been solved by most of the current ADAS systems. Vehicles are equipped with these features that can control their steering, accelerating, and braking under the monitor of human drivers. These systems materially depend on sensors such as cameras, radar, ultrasonic sensors, and LiDAR to collect data from the surrounding environments. In addition, the listed features heavily depend on computer vision applications such as *depth estimation*, *object detection*, *lane detection*, and *traffic sign recognition* algorithms to extract information from the collected data. The extracted relevant information is then used by the vehicle's computer to make driving decisions. Thus, the main challenge to ensure the correct functioning of ADAS is that the underlying sensors and computer vision applications work accurately (precise and consistent), reliably (robust and flexible), efficiently (processing information parsimoniously and cost-effective), and easily integrated (interoperable, replaceable) [18] [5].

The level 3 system, *Drive Pilot*, represents a milestone in the development of autonomous driving. However, it can only operate at speeds up to 40 mph (64.37 km/h) on suitable freeway sections. Besides, the *Drive Pilot* system uses computer vision applications to sense the environment around the vehicle, and a HD map to estimate the position of the vehicle. Therefore, multiple information fusion is indispensable for achieving the aim of autonomous driving. The high-level computer vision tasks such as vision-based path planning, and visual localization and mapping that enable vehicles to autonomously plan their trajectories or localize their positions have been widely explored in academic community [180].

Information fusion plays a critical role in autonomous driving. It integrates data from various sensors, such as cameras, radar, LiDAR, and HD maps, to achieve a comprehensive understanding of the vehicle's surroundings [181]. This allows the system to make more accurate decisions in terms of lane detection, obstacle detection, traffic sign recognition, and position localization. However, the complexities of processing multi-modal data in real time results in significant computational burdens. Moreover, misalignments between data sources, sensor noise, and latency issues can degrade the accuracy and safety of autonomous driving systems [182].

Effective information fusion must address these challenges by ensuring that the data from different sensors is synchronized and correctly interpreted. Achieving this milestone would enable higher-level tasks, such as vision-based path planning and visual localization, allowing vehicles to autonomously plan their trajectories and localize their positions. Although these tasks have been widely explored in the academic community, practical implementation still faces significant obstacles due to the challenges arising from information fusion. Therefore, the integration of information fusion into real-world autonomous systems remains an area that requires further development.

## 2.3 Conclusions and the Way Forward

The development of autonomous driving has encountered notable challenges, particularly in progressing beyond levels 2 and 3 systems. Although levels 4 and 5 enable fully self-driving vehicles, they demand exceptionally high standards of reliability, real-time processing, and decision-making across diverse and unpredictable driving scenarios, such as densely populated urban environments and complex road networks [178]. The remaining of this review will document how current sensor technologies (Section 3), AI (Section 4), and information fusion techniques (Sections 3-5) have not yet matured to the point where they can ensure safety under these conditions. Not only because of these technical limitations but also because of regulatory barriers (Section 11) and users' perceptions about safety (Section 9) and ethical worries (Section 10), manufacturers have shifted their focus toward enhancing levels 2 and 3 systems. These systems, while still requiring human intervention, can autonomously manage specific driving tasks such as steering, acceleration, and braking under certain conditions. This approach is more feasible with the current state of technology and, as we will see in Section 11, is better aligned with existing regulatory approval processes. Also, as discussed in Section 8, gaining consumer trust is easier with these systems since they maintain human involvement, improving both safety and user experience.

In contrast, levels 4 and 5 autonomy require vehicles to operate entirely without human intervention under all conditions. This requires integrating data from multiple sensors, such as cameras, LiDAR, radar, and HD maps, to enable precise and reliable real-time decision-making. However, challenges in information fusion, including synchronization issues, sensor noise, and latency, have hindered progress. Achieving full autonomy remains the ultimate goal, but progress is slow, and a clear timeline for widespread deployment has not yet been established. While research efforts to develop technologies for levels 4 and 5 have are discontinuing, manufacturers are currently prioritizing the improvement of level 2 and 3 systems. These systems offer a more immediate path to enhancing safety and the driving experience, while the complex requirements for full autonomy continue to present significant challenges for the future. Considering the delays in the development of level 4 and level 5 system, and that their functions remain largely speculative, in the following sections we will review the commonly used sensors, data sets, and environment perception tasks for current autonomous driving systems.

## 3. Environment Perception and Sensors

### 3.1. Introduction and Methodology

PAVs rely on a range of sensors and computer vision algorithms to interpret and respond to their environments. This section provides an overview of widely used sensors [60], highlighting their operating mechanisms, sensing modalities, data sizes, and challenges associated with each. We then present key data sets [15] developed for AV perception tasks and investigate their unique attributes, helping to inform suitable selection for various applications.

For papers on computer vision and environment perception we chose IEEE Xplore as the primary repository, given its influential role in the fields of computer science, electrical engineering, electronics, and related areas [21]. To focus on the applications of computer vision in autonomous vehicles, we selected "computer vision," "autonomous vehicle," "autonomous driving," and "ADAS" as the foundational keywords. These were combined with terms related to computer vision applications, such as "pedestrian detection", "cyclist detection", "vehicle detection", "lane detection", "traffic sign recognition", "sensor", and "dataset" for advanced searches through Google Scholar. Additionally, we reviewed several preprint papers [6], [22]–[32] widely recognized in the field for introducing state-of-the-art research or datasets.

TABLE II: A summarization of a number of surveys related to autonomous vehicle perception published from 2013 to 2023. The surveys covered in this table are selected based on the relevancy to the main topic of this work, publication year, and reputation of the publisher. "AD": Autonomous Driving, "ADS": Autonomous Driving System, "AV": Autonomous Vehicle, "CAS": Collision Avoidance System, "PD": Pedestrian Detection, and "SOTA": State-of-the-art.

| Title | Year | Description | Remarks |
|---|---|---|---|
| Looking at vehicles on the road: A Survey of Vision-Based Vehicle Detection, Tracking, and Behavior Analysis [10] | 2013 | Investigating vision-based methods for vehicle detection, tracking, and behavior understanding | Surveyed vision-based methods for vehicle detection, tracking, and behavior understanding. Only traditional methods are covered. |
| Recent Progress in Road and Lane Detection: A Survey [14] | 2014 | Survey on approaches and algorithms for road and lane detection | Analyzed the road and lane detection methods from the perspective of different function modules. Only traditional methods are covered. |
| Vehicle Detection Techniques for Collision Avoidance Systems: A Review [11] | 2015 | Survey on vision-based vehicle detection and tracking algorithms for CAS | Analyzed vehicle detection methods for CAS. Compared the performance of different sensors. Discussed motorcycle detection and tracking methods. |
| The Role of Machine Vision for Intelligent Vehicles [13] | 2016 | Reviewing machine vision for driver assistance and automated driving | Outlined the present and the potential future role of machine vision for driver assistance and AD. |
| When to Use What Data Set for Your Self-driving Car Algorithm: An Overview of Publicly Available Driving Datasets [15] | 2017 | Analyzing 27 publicly available data sets for AD | Compared 27 data sets from different perspectives. Provided guidelines for selecting data set for different tasks. |
| Autonomous Vehicle Perception: The Technology of Today and Tomorrow [16] | 2018 | Reviewing the AV perception methods | Presented an overview of the sensor, localization and mapping techniques for AVs. Discussed improvements for sensors and AV perception. |
| A Survey on 3D Object Detection Methods for Autonomous Driving Applications [8] | 2019 | Survey 3D object detection methods for AD applications | Reviewed 3D object detection in AVs. Analyzed the pros and cons of sensors. Discussed standard data sets. |
| Pedestrian Detection in Automotive Safety: Understanding State-of-the-Art [17] | 2019 | Survey pedestrian detection methods in the automotive application | Investigated the techniques used in PD for automotive application. Highlighted the demand for low-cost and robust PD solutions. |
| A Survey of Deep Learning Techniques for Autonomous Driving [18] | 2020 | Survey the current SOTA deep learning technologies used in AD | Investigated different AI and deep learning technologies used in AD. Tackled challenges in designing AI architectures for AD |
| LiDAR for Autonomous Driving: The Principles, Challenges, and Trends for Automotive LiDAR and Perception Systems [3] | 2020 | Reviewing LiDAR technologies and perception algorithms for AD | Introduced the principle of how LiDAR works. Analyzed the main development directions of LiDAR technology. |
| A Progressive Review: Emerging Technologies for ADAS Driven Solutions [5] | 2021 | Reviewing different functionalities of ADAS and its levels of autonomy | Progressively reviewed the principle of different sensors, and important ADAS features. Examined various multi-sensor systems used in ADAS. |
| Deep Learning in Lane Marking Detection: A Survey [9] | 2021 | Survey the deep learning-based methods for lane marking detection | Focused on deep learning-based lane marking detection. Provided in-depth analysis lane marking detection algorithms. |
| Deep Neural Network Based Vehicle and Pedestrian Detection for Autonomous Driving: A Survey [19] | 2021 | Survey the DNN-based methods for pedestrian and vehicle detection | Performed experimental comparison of several mainstream detectors for pedestrian and vehicle. |
| Detection of Motorcycles in Urban Traffic Using Video Analysis: A Review [20] | 2021 | Reviewing algorithms for motorcycle detection and tracking | Investigated the algorithms for motorcycle detection and tracking from videos. Motorcycle detection in urban environments. |
| A Review of Vehicle Detection Techniques for Intelligent Vehicles [12] | 2022 | Reviewing the vehicle detection methods for intelligent vehicles | Investigated vehicle detection with different sensors. Compared the performance of classical methods and deep learning-based methods. |
| Camera-Radar Perception for Autonomous Vehicles and ADAS: Concepts, Data sets and Metrics [6] | 2023 | Survey the camera and radar-based perception methods for ADAS and AVs | Analyzed the pros and cons of different sensing modalities. Presented an overview of the deep learning-based detection and segmentation methods. |

### 3.2. Sensors and Related Challenges

#### 3.2.1. Cameras

Cameras are the most commonly used image sensors that sense the visible light spectrum reflected from objects [60]. Compared with Radar and LiDAR, cameras are relatively cheap. Images from the camera give straightforward 2D information, which can be applied to object detection or lanes detection. The measure distance of cameras ranging from several centimeters to 100m. However, the performance of cameras is greatly reduced by light or weather conditions such as fog, haze, smock, and smog, which limits their applications to daytime and clear skies. Moreover, cameras also suffer from huge data management problems, due to the large volume of data they generate. A single high-resolution camera usually generates 20-60 MB data per second [61]. It poses great challenge to data processing systems as analyzing and interpreting high volumes of data in real-time requires significant computational resources.

#### 3.2.2. LiDAR

LiDAR is an active ranging sensor that measures the distance to objects based on the round-trip time of a laser light pulse [60]. Laser beams are low divergence to reduce power decay with distance, thus, it enables LiDAR to measure distance up to 200m. Benefit from the high accuracy distance measure ability, LiDAR is commonly applied to construct accurate and high-resolution maps. However, the LiDAR suffers from sparse measurements which is not suitable for detecting small targets. Furthermore, its measurement range and measurement accuracy could be influenced by weather conditions [62]. Finally, the high costs restrict the wide deployment of LiDAR on autonomous vehicles [63]. Specifically, the 16 lines Velodyne LiDAR costs almost $8000, while the Velodyne VLS-128E costs over $100000. LiDAR produces about 10-70 MB data per second, which is a challenge because today's onboard computing platforms are not always powerful enough to achieve a real-time processing speed [61].

Table III: this table summarizes some data sets for the perception tasks of autonomous vehicles that collected in the period from 2013 to 2023. The table presents types of sensors, the presence of adverse conditions (e.g., time, weather), the data set size, and the position of data collection. It also shows the types of the intended applications and annotation format. Therefore, Table III provides guidelines for readers to select the appropriate data set for the related applications. "Sensors": only visual sensors are illustrated in the Table, "K": thousand, "M": million, "USYD": The University of Sydney, "–": represents that no information is provided, and "◇": More than 1.5 years once a week continuously updated. "IS": Instance Segmentation, "LD": Lane Detection, "OD": Object Detection, "PD": Pedestrian Detection, "SS": Semantic Segmentation, "TLD": Traffic Light Detection, "TSD": Traffic Sign Detection, "VD": Vehicle Detection, "VP": Visual Perception.

| Year | Data set | Application | Sensors | Time | Weather | Image Frames | Annotation Type | Locations |
|---|---|---|---|---|---|---|---|---|
| 2013 | KITTI [64] | VP | RGB Camera LiDAR | Day | Real | 44K | 2D Boxes, 3D Boxes Road Surface, Pixel | Karlsruhe |
| 2016 | LISA TL [65] | TLR | RGB Camera | Day, Night | Real | 43016 | 2D Boxes | San Diego |
| 2016 | TT100K [66] | TSD | Panorama Camera | Diverse | Diverse | 100K | 2D Boxes, Pixel Mask | China |
| 2017 | BOSCH [67] | TLD | RGB Camera | – | Diverse | 13427 | 2D Boxes | San Francisco |
| 2018 | BDD100K [25] | VP | RGB Camera | Diverse | Diverse | 100K | 2D Boxes, Lane Markings, Drivable Area, Pixel | New York San Francisco |
| 2018 | KAIST [68] | VP | RGB Camera Thermal Camera LiDAR | Diverse | – | 95000 | 2D Boxes | Seoul |
| 2019 | $D^2$-City [26] | 2D OD, OT | RGB Camera | – | – | 1000 clips | 2D Boxes | 5 China cities |
| 2019 | NightOwls [69] | PD | RGB Camera | Down, Night | Diverse | 279K | 2D Boxes | Europe |
| 2019 | STL [70] | TLD | RGB Camera | Diverse | Diverse | 14800 | 2D Boxes, Pixel | San Francisco |
| 2020 | A2D2 [28] | VP | RGB Camera, LiDAR | Day | – | 41277 | 3D Boxes Pixel | 3 Germany cities |
| 2020 | A*3D [71] | 3D OD | RGB Camera LiDAR | Diverse | Diverse | 39k | 3D Boxes | Singapore |
| 2020 | MTSD [72] | TSD | RGB Camera | Diverse | Diverse | 105K | 2D Boxes | Global |
| 2020 | USyd [73] | VP | RGB Camera, LiDAR | Diverse | Diverse | ◇ | Pixel | USYD |
| 2021 | PVDN [74] | PVD | Gray Camera | Night | – | 59746 | Keypoints | – |
| 2022 | OpenMPD [75] | 2D/3D OB 2D/3D SS | RGB Camera LiDAR | Day | Sunny | 15000 | 2D Boxes, Pixel | Beijing |
| 2022 | CeyRo [76] | TSD, TLD | RGB Camera | Diverse | Diverse | 7984 | 2D Boxes | Sri Lanka |
| 2022 | DualCam [31] | TLD | RGB Cameras | – | – | 1845 | 2D Boxes | – |
| 2022 | KITTI-360 [77] | VP | RGB Camera LiDAR | – | – | 150K | 3D Boxes, Pixel | Karlsruhe |
| 2022 | K-Lane [78] | LD | RGB Camera LiDAR | Day, Night | – | 15382 | Lane lines | – |
| 2023 | S2TLD [79] | TLD | RGB Camera | Diverse | Diverse | 5786 | 2D Boxes | China |
| 2023 | ZOD [32] | 2D/3D OD IS, SS TSR, RC | RGB Camera LiDAR | Day, Night Twilight | Diverse | 100K | 2D/3D Boxes, Classification, Pixel | Europe |

Beyond traditional LiDAR applications, AI-driven algorithms have expanded LiDAR's potential by improving real-time data processing and navigation capabilities. For example, Zhang et al. [148] developed an AI-driven approach to process LiDAR data in real time, creating accurate 3D reconstructions of the environment. This combination of AI and LiDAR data enables PAVs to navigate complex terrains with improved precision, facilitating obstacle identification and path planning. By merging LiDAR data with AI techniques, we can overcome some inherent limitations of LiDAR, enhancing the AV's perception and decision-making processes.

### 3.2.3. Radar

Radar uses electromagnetic or radio waves to detect objects [60]. It can not only measure the distance to an object but also detect the angle and relative speed of the moving object. Generally, the working frequency of radar system is 24 or 77 GHz. The maximum measure distance of 24 GHz radar is 70m, while the maximum measure distance increases to 200m for the 77 GHz radar. Compared with LiDAR, radar is well suited for measurements in conditions with dust, smoke, rain, adverse light or rough surfaces [60]. In terms of the data size, each radar produces 10-100 KB per second [61].

### 3.2.4. Ultrasonic Sensors

Ultrasonic sensors measure the distance of objects via emitting ultrasonic waves [2]. The sensor head emits an ultrasonic wave and receives the wave reflected from the target. Therefore, the distance can be calculated by measuring the time between the emission and reception. Ultrasonic sensors are widely used in automobile self-parking and anti-collision safety systems. Ultrasonic sensors have the merit of being easy to use, highly accurate, and can detect very small changes in position. However, it has limited measure distance (less than 20m), and inflexible scanning methods. The price of an ultrasonic sensor is usually less than $100. The ultrasonic sensor has a similar data size as radar, which is 10-100 KB per second [61].

### 3.3. Data Sets

A crucial component for the safety of autonomous driving is the perception of the environment around the autonomous vehicles. In general, autonomous vehicles are equipped with multiple sensors along with sophisticated computer vision algorithms to capture necessary information from the driving environment. However, these algorithms usually depend on deep learning techniques, especially convolutional neural networks (CNNs), which drives the requirement for benchmark data sets. Some data sets are better suited for specific tasks than others, which is why several data sets for evaluating different components of autonomous driving systems have been collected by researchers from both academia and industry. Table II summarizes various data sets used for perception tasks in autonomous driving, collected between 2013 and 2023. In this table, we analyze these data sets based on sensor types, the presence of adverse conditions (e.g., time of day, weather), data set size, and data collection locations. Furthermore, we summarize the intended applications and annotation formats. Therefore, Table II is expected to serve as a guideline for readers in selecting an appropriate data set for their specific applications.

### 3.4. Environment Perception and Related Challenge

Perception refers to the ability of an autonomous vehicle to gather data through onboard sensors, extract necessary information and gain the understanding of the environment around the vehicle [2]. It is a fundamental component of PAVs because it provides them with necessary information on the driving environment for safe driving, thus unsafe behavior of existing PAVs has often been imputed to incomplete or undeveloped perceptual capabilities [183]. A PAV requires the capability to correctly interpret the driving environment, recognizing elements such as obstacles, traffic signs, and the free drivable areas in front of the vehicle. In general, environmental perception tasks are associated with computer vision, deep learning, and CNNs [183]. To operate safely, PAVs are expected to successfully complete four key computer vision tasks: depth estimation, object detection, lane detection, and traffic sign recognition. In this section, we provide an overview of these tasks.

### 3.4.1. The Depth Estimation Challenge

The objective of depth estimation is to estimate a dense depth map from the input RGB image(s) [80]. Active methods use sensors such as RGB-D cameras, LiDAR, or radar to measure the depth information from the environment. However, RGB-D cameras suffer from a limited measurement range which is not suitable for autonomous vehicles that run at high speed in outdoor environments. LiDAR and radar are limited to sparse coverage. Besides, the price of high accurate LiDAR is extremely expensive which increases the cost of autonomous vehicles. Compared with LiDAR and radar, RGB cameras are cheaper, and they can provide richer information about the environment. Therefore, passive depth estimation methods based on cameras have been developed by both academia and industries.

The most common passive methods for depth estimation are based on stereo vision or monocular vision. Stereo depth estimation aims to find the correspondence between two rectified images from two cameras to predict the disparity between these two images [81]. The foundation of stereo depth estimation is similar to the depth perception of human eye and is on the basis of triangulation of rays from two overlapping viewpoints. In recent years, many stereo depth estimation methods [82]–[85] have been developed. The produced depth maps contain distance information from the surface of objects to the camera, which is of great importance for the PCW system in ADAS. For example, Subaru's *EyeSight* driver assist system utilizes stereo RGB cameras to determine the distance between the vehicle and pedestrians, cyclists and vehicles.

The stereo depth estimation algorithms require that both images have been rectified. The transformation process necessary to rectify the image is contingent upon the calibration process. However, the calibration process requires taking several images of a known calibration pattern (e.g., the checkerboard method), which makes the calibration relatively tedious. Therefore, stereo depth estimation methods are sensitive to various environmental conditions (e.g., mechanical shock) that can potentially change the physical structure of the camera [81].

Due to the recent advances in computer vision and deep learning, estimating depth maps from monocular RGB images is becoming more convenient. As a class of deep learning algorithm, CNNs use convolutional operation to replace matrix multiplication to process data with the format of multiple arrays, such as a RGB image consisting of three 2D arrays including pixel intensities in three color channels [86]. Therefore, they are specifically used for image recognition and tasks that involve the processing of pixel data. In 2014, Eigen et al. [87] developed the first CNN-based monocular depth estimation method and demonstrated the prospect of using CNN to predict depth maps from monocular RGB images. Then, inspired by [87], many monocular depth estimation networks [23], [88] have been introduced. However, these methods depend on extremely deep and complex network architectures that require high performance GPUs to run in real-time. To improve the running speed of monocular depth estimation, real-time CNNs [29], [89]–[91] have been developed. Compared to stereo depth estimation, monocular depth estimation does not require extrinsic calibration but usually achieves inferior depth accuracy.

### 3.4.2. Object Detection Challenges

#### 3.4.2.1. Generic Object Detection Challenge

Generic Object Detection aims to search for the instances of objects from a set of predefined classes (e.g., cat, dog, basketball, fridge) from input images. If present, the detector returns the spatial location and extent of each instance [92]. It places emphasis on detecting a broad range of classes of objects. The detectors are divided into two groups: two-stage detectors and one-stage detectors. The two-stage detectors begin by extracting a set of region proposals and then classify each of them via a separate network, while the single-stage detectors directly predict class probabilities and bounding box offsets from the input image in a unified network. The representative two-stage detectors are R-CNN [93] and its successors [94], [95].

R-CNN [93] first applies selective search algorithm [96] to extract a set of region proposals from the input image. The extracted region proposals are then resized to a fixed size and passed through a CNN to extract feature maps. Finally, the class-specified linear SVM classifiers are applied to predict the presence of an object within each region and to recognize object classes. One year later, He et al. [97] developed the spatial pyramid pooling network (SPPNet). The core contribution of the SPPNet is a spatial pyramid pooling (SPP) layer that allows CNNs to produce a fixed- length feature representation from the entire image. Based on the R-CNN and SPPNet, Girshick proposed Faster R-CNN [94]. Instead of separately learning a detector and a bounding box regressor as in R-CNN or SPPNet, Fast R-CNN jointly to learn classify object proposals and regress their spatial locations. Meanwhile, Ren et al. [95] designed a Region Proposal Network (RPN) for generating region proposals. RPN shares the fully convolutional layers with the detection network, therefore it works with little or no additional computations.

In 2016, Joseph et al. [98] treated object detection as a regression problem and designed the first CNN-based one-stage object detector, named YOLO. Unlike two-stage detectors, YOLO divides the input image into regions and simultaneously predicts bounding box and probability for each region. Liu et al. [99] introduced the SSD algorithm, which achieves better performance than YOLO in terms of running speed and accuracy. Benefitting from the multi-reference and multi-resolution detection techniques, SSD achieves competitive accuracy with two-stage detectors such as Faster R-CNN. The subsequent versions [24], [27], [100], [101] of YOLO that were developed after SSD outperform most of existing object detection algorithms in inference speed and accuracy through applying optimized structures.

Recently, YOLO-v8 [183] has been introduced with significant advancements over previous versions, including an improved model architecture and training techniques. It incorporates an enhanced backbone and neck structure with improved feature pyramid network (FPN) and path aggregation network (PANet) components, allowing for better handling of multiscale features. Besides, YOLO-v8 introduces more efficient loss functions and anchor-free prediction mechanisms, enabling it to handle variable object sizes and shapes with increased precision. YOLO-v8 achieves state-of-the-art performance in both accuracy and inference speed, making it a competitive choice among single-stage detectors. Based on these generic object detectors, detectors aim to search specific class of object from images have been developed. Readers can refer to [92], [102] for more details on generic object detection.

The key challenges associated with generic object detection stem from the difficulty of accurately identifying objects across a wide variety of classes while maintaining efficiency in inference speed. One major problem lies in the trade-off between accuracy and speed. Specifically, two-stage detectors such as Faster R-CNN [95], produce high accuracy but are computationally expensive due to the dependence on region proposals and subsequent classification. This makes them slower and less suited for real-time applications. By contrast, one-stage detectors like YOLO [98] and SSD [99] are designed to be faster by avoiding the region proposal stage and directly predicting bounding boxes and class probabilities. However, this often results in lower precision, particularly when detecting smaller or cluttered objects.

Another key issue is the difficulty in balancing the detection of objects across various scales and aspect ratios. Object detectors must be robust across different sizes, orientations, and occlusions of objects, which complicates the task. For instance, while SSD benefits from multi-resolution detection, ensuring consistency across scales remains a significant challenge. Furthermore, detecting objects in cluttered or complex environments introduces further difficulty, as occlusion and background noise can mislead detectors, resulting in false positives or missed detections. Therefore, improving the ability of detectors to cope with such variations without sacrificing speed or accuracy is a persistent problem in object detection.

#### 3.4.2.2. Class-Specific Object Detection Challenge

Compared with generic object detection, the objective of class-specific object detection is to detect a specific class of object such as cyclist, pedestrian or vehicle. In the PCW feature of ADAS, the class-specific object detection enables the vehicles to detect the appearance of the cyclist, pedestrian or vehicle in front of it. When the PCW determines that the probability of a frontal collision with the detected frontal pedestrian, cyclist or vehicle is high, it activates the visual and audible alerts to remind the driver to take evasive action. If the system detects the driver failed to take evasive action, the AEB system can be applied automatically to stop the vehicle.

Besides, if an insufficient braking input is detected, the system can increase the braking force to provide full braking response. Therefore, it can help reduce the risk of a frontal collision. In real-world environments, cyclists, pedestrians and vehicles may be moving in any direction.

As a result, the possibilities of the shape of these objects are unlimited. Additionally, different dressing styles or colors of pedestrians and cyclists, and different colors of vehicles, makes it complex to represent cyclists, pedestrians and vehicles with a unique set of templates. In this subsection, we review algorithms for cyclist detection, pedestrian detection and vehicle detection. Those algorithms have characteristics and challenges in the real-world, such as vastly different scales, poor appearance conditions, and extremely severe occlusion in crowd scenarios [103].

Class-specific object detection, particularly for detecting cyclists, pedestrians, and vehicles, faces several critical challenges in real-world environments. One major issue is the variability in the appearance and movement of these objects. Cyclists, pedestrians, and vehicles can move in any direction, and their shapes can vary significantly due to different orientations and poses. Moreover, the variety in clothing styles, colors, and vehicle types increases further complexity, making it difficult to represent these objects with a fixed set of templates. Furthermore, detecting these objects in adverse conditions, such as varying lighting, weather, and crowded scenes, is particularly challenging. Severe occlusion, especially in cluttered environments, further improves detection difficulty, as parts of the objects may be obscured, leading to potential false negatives or inaccurate localization. These problems highlight the requirement for robust algorithms that can adapt to the dynamic nature of real-world scenarios and ensure reliable detection in complex, unpredictable environments.

### 3.4.2.2.1. The Pedestrian Detection Problem

Pedestrian detection refers to the task of detecting pedestrians from images, which is a basic component of the PCW system. Besides, the automotive night vision system in some certain premium vehicles also featured pedestrian detection [104]. In the field of computer vision, Dalal and Triggs [105] proposed the classical pedestrian detection method that combines histograms of oriented gradients (HOGs) and linear support vector machine (SVM). This work is a milestone in pedestrian detection and has been cited more than 46,000 times. The proposed method produces promising accuracy, but it is difficult to run in real-time. Besides, Zhang et al. [106] analyzed the relation between body parts and different channels of features produced by pedestrian detector and proposed to use channel-wise attention to solve the occlusion problem for pedestrian detection. Later, Li et al. [107] developed a YOLO-based method for pedestrian detection in hazy weather. Furthermore, they collected a data set that includes 1195 pedestrian images in hazy weather. This data set is further augmented through six image augmentation techniques to train the developed pedestrian detector. In 2020, Zhang et al. [103] designed a pedestrian detector for crowded scenes. In particular, they treat pedestrian detection as a feature detection problem that combines semantic features to model the semantic differences between each instance in crowed environments.

The abovementioned methods all detect pedestrians from RGB images. Compared to RGB cameras, thermal cameras are insensitive to ambient light and capture less texture. Therefore, they are robust in bright sun glare scenarios. In 2020, Nowosielski et al. [108] developed a nighttime pedestrian detection system for supporting the driver during night driving. The developed system detects pedestrians from thermal images through YOLOv2 detector in an ODROID XU4 microcomputer platform. Later, Kim et al. [109] introduced an uncertain-aware multi-modal (color and thermal) pedestrian detection framework, which includes an uncertainty-aware feature fusion (UFF) module and an uncertainty-aware cross- modal guiding (UCG). Based on the aleatoric uncertainty that measures the uncertainty in the observations, the UFF defines a Region of Interest (RoI) uncertainty to quantify the ambiguity of the detected RoIs. In addition, the UCG applies predictive uncertainty to alleviate the discrepancy between the color modality and thermal modality, which makes the feature distributions of the two modalities become similar. Therefore, the features of the pedestrians and background are easily distinguished. Recently, Dasgupta et al. [110] designed a multimodal feature fusion-based pedestrian detection method. To fuse the features extracted from RGB and thermal images, a feature embedding module is designed to get the multimodal features. Then, the multimodal features are passed to the detection decoder to produce pedestrian bounding boxes.

### 3.4.2.2. The Cyclist Detection Problem

In one of the early studies on cyclist detection, a vision-based cyclist detection method was developed by Tian et al. [111]. The authors applied cascaded detectors with different classifiers and shared features to detect cyclists from multiple viewpoints. One year later, Li et al. [112] collected a stereo vision-based cyclist detection data set that includes 22161 annotated cyclists instances. Besides, they designed a stereo-proposal based Fast R-CNN (SP-FRCN) to detect cyclists in images. The SP-FRCN uses stixel representation to generate region proposals from stereo data.

Meanwhile, Li et al. [113] proposed a unified framework to simultaneously detect cyclists and pedestrian from images. The proposed framework applies a detection proposal method to produce a series of object candidates. Then, these object candidates are fed to a Faster R-CNN based model for classification. Finally, a post-processing step is used to further improve the detection performance. Wang and Zhou [114] proposed a Fast R-CNN [94] based unified framework for cyclist and pedestrian detection in driving environments. The proposed framework uses a multilayer feature fusion method to tackle the challenges of small-sized targets and changeable background environment. Two years later, Annapareddy et al. [115] proposed a pedestrian and cyclist detection method from thermal images through Faster R-CNN. The proposed method produces promising results on the KAIST Multispectral

Pedestrian dataset [68].

### 3.4.2.3. The Vehicle Detection Problem

García et al. [116] proposed a sensor fusion method for detecting vehicles in inter-urban scenarios. The proposed method applies the unscented Kalman filter (UKF) and joint probabilistic data association to fuse the data from 2D LiDAR and monocular camera and achieves promising vehicle detection results in single-lane roads. In [117], Yang et al. presented a YOLOv2 based real- time detector for the joint detection of pedestrian and vehicle. Wang et al. [118] performed a comparative evaluation for five popular deep learning-based object detectors, (e.g., Faster R-CNN [95], R-FCN [119], SSD [99], RetinaNet [120], and YOLOv3 [24]) in vehicle detection on the KITTI dataset [64]. They compared the performance of these detectors in terms of the detection time, recall, and precision metrics. We suggest readers refer to [118] for more details. Wu et al. [121] presented a fully convolutional neural network, named SqueezeDet, to simultaneously detect vehicle, pedestrian and cyclist in images. Being designed as a single-stage detector and using the SqueezeNet as the backbone, SqueezeDet achieves real-time speed (57.2 fps on an Nvidia Titan X GPU) and reduces the model size for energy efficiency.

Chen et al. [122] constructed a lightweight vehicle detector which achieves three-times faster than YOLOv3 [24] while only having 1/10 size of model. Murthy et al. [123] proposed a lightweight real-time method for pedestrian and vehicle detection and named as EfficientLiteDet. EfficientLiteDet is built on top of Tiny-YOLOv4 through inserting one more prediction head to achieve multi-scale object detection. The conventional vehicle detection methods depend on directly visible vehicles in images, which is a drawback com- pared to human visual perception. Because humans usually use visual cues caused by objects to reason about information or anticipate occurring objects. This phenomenon is more obvious in nighttime driving scenarios where human drivers foresee the oncoming vehicles through analyzing illumination changes in the environment or the light reflections caused by the headlamps of oncoming vehicles [124]. Drivers utilize this provident information to adapt their driving operation accordingly, for example, switch the high beam to low beam in advance to avoid glares for the oncoming drivers. While the computer vision systems are usually trained to solve one specific task, which is formulated as a mathematical problem. For instance, in object detection, the objects are annotated by bounding boxes, and the task is to predict and classify these bounding boxes [92].

According to [125], human drivers detect the oncoming vehicles on average 1.7s faster than the computer vision system. This non-negligible time discrepancy could be attributed to the characteristic that ordinary object detection systems assume that objects have clear and visible boundaries. To solve the discrepancy between human and ordinary vehicle detection algorithms, especially the vehicle detection at nighttime, many researchers presented their works in provident vehicle detection

(PVD) [30], [74], [124]–[126]. PVD is a technique that detects the appearance of vehicles through the light reflections caused by their headlamps. It is the foundation of the HBA system which uses a front-mounted camera located in the upper-portion of the windscreen to detect the light sources head of the vehicle and automatically switch the headlamps between low beams and high beams to avoid blinding of oncoming drivers [127].

### 3.4.2.4. The Lane Detection Problem

The task of lane detection is to detect the lane areas or lane markings through camera or LiDAR [128]. Lane detection allows the vehicle to properly localize itself within the road lanes, it is a fundamental component for *LDW* and *LKA* systems, minimizing the chances of collision. The LDW system detects the lane markings while the vehicle is on a straight or slightly curved road. When the LDW system determines that the vehicle deviates from its lane, it notifies the driver through audible and visual alerts. While LKA is more advanced than LDW, as it can apply corrective steering to help guide the vehicle back to the middle of detected lanes. According to the type of sensing sensors, the current lane detection methods can be categorized into three types, camera-based methods, LiDAR-based methods, and multi- modal fusion-based methods.

In 2014, Kim and Lee [129] developed a lane detection method that combines a CNN with random sample consensus (RANSAC) algorithm. The RANSAC algorithm works through randomly selecting a subset of samples from the given data set and uses the selected samples to estimate model parameters. This process is repeated a large number of times until the best model is found. CNN is used to extract lane candidates in the image. Subsequently, the extracted lane candidates are passed to the RANSAC algorithm to detect road lanes. The proposed method can be regarded as an approximation of the mapping function between the input and output. Two years later, Gurghian et al. [130] proposed an image classification-based lane detector and named as DeepLanes. DeepLanes is a deep CNN that trained on a data set consisting of RGB images from two laterally mounted down-looking cameras. Benefiting from the more complex network, DeepLanes achieves better performance than [129]. However, it depends on laterally mounted down-looking camera, which limits its application scenario.

Neven et al. [131] formulate lane detection as an instance segmentation problem where each lane is treated as an in-stance within the lane class. They designed an end-to-end multi-task network which consisting of a lane segmentation branch and a lane embedding branch. The lane segmentation branch produces a binary lane mask indicating which pixels are located in a lane and which not. The lane embedding branch clusters the segmented lane pixels into different lane instances. By splitting the lane detection task into two steps, the proposed method alleviates the lane change problem and can detect a variable number of lanes. Recent advancement of object detection motivates researchers detect lanes through detecting a series of points (e.g., every 10 pixels in the vertical axis) [132]. Inspired from the region-based object detector, Faster R-CNN [95], Li et al. [133] developed a one-stage lane line detector, named Line-CNN. Line-CNN runs at about 30 fps on an Nvidia Titan X GPU. Later, Tabelini et al. [132] proposed an anchor-based mechanism to aggregate global information for lane detection. It achieves state-of-the-art accuracy performance through using a lightweight backbone network.

The camera-based lane detection methods can meet the high frame rate requirements of driving scenes. Using RGB cameras, these methods are sensible to environment illumination, especially the dramatic changes in light. Therefore, their performance may decrease considerably at nighttime. LiDAR sensors perceive the environment through emitting light, which are not sensitive to environment illumination. Hence, lane detection also has been solved through using LiDAR measurement as the input [134], [135]. Hata and Wolf [134] proposed an Otsu thresholding-based method to segment LiDAR point clouds into asphalt and road markings. [135] cast road area detection as a pixel-wise semantic segmentation task in point cloud's bird's eye view (BEV) images through a fully convolutional network (FCN).

Compared to camera, LiDAR provides accurate distance measurement and retains rich 3D information in the environment. However, it only produces sparse and irregular point cloud data, which can result in the existence of empty voxels. Therefore, multi-modal fusion-based methods have been developed. Bai et al. [136] introduced a method that combines camera with LiDAR to detect lane boundaries in 3D space. They first convert the point cloud data to BEV and predict a dense ground height using a CNN. The predicted dense ground height is then fused with the BEV image to perform lane detection. Zhang et al. [137] designed a channel attention-based multi-modal information fusion method for lane detection. Unlike [136], they fused features learned from RGB image and point cloud data through a channel attention mechanism that enables camera and LiDAR fusion information to be used simultaneously across channels.

### 3.4.2.5. The Traffic Sign Recognition Problem

Traffic signs are signs put at the side of roads bearing symbols or words of warning or direction to pedestrians and drivers. A traffic sign recognition system usually concerns two related subjects: *traffic sign recognition (TSR)* and *traffic sign detection (TSD)*. TSR is a fine-grained classification to identify the type of the detected traffic signs, while TSD aims to localize the traffic signs in an image. We review publications on TSR and TSD in this subsection. In autonomous driving systems, TSR is a safety component that recognizes traffic signs through a camera and conveys the information displayed on the sign to the driver via the multi-information display. TSR helps prevent the driver from overlooking traffic signs. The current ADAS systems apply TSR algorithm to recognize *speed limit, do not enter, and traffic stop signs*. When TSR determines the vehicle speed exceeds the speed limit sign indicated in the active driving display, the system notifies the driver through visual and audible warning. Therefore, it can enhance driving safety and comfort by helping drivers adapt the maximum speed of the vehicle to a particular limit.

Both TSR and TSD have been explored by researchers from the communities of computer vision and autonomous driving. In 2011, Stallkamp et al. [138] introduced the German Traffic Sign Recognition Benchmark (GTSRB), a large scale and real-world data set containing 50,000 traffic sign images in 43 classes. Two years later, Houben et al. [139] released the German Traffic Sign Detection Benchmark (GTSDB) which has 900 images containing 1206 traffic signs. These two data sets enabled researchers to analyze and compare the performance of different algorithms on the same benchmarks. It is worth noting that traffic signs in the GTSRB benchmark occupy most of the image, algorithms only need to classify the subclass of the sign. Furthermore, the GTSDB benchmark only annotated four categories of traffic signs. Therefore, these benchmarks are not representative for the real-world tasks where traffic signs in an ordinary image are usually less than 1% of the image [66].

In 2016, Zhu et al. [66] collected a large-scale traffic sign data set from Tecent Street View panoramas, named TT100K. The TT100K data set has 100000 images containing 30000 traffic sign instances. Based on the TT100K data set, they trained two CNNs for TSR. One year later, Luo et al. [140] proposed a TSR system to recognize both symbol-based and text-based signs in video sequences. They first use MESRs to extract traffic sign regions of interest (ROIs) from images. Then, a multi-task CNN is trained to refine and classify the ROIs. Lee and Kim [141] designed a CNN to simultaneously detect the position and boundary of traffic signs.

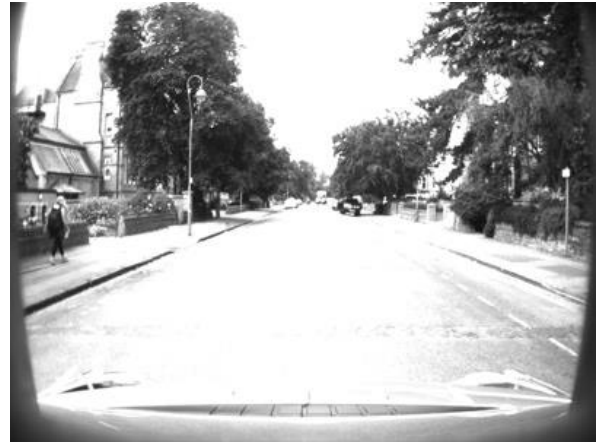

Fig. 3. Example image with sun glare. The image is from the RobotCar data set [163].

Meanwhile, Li and Wang [142] designed a real-time TSR method through combining Faster R-CNN [95] with MobileNet [22]. Furthermore, they applied the color and shape information to refine the localization of small traffic signs. Kamal et al. [143] formulated the TSD as an image segmentation problem and designed a modular CNN architecture that stacks SegNet and U-Net to solve it. To tackle the mis- recognition of small traffic signs in the image, Min et al. [144] combined the semantic scene understanding and structural traffic sign location for TSR. They designed a lightweight RefineNet to segment objects from the scene to obtain the information regarding the spatial positional at pixel level. Subsequently, a scene structure model which is based on the constraints of spatial positional relationships between traffic signs and other objects is built to establish the trusted search regions. Experimental results demonstrated that the proposed method could alleviate the mis-recognition of small traffic sign in straight road and curvy road scenes. While for complex scenes such as intersections, it still has ineffective recognition.

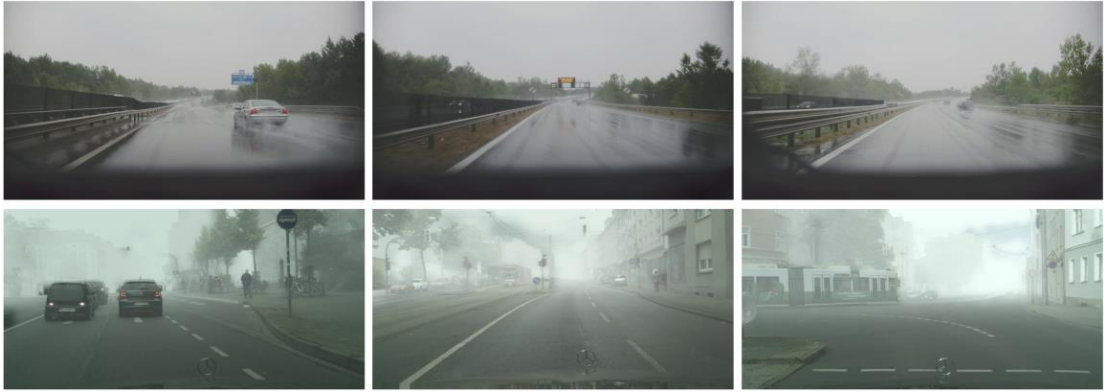

Fig. 4. Examples of adverse weather. From top to bottom: images with rainy weather [78] and images with foggy weather [161].

### 3.4.3. Additional challenges and problems for environment perception

#### 3.4.3.1. The Sun Glare Problem

Sun glare is a commonly encountered environmental hazard, it brings about over-exposure in the image and degrades the performance of computer vision algorithms [162]. In autonomous driving scenarios, the influence of sun glare can be classified into two categories, direct and indirect. The direct influence occurs in cases where the sun is low, and the glare directly hits the onboard camera. For the indirect influence, it results from the sunlight reflected from the wet road or highly specular surface. The indirect influence may result in the detection of lane boundary or road markings impossible, because the region with the glare effect is overexposed. In some situations, the misdetection of lane markings may negatively influence the decision on driving direction of autonomous vehicles.

#### 3.4.3.2. The Adverse Weather Problem

Autonomous driving systems typically depend on cameras and LiDAR to sense the environments around vehicles. However, in cases where the weather is poor, such as heavy rain or thick fog, the information captured by these sensors can be disrupted and thereby impact the accuracy of the detection. Specifically, cameras tend to perform poorly in low-visibility conditions like heavy fog, rain, or snow, where the captured images become blurred. LiDAR is also particularly impacted by adverse weather, as laser pulses can be scattered by rain, snow, or fog, leading to noisy or incomplete data that complicates accurate object detection. These conditions pose significant challenges for both sensors, as degraded detection accuracy may result in incorrect driving decisions and compromise the safety of autonomous driving [60]. Therefore, autonomous systems must be designed with advanced algorithms and feature fusion techniques to mitigate these effects and maintain reliable performance in adverse weather.

#### 3.4.3.3. The oversimplified testing conditions problem

Until August 2020, there were five fatalities happened for level 2 autonomous driving [63]. Among those fatalities, four of them were from Tesla and one from Uber. Specifically, all four Tesla accidents were attributed to perception failure, while Uber's fatality resulted from the system's failure to correctly detect pedestrian behavior. These failures highlight the limitations of autonomous systems in real-world environments, which are far more complex and unpredictable than the controlled conditions often used in field testing, such as good weather or light traffic.

The failure of perception systems can be attributed to several factors. Autonomous driving relies heavily on sensors, such as cameras, LiDAR, and radar, to perceive the surroundings. However, these sensors are susceptible to various challenges in real-world scenarios. For instance, dynamic environments with a variety of pedestrians, cyclists, vehicles, and unpredictable human behavior can lead to misclassification. Poor weather conditions, such as rain, fog, or snow, further degrade sensor performance, while occlusion by other vehicles or obstacles can prevent the system from fully perceiving critical elements in the environment.

Furthermore, the complexity of the world lies in the variety of objects, diverse road structures, and ever-changing environmental conditions, all of which create scenarios that autonomous systems may not have encountered during training [189]. This complexity makes it difficult for systems to generalize effectively, leading to detection errors and incorrect decision-making. As a result, the gap between controlled testing environments and real-world conditions poses a significant challenge to the development and safety of autonomous driving systems. Therefore, improving the robustness of perception algorithms and ensuring better real-world adaptability remains crucial.

#### 3.4.3.4. The Data Size Problem

The size of the datasets to be manipulated represents a problem in terms of both storage capability and real-time processing speed. In order to achieve the objective of fully autonomous driving, autonomous vehicles are equipped with multiple sensors such as camera, LiDAR, radar, and ultrasonic sensor to perceive the driving environment. According to [61], an autonomous vehicle produces about 4000 GB of data a day, which is equal to the mobile data produced by almost 3000 people. The huge amounts of data pose significant challenges to communication, storage, and computing platforms [163]. Although onboard computing and storage technologies have been developed rapidly, they are still limited compared with the scale of data to be stored and processed. To achieve better driving performance than the best human driver who takes actions within 0.1 to 0.15s, the autonomous driving systems must achieve a real-time running speed in real-world traffic environment within 0.1s [164]. This requires a significant amount of computing power. Although the high-performance GPUs can provide the low latency computation, their high-power consumption (e.g., the power of Nvidia Drive AGX is 300W) may significantly degrade the driving range and fuel efficiency of autonomous vehicles.

#### 3.4.3.5. The Affordability Problem

High costs associated with technology discourage users from buying PAVs and, subsequently, deter developers from investing in R&D of new models. High costs may derive from the inherent price of the raw materials, the hardware components (especially sensors), or the software development (especially those require large datasets and great computing power and time for training and optimization) and testing for compliance. Autonomous vehicles depend on a series of onboard devices to support their normal functions. In addition to various sensors such as cameras, LiDAR, radar, and ultrasonic sensor, autonomous vehicles also require communication devices, computing platform, and extra power supply. According to [165], the average cost to build a conventional non-luxury vehicle in the US is around $30000, while for a fully autonomous vehicle, the total cost is increased to $250000. However, some surveys [148], [149], [154] on public opinion on autonomous vehicles demonstrate that a majority of people are unwilling to pay extra money for autonomous vehicles (see Section 9). To gain the trust of consumers, automotive manufacturers must explore new solution to reduce the cost of sensors mounted of PAVs.

### 3.5. Future Directions and Promising Approaches to Environment Perception

#### 3.5.1. Universal Data Sets for Long-term Autonomous Driving

Autonomous vehicles significantly rely on extensive real-world data to develop, test, and validate the performance of algorithms. However, as shown in Table II, existing data sets collected from 2013 to 2023 largely focus on immediate algorithmic needs, emphasizing controlled scenarios in specific urban areas. This narrow focus limits their applicability to the diverse and evolving challenges in long-term autonomous driving, particularly the variation in visual perception caused by environmental changes, such as seasonal shifts and construction [184] [185]. Addressing this limitation requires large-scale, diverse data sets that capture a wide range of driving conditions across locations and timeframes, providing critical insights for both academia and industry. The primary challenge lies in the sheer scale and complexity of such data collection, including the difficulty of capturing and maintaining data that reflects a constantly changing environment.

#### 3.5.2. Mobile Edge Computing for Autonomous Vehicles

To support the safety and reliability of PAVs, onboard sensors produce vast quantities of data that must be processed by deep neural networks (DNNs) at real-time speeds. However, balancing the high-volume cost of computing devices with the processing power needed for these complex models presents a major challenge. Mobile edge computing (MEC) emerges as a promising solution, integrating telecommunications with cloud capabilities to offer processing services directly from the network edge [188]. This approach reduces latency, supporting the delay-sensitive applications critical for AV operations. The edge servers serve as processing hubs, while PAVs act as clients accessing this processed data. However, MEC introduces challenges, including maintaining network reliability, managing data transfer latency, and ensuring robust security at the edge, all essential to meet the safety requirements of autonomous driving.

#### 3.5.3 Real-time and Lightweight CNNs for Autonomous Driving

To enhance perception accuracy, autonomous vehicles usually employ CNNs to process sensor data from cameras and LiDAR. In general, the development trend of CNNs is to design very deep networks to boost accuracy [189]. However, these networks demand significant memory and computational power, resulting in challenges for AV applications that rely on resource-constrained hardware. Developing real-time, lightweight CNNs address these constraints by enabling faster, more energy-efficient models that enhance safety and robustness. The primary challenge in creating these lightweight architectures is achieving a balance between reduced computational demands and maintaining sufficient accuracy to detect objects, pedestrians, and obstacles under diverse conditions.

#### 3.5.4. Risk Assessment for Autonomous Vehicles

The performance of PAVs is significantly affected by varying weather, lighting, road condition, and other highly dynamic aspects of the real-world driving environment, such as the unpredictable behaviors of other vehicles, pedestrians, and cyclists. These critical factors introduce substantial variability and uncertainty, increasing the possibility of autonomous driving errors. Addressing these issues requires sophisticated risk assessment mechanisms that can adapt to and anticipate changes in environmental conditions and human behavior [186]. Without such measures, PAVs may be less able to respond appropriately to sudden changes, resulting in increased risk to passengers and other road users.

To enhance safety, risk assessment algorithms should focus on monitoring and evaluating specific elements that impact autonomous decision-making [186]. This includes real-time analysis of weather patterns, changes in visibility, and road surface quality, which may impact vehicle stability and sensor accuracy. In addition, these algorithms should assess the behavior of surrounding agents such as vehicles, pedestrians, and cyclists to predict potential conflicts or unexpected maneuvers. Through continuous data collection and advanced predictive modeling, risk assessments can identify hazardous situations, allowing PAVs to respond ahead of time. This aims to reduce accidents by analyzing dangerous conditions and activating adaptive responses, ultimately contributing to a safer autonomous driving experience (see Section 8).

### 3.5.5. Conclusions and the Way Forward

PAVs rely heavily on visual and sensory data to understand their environment and make decisions. Traditional perception systems primarily use separate modalities, such as cameras, LiDAR, and radar, to perceive scenes. However, to enhance situational awareness (see Section 8) and contextual understanding (Sections 4 and 5) is crucial to appropriately interpret and navigate complex driving scenarios. As we will discuss in detail in Section 5, integrating vision-language models probably represents the most promising approach to achieve this goal. Multimodal large vision-language models, which can process and interpret data from various sources and associate it with language-based context, represent a promising future direction for autonomous driving [187]. As discussed in greater detail in Section 5, these models enable PAVs to understand more abstract cues, such as road signs, pedestrian behaviors, and complex scene elements that require both visual features and semantic context. By leveraging language-based prompts and descriptors, these systems could interpret subtle situations, providing PAVs with a deeper and more adaptable understanding of their surroundings.

Considering the high computational demands and hardware limitations in PAVs, the primary challenges lie in developing and training large-scale multimodal models capable of real-time processing. In addition, ensuring these models' interpretability and reliability across varied driving conditions, such as different weather, lighting, and traffic scenarios, remains a critical challenge. In the future, overcoming these challenges could improve AV safety and decision-making by enabling a richer understanding of dynamic environments.

# 4. AI for Path Planning and Mapping

## 4.1. Introduction and Methodology

Path planning and mapping are foundational to the operational capabilities of autonomous vehicles (AVs), enabling safe and efficient navigation through dynamic environments. Path planning involves the computation of feasible, collision-free trajectories, while mapping focuses on constructing accurate representations of the vehicle's surroundings. Both tasks are inherently challenging due to uncertainties in dynamic environments, sensor inaccuracies, and computational constraints. This section critically reviews AI-driven approaches to path planning and mapping, emphasizing their contributions, limitations, and unresolved challenges.

Publications reviewed in this section were selected based on their relevance to AI applications in path planning and mapping for PAVs. Searches were conducted using IEEE Xplore, Google Scholar, and SpringerLink with keywords such as "path planning," "autonomous vehicles," "AI for mapping," "reinforcement learning in AV," and "SLAM for AV." Emphasis was placed on peer-reviewed journals and high-impact conferences from 2010 to 2024. Widely cited technical reports and preprints were included selectively to ensure groundbreaking contributions were considered.

## 4.2. Path Planning Challenges and Attempted Solutions

### 4.2.1. Real-time Decision-Making in Dynamic Environments

Autonomous vehicles must make rapid, accurate decisions while navigating dynamic environments with moving obstacles, erratic human behaviour, and fluctuating traffic conditions. These challenges, exacerbated by computational constraints and unpredictable scenarios, require advanced solutions that traditional rule-based systems struggle to address. Attempted solutions include Reinforcement Learning, Model Predictive Control, Behaviour Cloning.

#### 4.2.1.1. Reinforcement Learning (RL)

Reinforcement Learning (RL) [230] is a data-driven approach that enables PAVs to learn optimal driving behaviours through interactions with simulated environments. In RL, an agent explores different driving scenarios, receiving rewards for successful actions (e.g., avoiding collisions) and penalties for failures (e.g., violating traffic rules). Over time, this trial-and-error process allows the agent to develop policies that maximize long-term safety and efficiency. For example, RL models such as Deep Q-Networks (DQN) and Proximal Policy Optimization (PPO) has been applied to train PAVs for tasks such as merging into dense traffic by simulating millions of scenarios where vehicles must balance caution with assertiveness [190, 191, 193, 194]. Despite its adaptability, RL faces challenges in real-world deployment, including the need for significant computational resources and struggles with rare or unforeseen scenarios.

#### 4.2.1.2. Model Predictive Control (MPC)

MPC formulates path planning as an optimization problem, predicting future vehicle states over a finite time horizon to generate optimal trajectories [195]. This method excels in handling dynamic constraints, such as navigating intersections and highway lane changes, where it continuously adjusts trajectories to ensure safety and compliance with traffic rules [191]. However, computational demands can limit real-time applicability, prompting recent advances like parallel processing and adaptive horizon lengths to mitigate this issue [196, 200].

### 4.2.1.3. Behaviour Cloning

Behaviour cloning leverages supervised learning to train PAVs to mimic human drivers. Using large datasets of recorded human driving behaviours, this approach creates models that map sensory inputs to corresponding driving actions [197, 198, 201]. For example, Tesla's Autopilot system employs elements of behaviour cloning to handle tasks like lane following and stop-and-go traffic [202]. While it facilitates rapid policy development, its reliance on training data limits generalization to unfamiliar scenarios, posing safety risks. Enhancements, including diverse data augmentation and fail-safe mechanisms, aim to enhance robustness.

### 4.2.1.4. Hybrid Approaches

Reinforcement Learning offers adaptability for complex behaviour but requires substantial real-world deployment challenges. MPC provides a robust framework for dynamic constraints, though computationally intensive. Behaviour Cloning accelerates policy development but risks overfitting. Integrating RL, MPC, and Behaviour Cloning can harness their respective strengths, enhancing AV path-planning robustness and efficiency. Combining RL with MPC integrates the adaptability of RL with MPC's trajectory optimization capabilities. For example, hybrid models have demonstrated up to a 15% reduction in energy consumption and smoother transitions in lane changes compared to standalone RL systems in eco-driving scenarios in dense traffic by balancing strategic planning and trajectory control [203, 204, 205, 206]. These findings underscore the computational trade-offs necessary to achieve a balance between high-level strategic planning and the demands of real-time operational responsiveness.

### 4.3. Handling Sensor Uncertainty and Failures

Environmental factors such as fog, rain, or sun glare can degrade the performance of sensors like LiDAR, radar, and cameras. These inaccuracies affect obstacle detection, lane recognition, and overall situational awareness. Sensor failures or misreading may lead to poor decision-making and unsafe trajectories. Addressing this issue requires robust algorithms and systems that can account for and mitigate uncertainty. Solutions include Probabilistic Path Planning Algorithms, Sensor Fusion, AI-augmented perception, and Redundant Sensory Systems.

### 4.3.1. Probabilistic Path Planning Algorithms

Probabilistic algorithms, such as Partially Observable Markov Decision Processes (POMDPs), explicitly account for uncertainties in sensor inputs. POMDPs model possible states of the environment and compute policies that maximize expected outcomes despite incomplete or noisy data. For example, POMDP-based approaches have been used to navigate PAVs in low-visibility conditions by considering probabilities of obstacles being present in ambiguous regions. These algorithms enhance robustness but often require significant computational resources to evaluate multiple possibilities in real time. [207, 208].

### 4.3.2. Sensor Fusion

Sensor fusion techniques combine data from multiple sensors, such as LiDAR, cameras, and radar, to create a more reliable representation of the environment. By integrating inputs from sensors with complementary strengths, sensor fusion reduces the impact of individual sensor failures. For instance, radar performs well in adverse weather, while cameras excel in detecting fine details under clear conditions. Fusion algorithms, such as Bayesian inference or neural network-based methods, merge these inputs to provide PAVs with robust situational awareness [209, 210, 211, 212].

### 4.3.3. AI-Augmented Perception

AI models can augment sensor capabilities by predicting or reconstructing missing data. Deep learning algorithms trained on large datasets of diverse weather conditions can infer likely obstacles or road markings, even when direct observations are obscured. For example, convolutional neural networks (CNNs) have been used to reconstruct lane markings in foggy conditions by analyzing visible patterns in adjacent areas [213].

### 4.3.4. Redundant Sensor Systems

Redundancy in sensor systems mitigates the risk of single-sensor failures. For instance, combining stereo cameras with radar ensures that critical information, such as obstacle distance and speed, is still available even if one system malfunctions. This approach enhances system reliability but increases hardware costs and power consumption [212].

Sensor uncertainty and failures remain critical challenges for PAVs, as perception systems must reliably interpret environmental data under diverse conditions. Probabilistic path planning algorithms, such as POMDPs, effectively handle uncertainty by modelling multiple possible environmental states, making them suitable for low-visibility scenarios. However, their high computational complexity limits real-time applications, requiring further optimization [214, 215]. Sensor fusion offers a robust solution by combining data from LiDAR, radar, and cameras to reduce the impact of individual sensor failures. Neural network-based fusion techniques have shown significant improvements in adverse weather conditions, though challenges remain in calibration and cost. Similarly, AI-augmented perception enhances situational awareness by reconstructing missing or obscured data using deep learning. While effective, these models depend on diverse training datasets to ensure reliability in edge cases. Redundant sensor systems provide fail-safes against single-sensor failures, increasing system reliability but at a cost of higher complexity and energy demands. Advances in lightweight and cost-efficient sensors could make this approach more viable for mass-market PAVs.

### 4.4. Balancing Safety and Efficiency

PAVs must balance safety and efficiency, particularly in scenarios like merging, overtaking, and navigating intersections. Overly cautious behaviour may lead to inefficient driving, causing delays and disrupting traffic flow, while aggressive manoeuvres can increase the risk of collisions. Striking the right balance between these competing objectives is a critical challenge for PAVs, requiring advanced algorithms that can adapt to dynamic environments while maintaining both safety and operational efficiency. Solutions include Game-theoretic approaches, Risk-aware Planning Algorithms, Driver Behaviour Prediction, and Personalized Driving Strategies.

#### 4.4.1. Game-Theoretic Approaches

Game-theoretic approaches model interactions between PAVs and other road users as multi-agent games. These models optimize decision-making by predicting the actions of other agents (e.g., human drivers) and selecting strategies that minimize conflict. For example, researchers have used Nash equilibria to enable PAVs to choose manoeuvres that are both safe and efficient in scenarios like highway merging and roundabouts [206, 207]. While game-theoretic models excel in representing interactive behaviours, they can be computationally expensive in real-time applications, especially with multiple agents [208].

#### 4.4.2. Risk-Aware Planning Algorithms

Risk-aware algorithms quantify safety metrics, such as time-to-collision or safe following distance, and integrate these into the decision-making process. These models prioritize safety by penalizing high-risk actions while optimizing for efficiency within acceptable risk thresholds. For instance, adaptive risk metrics allow PAVs to navigate aggressively in low-risk conditions (e.g., sparse traffic) but cautiously in high-risk environments (e.g., crowded intersections). Studies have shown that this approach enhances both safety and throughput in urban traffic scenarios [219].

#### 4.4.3. Driver Behaviour Prediction

Understanding and predicting the behaviour of human drivers is essential for balancing safety and efficiency. Machine learning models trained on real-world driving data can anticipate lane changes, braking, or acceleration patterns of surrounding vehicles. These predictions may enable PAVs to adjust their actions, such as yielding to aggressive drivers or merging into faster lanes. Recent advancements in recurrent neural networks (RNNs) and transformers have improved the accuracy of such predictions, leading to smoother and safer interactions on the road [220, 221].

#### 4.4.4. Personalized Driving Strategies

Personalized driving strategies allow PAVs to adjust their level of assertiveness based on user preferences or the situational context. For example, some users may prioritize safety over efficiency, while others may prefer more assertive behaviours to minimize travel time. These strategies are implemented using customizable parameters in the decision-making algorithms, enabling PAVs to dynamically adapt to individual preferences without compromising general safety. For example, studies conducted on personalized driving behaviour for PAVs, proposing a user-oriented approach that adjusts driving assertiveness based on passenger preferences [222]. Decision-making algorithms incorporate customizable parameters to adapt the AV's level of assertiveness. For example, a passenger might indicate a preference for highly cautious driving in busy urban areas, while another might prioritize efficiency on highways to reduce travel time.

The balance between safety and efficiency is critical for ensuring that PAVs integrate seamlessly into mixed traffic environments without causing disruptions or safety concerns. Game-theoretic models provide foundation for strategic decision-making in interactive scenarios, but their computational demands limit real-time applications. Risk-aware algorithms address this by quantifying safety metrics and optimizing for efficiency within defined thresholds, making them practical for real-time deployment. However, these models often require fine-tuning to avoid overly conservative behaviour that reduces throughput. Driver behaviour prediction adds a proactive dimension to AV decision-making, enabling smoother interactions with human drivers. The use of machine learning models to anticipate human actions has proven effective in reducing conflicts and enhancing traffic flow.

Similarly, personalized driving strategies contribute to user satisfaction, a key factor for public acceptance of PAVs. By allowing dynamic adjustment of assertiveness levels, these strategies can address diverse user expectations while maintaining overall safety. Combining these solutions offers the most promise for balancing safety and efficiency. For instance, integrating driver behaviour prediction with risk-aware decision-making can improve AV adaptability in complex scenarios. Furthermore, advancements in computational efficiency and real-time processing will be essential for deploying game-theoretic models at scale. Addressing these challenges will require interdisciplinary collaboration across fields such as AI, human factors, and traffic engineering.

### 4.5. Real-Time Localization and Mapping Challenges

PAVs depend on accurate, up-to-date maps for navigation and decision-making. Creating and maintaining these maps in real-time remain two critical challenges, particularly in dynamic environments where the landscape changes frequently. Below, we address two major challenges in mapping and explore solutions supported by recent advancements in the literature. Simultaneous Localization and Mapping (SLAM) is a critical component for PAVs to perceive their environment and localize within it. SLAM algorithms allow PAVs to build maps of their surroundings while maintaining real-time awareness of their position. However, achieving high accuracy under dynamic and cluttered conditions remains a significant challenge. Moving objects, changing environments, and computational constraints often degrade SLAM performance. Possible solutions include AI-Augmented SLAM and Multi-Modal SLAM.

### 4.5.1. AI-Augmented SLAM

AI-augmented SLAM integrates AI, particularly deep learning, into traditional SLAM pipelines to improve feature extraction, loop closure detection, and robustness in dynamic environments. Traditional SLAM methods rely on fixed algorithms for detecting features and aligning them, which can struggle in low-visibility or rapidly changing conditions. Deep learning models address these limitations by learning complex patterns in data. For example, ORB-SLAM2, a widely used SLAM system, was enhanced by integrating deep neural networks for key point detection and semantic understanding. This integration reduced localization errors in cluttered environments by better handling ambiguous or dynamic features [222]. Similarly, Tang et al. [223] proposed LCDNet, a deep learning-based loop closure detection method for LiDAR-based SLAM. Their method significantly improved performance in highly dynamic settings, such as urban traffic, by learning discriminative features that distinguish between static and moving objects.

### 4.5.2. Multi-Modal SLAM

Multi-modal SLAM combines data from multiple sensor types, such as LiDAR, cameras, and inertial measurement units (IMUs), to enhance robustness and scalability. Each sensor type has unique strengths: LiDAR provides accurate depth information, cameras offer high-resolution visuals, and IMUs deliver motion data. By fusing these modalities, multi-modal SLAM systems address limitations of individual sensors, such as LiDAR's sensitivity to rain or cameras' poor performance in low light. Campos et al. [224] introduced ORB-SLAM3, which integrates visual-inertial and multi-map SLAM capabilities to improve performance in challenging environments, such as tunnels or poorly lit areas. Leutenegger et al. [225] developed Semi-Direct Visual-Inertial Odometry (SVO), combining visual and inertial data to achieve accurate localization even in GPS-denied environments. These methods enable robust and scalable SLAM in diverse scenarios. AI-augmented and multi-modal SLAM approaches have advanced mapping capabilities. However, ensuring computational efficiency and robustness in real-world conditions requires further innovation. Balancing the complexity of AI models with real-time processing constraints remains a critical area of research.

### 4.6. Large-Scale Map Updates Challenge

AVs rely on high-definition (HD) maps for navigation, which provide detailed representations of roads, traffic signs, lane boundaries, and other critical information. However, maintaining up-to-date maps in dynamic environments, such as urban areas with frequent construction or temporary obstacles, is a logistical and computational challenge. Attempted solutions include strategies based on Crowdsourced Mapping and Edge Computing.

### 4.6.1. Crowdsourced Mapping

Crowdsourced mapping involves collecting and aggregating data from a fleet of PAVs to dynamically update HD maps. As PAVs traverse various routes, their onboard sensors—such as cameras, LiDAR, and GPS—capture real-time environmental data. This information is then transmitted to a central server, where it is processed to detect changes in the environment, such as new road features or alterations due to construction. By leveraging the collective data from multiple vehicles, crowdsourced mapping enables rapid and scalable updates to HD maps, ensuring they reflect the current state of the road network. For instance, Dabeer et al. [226] presented an end-to-end system that utilizes consumer-grade components to crowdsource precise 3D maps with semantically meaningful landmarks, such as traffic signs and lane markings. Their approach demonstrated that a fleet of vehicles equipped with standard sensors could collaboratively generate and update HD maps efficiently, reducing reliance on specialized mapping vehicles. Similarly, Liu et al. [227] proposed LiveMap, a real-time dynamic mapping system that aggregates perception data from connected vehicles within an edge computing framework. LiveMap processes data from multiple vehicles to create an up-to-date representation of the driving environment, enhancing the AV ability to navigate dynamic and complex scenarios.

### 4.6.2. Edge Computing

Edge computing involves processing data closer to its source—such as on the vehicle itself or nearby edge servers—rather than relying solely on centralized cloud servers. This approach reduces latency, conserves bandwidth, and enables real-time data processing, which is crucial for the timely updating of HD maps in rapidly changing environments. By decentralizing computations, edge computing allows PAVs to receive and contribute to map updates more efficiently, enhancing their responsiveness to environmental changes. For example, Zhang et al. [228] introduced EdgeMap, a crowdsourcing-based edge computing framework designed for real-time HD map updates. EdgeMap processes data locally on edge servers, minimizing the bandwidth required for cloud communication and ensuring low-latency updates. This system effectively balances computational demands across the network, allowing vehicles to access accurate maps without overburdening central infrastructure. Additionally, Liu et al. [229] developed LiveMap, a real-time dynamic mapping system that operates within an automotive edge computing network. LiveMap detects, matches, and tracks objects on the road by processing data from connected vehicles, thereby enhancing the AV situational awareness and ability to navigate complex environments.

Both crowdsourced mapping and edge computing offer promising solutions to the challenges of maintaining up-to-date HD maps for PAVs. Crowdsourced mapping leverages the collective data from multiple vehicles to provide comprehensive and timely updates, enhancing the scalability and accuracy of map maintenance. However, it requires robust data aggregation methods and mechanisms to ensure data quality and security. Edge computing complements this approach by enabling real-time data processing closer to the data source, reducing latency and bandwidth usage. This decentralization allows for more responsive map updates, which is critical in dynamic environments.

### 4.7. Conclusions and the Way Forward

Integrating crowdsourced mapping and edge computing can potentially provide a more robust framework for HD map maintenance. As documented in Section 6, implementing edge computing requires substantial investment in infrastructure and effective management of distributed resources. By combining the scalability of crowdsourced data collection with the responsiveness of edge computing, PAVs can achieve a more accurate and up-to-date understanding of their operating environments, thereby enhancing navigation safety and efficiency. As documented in Section 7, future research should focus on developing standardized protocols for data sharing and distributed processing, as well as addressing challenges related to data privacy, security, and the interoperability of systems across different vehicle manufacturers and service providers.

## 5. Common Sense Reasoning

### 5.1. Introduction and Methodology

Despite the significant improvements of sensors and machine learning applications, perceptual systems for autonomous driving still fall short of human drivers' capabilities. Besides the inherent difficulty to detect and recognize objects and environmental features, human-like driving also presuppose the capability to respond appropriately to the presence of these objects and features. This implies an understanding of the interrelations among objects or objects' properties; the intentions and proximal aims manifested by agents; the practical significance of the scenario at hand. This amalgamation of seemingly mundane yet intricately interconnected concepts, including the nuanced understanding of our environment and the mental states of other interacting agents, has posed a longstanding challenge for AI since its inception.

Researchers in AI usually refer to *commonsense* [231] to indicate the comprehensive reservoir of practical knowledge that artificial systems need to successfully navigate complex and open-ended real-life scenarios characterized by a great number of massively interconnected variables [232]. Commonsense has continued to be one of the thorns in the side of AI [233, 234], until the advent of LLMs [235]. Already by 2021 they achieved the same success of humans on the Winograd Schema Challenge [236], since then the standard test of commonsense unreachable by AI, prompting soon for much more demanding tests [237].

In light of this new scenario, this section focuses exclusively on language models—specifically, Multimodal Language Models (MLMs) with visual capabilities—as they are currently the only technologies capable of providing a degree of common sense. This is a rapidly expanding yet relatively nascent line of research. A search on Google Scholar for the terms "multimodal large language models" and "autonomous driving" yielded 60 hits up to 2023, which increased to ~500 by October 2024. Interestingly, a comparison with the number of hits for the term "multimodal large language models" alone returned ~4000 hits. This indicates that approximately 10% of MLM research is primarily applied to autonomous driving. The following section briefly characterizes the type of tools represented by MLMs, providing some technical details, and then addresses the areas where common sense would be most beneficial in autonomous driving systems, particularly in corner cases.

### 5.2. The Grounding Problem

From a theoretical standpoint, the introduction of MLMs offers a solution to the long-standing problem in AI known as the *Grounding Problem*, which pertains to how to create a connection between the internal representations of an AI system and the real-world entities [238]. Traditionally, this issue was referred to as the *Symbol Grounding Problem*, due to the symbolic nature of conceptual representations in classical AI. In the context of LLMs, it has been rebranded as the *Vector Grounding Problem* [239]. Equipping a system with the ability to perceive the external world necessitates creating the most effective connections with real-world objects, thereby addressing the grounding problem, whether symbolic or vectorial. However, this path presents significant challenges as, until a few years ago, effective vision systems were primarily based on deep convolutional neural networks (DCNNs), including those used for autonomous driving, whereas LLMs are built on the Transformer architecture [240].

A significant first step was realized with the Vision Transformer (ViT) [241], which adapted the attention mechanism of the Transformer from language to vision for the first time. In ViT, instead of operating on word embedding vectors, the model processes 16x16 patches of the input image. While ViT does not handle language processing and serves primarily as an alternative to established DCNNs, its introduction laid the groundwork for integrating the Transformer architecture with images. Notably, ViT has become a crucial component in facilitating true integration between language and vision, as exemplified by CLIP (Contrastive Language-Image Pre-training) [242]. In CLIP, contrastive learning ensures that outputs are similar when derived from the same input entity and dissimilar when derived from different entities. This involves separate encoders for the text describing an image and the image itself, generating latent vectors of equal length. Initially employed for tasks such as image captioning, simple question answering, and selecting images based on textual descriptions, CLIP's significance extends beyond these applications. It represents an effective approach to progressively integrating linguistic and visual domains. Variants of CLIP, such as ALIGN (A Large-scale Image and Noisy-text embedding) [243], have further advanced this integration.

Fig. 5. Representation of areas with more complex or underdeveloped road networks

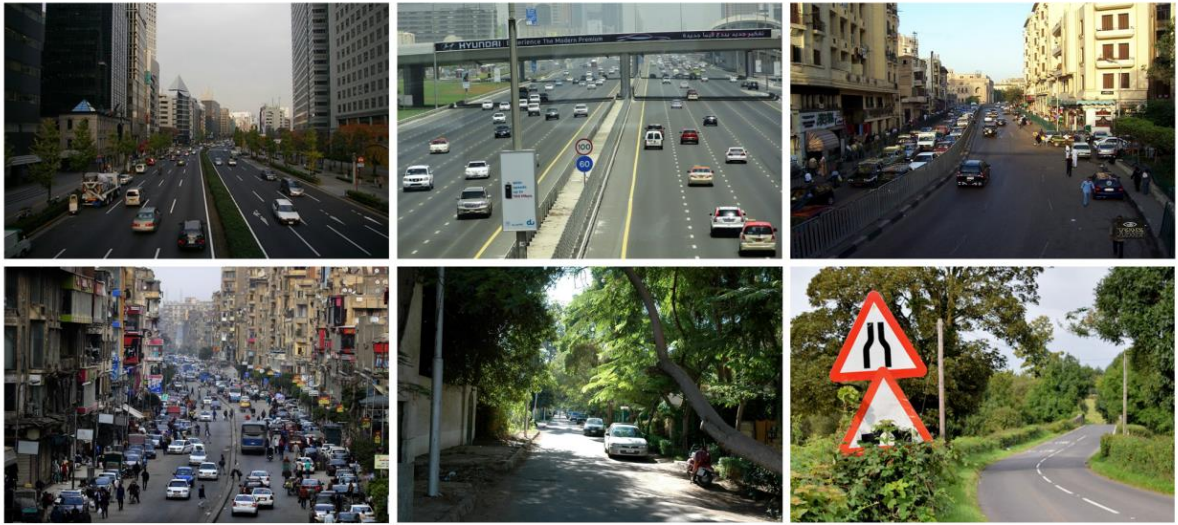

## 5.3. The Semantic Integration Problem

The need to simultaneously train the visual and linguistic components presents a challenge, as it often requires sacrificing the robust inferential semantic network characteristic of advanced LLMs, which is typically built upon extensive text corpora. To preserve this valuable knowledge base, various strategies have been explored to minimize the impact on the synaptic weights of the linguistic module when exposed to visual stimuli. For instance, in [244], the linguistic encoder remains unaffected, with only the gradients in the attention layer used for backpropagation of errors in the visual encoder. Alternatively, the LLaMAAdapter V2 model [245] takes a middle-ground approach, where only selected Transformer layers within the linguistic module are influenced by visual exposure during training.

LLaMA-Adapter V2, alongside GPT-3.5, is central to *Drive Like A Human*, one of the first partial adoptions of MLM in autonomous driving [246]. Prior to this, LLMs had already been tested in the driving loop, as seen with *GPTDriver* [247], which required a separate system, called *Unified Autonomous Driving* (UniAD) [248], capable of interpreting the scene due to the model's inherent lack of perceptual abilities. This setup involved a detailed translation process that converted all relevant information into a linguistic format, including salient objects, their spatial coordinates, and velocities. *Drive Like A Human* marks a step forward, operating in a closed loop that achieves full integration—not in the perceptual realm, but in the domain of control. It utilizes the ReAct system [249], which introduces a general prompting paradigm for combining reasoning and action with LLMs to tackle diverse decision-making tasks. Through the synergy of LLaMA-Adapter V2, GPT-3.5, and ReAct, *Drive Like A Human* demonstrates its ability to leverage common sense reasoning in specific scenarios. For instance, when presented with an image of a pickup truck carrying several traffic cones in its bed, the system does not mistakenly interpret the presence of the traffic cones as a hazard. The purpose of this pioneering work was explicitly to promote a new research direction, and it has indeed done so.

## 5.4. Towards the Development of a Research Agenda

In early 2024, the introduction of GPT-4V marked a significant milestone—a comprehensive MLM capable of processing linguistic inputs, including conversational interactions, alongside multiple images [250]. Shortly after its release, open-source alternatives became available, such as LLaVa (*Large Language-andVision Assistant*), which uses Vicuna for the language component and CLIP for the vision component [251], and InternVL, based on ViT and Llama-2 [252]. It was at this point that the avalanche effect in research proliferation occurred, as mentioned at the beginning of the section, with a jump from 85 works in 2023 to 573 as of September 2024. Notably, more than half of these publications are surveys, reviews, and position papers. Among the most recent and comprehensive surveys, we recommend [253, 254, 255]. Of those focusing on actual developments, only a few of the most significant are briefly listed here.

The DME-Driver (*Decision-Maker Executor-Driver*) system [256] is based on GPT-4V and tested on a dataset created by the authors, HBD (Human-Driver Behavior and Decision-Making), which integrates human driver behavior logic with detailed environmental perception. Both the dataset and the system's testing environment are based in the CARLA simulator [257]. Despite its name, the system *DriveGPT4* [258] is not based on GPT-4; it is a smaller MLM trained directly using 56K samples from the BDDX (*Berkeley DeepDrive eXplanation*) dataset [259] and an additional 223K samples of general multimedia instruction-following examples. Similarly, *DriVLMe* [260] relies on the BDD-X dataset for training Vicuna7B, but with additional training from what the authors call "embodied" and "social" experiences. The former refers to a collection of ego-car data in the CARLA simulator, while the latter involves real human dialogues during driving.

The work in [261] aims to adopt MLMs while keeping computational demands manageable, using video token sparsification and leveraging InternVL [252]. Two other systems, *AgentsCoMerge* [262] and *LMMCoDrive* [263], focus on vehicle communication. The former specializes in collaborative decision-making for ramp merging, while the latter is a more general MLM-based approach to cooperative motion planning for fleets of vehicles. In [34], a dataset called DrivingContexts is created, featuring a classification of 24 driving and environmental contexts, along with 2.6 billion context-query pairs relevant to driving. The *ContextVLM* system, based on LLAVa, classifies the current context accordingly. Thus, over the past year, there has been a progressive consolidation of the research agenda regarding the use of MLMs in autonomous driving, with the establishment of specific benchmarks and cross-comparisons among various proposed solutions. The next subsection addresses some particular cases.

### 5.5. Corner Cases

In the context of driving, the absence of a comprehensive framework rooted in common sense leaves autonomous driving systems ill-equipped, particularly when confronting what are often termed *cornercase scenarios*, uncommon situations that differ from the more limited reality captured by the datasets used to train vision systems. The conventional approach to addressing these corner cases—augmenting datasets to encompass a wider range of situations—yields incremental improvements. However, it falls short of overcoming the fundamental challenge of navigating an open world akin to the dynamic environments encountered on roadways. A striking example of a corner case can be found in one of the early exploratory studies on using GPT-4V in autonomous driving [265], which employed real-world video clips. In one of these clips, a small airplane performs an emergency maneuver on the highway, ending up in the outer lane with a wing extending into the driving lane. This scenario would pose a significant challenge for traditional visual recognition systems. Yet, GPT-4V accurately identifies the airplane and provides appropriate driving instructions to safely navigate around the obstacle.

The first step toward moving beyond anecdotal cases like the one described above is the creation of corner-case benchmarks. The initial benchmark, CODA [266], contains about 1,500 videos with cases grouped into 30 categories. From this, CODA-LM [267] was later derived, tailored for testing MLMs by supplementing each case in CODA with textual annotations concerning both the perception of the event and driving suggestions. In September 2024, as part of the European Conference on Computer Vision, a challenge based on CODA-LM was organized. The winning system [268] uses a twostage approach: first, LLaVA is fine-tuned to generate a coarse textual description and driving suggestion, followed by GPT-4 refining the output. This system achieved an accuracy of 83% for corner-case perception and 74% for driving suggestions compared to CODA-LM annotations. Meanwhile, the system in [269] is based on the current most advanced MLM from OpenAI, GPT-4o. Additionally, the LiteViLA (*Lightweight*

*Vision-Language model*) system [270] aims to reduce computational demand, leveraging the TinyLLaVA MLM [271], with only a limited drop in performance to 66% in driving suggestions.

### 5.6. Safety, Reliability, and Feasibility Challenges

The need for safety in autonomous driving imposes particularly stringent accuracy requirements, and it remains an open question whether MLMs can meet these standards. While their integration as a source of common sense offers the potential to enhance safety, the system becomes impractical if the accuracy is insufficient to substantially reduce false positives. From the perspective of on-board deployment, the same precautions applied to other deep learning modules in autonomous vehicle perception, such as protections against adversarial attacks, should be considered. These attacks involve minor disturbances in visual signals—such as altered road signs—that can induce significant errors in the visual recognition system [272, 273]. Although there have been no documented instances of such attacks being executed, they remain a potential vulnerability. Some studies [274, 275] suggest that MLMs, rather than introducing new vulnerabilities, could serve as an additional layer of defense. Nonetheless, this area requires further investigation.

### 5.7. Supporting Vulnerable Road Users

Given the critical importance of safety, autonomous driving systems pay special attention to vulnerable road users (VRUs), particularly pedestrians and cyclists [276]. Pedestrians display less predictable behaviors compared to other road users, whose movements more closely align with standard traffic patterns and are only partially influenced by road features. As a result, interactions between vehicles and pedestrians present a significant challenge for autonomous driving systems [277, 278]. It is, therefore, unsurprising that researchers are exploring how language models might address this challenge. In a study leveraging GPT-4V [279], driving scene frames are fed directly into the MLM. This primarily exploratory study involves designing prompt templates that end with questions like, "Is the pedestrian crossing in front of our car?" or "What is the pedestrian's action (standing/walking)?" These templates draw examples from standard pedestrian behavior datasets such as JAAD, WiDEVIEW, and PIE. Even in complex and ambiguous scenarios, GPT-4V demonstrates an ability to offer plausible interpretations of pedestrian behavior, thereby providing optimal suggestions for driving actions.

### 5.8. In-between Narrow and General AI: a Theoretical Challenge

Leveraging MLMs in autonomous driving systems presents considerable theoretical challenges, particularly when it comes to accurately capturing the nuanced patterns of human cognition. Efforts to emulate human drivers' cognitive abilities have historically focused on replicating individual functions one at a time [280]. However, achieving commonsense reasoning requires MLMs to simulate a broader range of cognitive processes that humans engage in during driving. Specializing an MLM for driving while retaining its broader commonsense capabilities is not straightforward. The challenge aligns with the classic tension between *Artificial General Intelligence* (AGI) and Narrow AI [281]. AGI aims for intelligent behavior across any context, while Narrow AI focuses on specific applications. While common sense is central to AGI, autonomous driving falls within the domain of Narrow AI. Among the research efforts discussed, some lean heavily toward Narrow AI [258], with concerns that a model specialized solely for driving tasks might fail when confronted with unexpected situations. In contrast, other approaches, such as those in [255], aim to infuse domain-specific expertise into MLMs while preserving their general adaptability.

A key advantage of Narrow AI is its lower computational demand, a critical benefit for in-vehicle applications. On the AGI side, however, current high-performing MLMs with advanced commonsense capabilities are often unsuitable for real-time use due to their extensive inference times. There is hope in ongoing efforts to create language models that balance high performance with reduced computational requirements [282, 271], as seen in the LiteViLA system [270] and techniques like video token sparsification [261].

In summary, while the commonsense reasoning that MLMs could bring to autonomous driving systems has the potential to greatly improve their ability to handle a wide range of scenarios, including challenging corner cases, their widespread integration into vehicular systems remains a longer-term objective.

# 6. Road Infrastructure

## 6.1. Introduction and Methodology

Despite significant advancements in the underlying technologies, the widespread deployment of advanced autonomous driving systems has been hindered by a multitude of challenges, many of which are inextricably linked to the state of our current road infrastructure, which encompasses elements such as road geometry, cross-section design, pavement quality, markings, signage, intersections, and traffic control systems. To address these challenges, a systematic review of top-tier research articles from the past decade has been conducted to investigate the compatibility issues and integration solutions between current road infrastructure and Level 4 and Level 5 autonomous vehicles. This review extracts articles from reputable databases like Scopus and Web of Science (WoS) to understand how key infrastructure components—such as road geometry, pavement conditions, signage, and traffic control systems—affect the performance of advanced autonomous driving technologies. The literature review presented in this section uses a search strategy with keywords like "autonomous vehicles," "Level 4," "Level 5," "road infrastructure," "compatibility," "adaptations," and "futureproofing." It focuses on articles that examine how existing infrastructure affects the performance of PAVs and suggest improvements for better integration.

Road infrastructure plays a critical role in enabling the safe and efficient operation of PAVs. These physical and digital components act as the foundational framework upon which PAVs must navigate, perceive their surroundings, and make informed decisions. However, the existing infrastructure, primarily designed with human drivers in mind, may not be fully compatible with the unique requirements and capabilities of Level 4 and Level 5 PAVs [283].

One of the primary challenges is the inconsistency and unpredictability of current road conditions, which humans are very good at handling. Faded or obstructed markings, non-standard signage, and temporary construction zones can confuse the perception systems of PAVs, hindering their ability to accurately interpret the road environment [284]. Furthermore, unexpected events such as traffic accidents, road closures, or the presence of human-driven vehicles can introduce dynamic and unpredictable situations that autonomous systems may struggle to anticipate and respond to appropriately [285].

Moreover, the transition to higher levels of automation necessitates a re-evaluation of existing design standards and guidelines. Many of these standards are based on human driver characteristics, such as perception-reaction times and eye heights, which may no longer be relevant in the context of PAVs [286], [287]. As a result, a comprehensive review and potential revision of design philosophies, specifications, and guidelines are required to accommodate the unique operational requirements of Level 4 and Level 5 PAVs [288], [289]. One of the most challenging issues is how an autonomous vehicle could adapt itself to the dynamics of road regulation and traffic conditions. Such a situation can happen anytime and anywhere because

- Road regulations (right of road) may vary with countries, states, regions or even segments of a road (see Figure XXX for an example and Section 11) [290].
- Owners may not be aware of changing traffic rules for old vehicles [291].
- Emergency situations can occur any time [292].
- Road infrastructure may not be available in all segments of a road (see Figure 1 for another example)[293].

In the future, when all cars are autonomous, the situation of road traffic can be dramatically different from what we have now and, therefore, will require different methods and infrastructures for road management and traffic control [294, 295].

In the next section, we will see that with the new technologies for vehicle-based communication and intelligent traffic control, traditional vision-based traffic control facilities, such as traffic lights, roundabouts, and stop signs, are likely to be replaced by less visible but more efficient and more effective algorithmic controlled road facilities [313]. This section, however, delves uniquely with road infrastructures and how they are required to implement higher levels of vehicle automation. By examining the current state of physical and digital infrastructure elements, identifying potential challenges and limitations, and exploring the necessary adaptations and upgrades, we aim to shed light on the critical factors impeding the widespread adoption of Level 4 and Level 5 PAVs. Through a comprehensive analysis of existing research literature, this section seeks to provide insights into the infrastructure-related barriers and propose strategies to overcome them, paving the way for a future where fully autonomous vehicles can seamlessly navigate our transportation networks.

| Geometric Element | Potential Problem with Current Standards | Suggested Considerations /Adaptations |
|---|---|---|
| **Sight Distances** | Designed for human perception-reaction time, which is longer than PAVs | Shorter sight distances may be feasible due to PAVs' quicker reactions [286], [289], [293] |
| **Horizontal Curves** | Not all horizontal curves accommodate PAVs' predictable path following | Incorporate AV kinematic models for optimized curve design [289], [293] |
| **Vertical Curves** | Crest vertical curves can limit sensors' line of sight | Design revisions for consistent visibility and speed maintenance [286, [293] |
| **Lane Widths** | Wider lanes cater to human driver error margins | Narrower lanes may suffice for PAVs with precise locational capabilities [289] |
| **Intersection Design** | Complicated designs may result in suboptimal path prediction for PAVs | Simplification or inclusion of AV-guidance signals could improve flow [289] |

Table IV: Various Geometries and Their Respective Problems for Autonomous Vehicles

## 6.2. The Inconsistency and Unclarity Challenge: Road Markings, Signage, and Traffic Control Systems

Proper delineation, consistent standards, and maintenance of road markings, traffic signs, and control signals are critical to the current operation of PAVs [294]. These physical elements serve as crucial visual cues that allow PAVs to navigate, interpret traffic rules, and make informed decisions [284]. However, the presence of faded or missing lane markings, such as those on heavily weathered highways, can lead to erratic lane-keeping behavior or unintended lane departures by PAVs [284], [294]. Obstructed or non-standard signs, like a yield sign obscured by vegetation or temporary construction zone markings deviating from standard practices, can confuse the perception systems of PAVs, hindering their deployment [290]. Therefore, the readiness and adoption of PAVs can vary significantly between regions based on the condition of their road infrastructure [295]. Well-maintained and clearly defined road markings and signage in regions like Abu Dhabi may facilitate faster adaptation of AV technology, enabling seamless navigation and adherence to traffic rules. In contrast, areas with more complex or underdeveloped road networks, characterized by faded markings, non-standard signage, and intricate road layouts, may require extensive infrastructure upgrades before widespread AV deployment can be achieved [296].

## 6.3. The Unpredictability Challenge: Poor Road Conditions and Messy Human-Vehicle Interactions

Level 4 and Level 5 PAVs need to be able to address dynamic and unpredictable road conditions seamlessly, which is a significant challenge with the current infrastructure. Unexpected events such as road closures due to construction or detours due to accidents can introduce unforeseen situations that PAVs may struggle to navigate. For instance, an AV encountering a sudden detour with minimal signage may fail to reroute effectively [295]. Similarly, the presence of human drivers can introduce unpredictability, like a driver waving another car through at an uncontrolled intersection, a subtle human cue that PAVs find difficult to interpret [297]. These complex driving scenarios often rely on subtle human cues, gestures, and unspoken agreements that are difficult to replicate in autonomous systems. For example, in a merge scenario, human drivers may use eye contact or hand gestures to negotiate the right of way, which current AV systems cannot comprehend reliably. Improving road infrastructure by implementing strategic approaches that facilitate safer and more efficient interactions between human-driven and autonomous vehicles, such as dedicated AV lanes or intelligent traffic management systems, is crucial for widespread AV deployment, especially in complex traffic situations like roundabouts or merging lanes [298].

## 6.4. The Digital Infrastructure Fragility Challenge: Lack of Robust Connectivity

Higher levels of AV require reliable and low-latency connectivity for high-definition mapping, real-time traffic data, and communication with other vehicles and infrastructure, emphasizing the need for real-time information exchange and integration of heterogeneous technologies for achieving higher levels of autonomous driving [299]. For instance, an AV without reliable connectivity may struggle to receive up-to-date high-definition maps or traffic data, leading to navigation issues or delayed responses to dynamic traffic conditions. The current digital infrastructure, including cellular network coverage, fiber-optic cable networks, and dedicated short-range communication (DSRC) infrastructure for vehicle-to-infrastructure (V2I) communication, may need to be improved in many areas to support the data demands of Level 4 and Level 5 PAVs [300]. For example, in areas with limited cellular coverage or outdated DSRC infrastructure, PAVs may face challenges in communicating with traffic signals or receiving crucial information from roadside units [301].

This aligns with the abstract's focus on addressing the challenges of technological augmentation of road infrastructure to support intelligent transport system (ITS) services and the need for enhancements in CCAM (Connected Cooperative and Automated Mobility) infrastructure imperative for intelligent road systems and decision-making of PAVs. Additionally, real-time data from sensors, IoT devices, and other sources are crucial for PAVs to navigate and make decisions effectively, echoing the abstract's emphasis on integration of various disciplines and technologies, such as sensors, communication, computation, and AI, to achieve higher levels of AV [302], [303]. Without a robust and widespread digital infrastructure in place, addressing the integration of these heterogeneous technologies and enhancing CCAM infrastructure, the deployment of higher levels PAVs will be restricted [302]. For instance, in areas with limited connectivity and outdated infrastructure, PAVs may not be able to leverage real-time sensor data or communicate effectively with other vehicles and road systems, hindering their ability to operate at higher levels of autonomy [283].

### 6.5. The Road Design Challenge: Re-Evaluation of Standards and Geometric Elements

Current road design standards and geometric elements, such as sight distances, horizontal and vertical curves, are largely based on human driver characteristics. However, with PAVs' enhanced perception abilities through sensors, shorter reaction times, and different operational requirements, a re-evaluation and potential revision of design philosophies, specifications, and guidelines will be necessary [286], [289], [293]. For instance, the required sight distance for a horizontal curve, which is determined based on human perception-reaction times, may need to be reduced for PAVs due to their faster reaction capabilities as depicted in Table IV. Standard reaction times are generally assumed to be around 2.5 seconds for perception and reaction under good conditions. This includes time for the driver to perceive a situation and initiate a response [287]. AV systems, like those from Tesla, can process information and react within milliseconds. While specific numbers vary depending on the situation, PAVs typically have significantly shorter reaction times due to their advanced sensor suites and processing capabilities [292].

In another scenario, the design of lane widths and shoulder widths, which are currently based on human driver behavior and vehicle dimensions, may require modifications to cater to the precise maneuvering capabilities and potentially smaller form factors of PAVs [304], [305]. By re-evaluating and revising these design standards and geometric elements, the infrastructure can be better optimized for the integration of PAVs, ensuring safer and more efficient operations as their adoption increases on public roads [306].

### 6.5.1. Modifications to Road Cross-Section Elements and Pavement Design

Current road cross-sectional elements, such as lane widths, shoulders, medians, and barriers, were originally designed with human drivers in mind, accounting for human perception and reaction times, as well as variability in steering accuracy [307], [308]. However, these standards may not fully align with the precise positioning capabilities of PAVs, which are less prone to lane wandering and can maintain consistent paths [309]. This difference creates a need to adjust certain elements to accommodate AV-specific requirements. For instance, in areas with dedicated AV lanes, narrower lane widths could be feasible due to PAVs' ability to maintain precise positioning. Additionally, in scenarios like vehicle platooning, where multiple PAVs travel closely together, the lanes could be designed for optimal alignment and spacing.

At the same time, wider shoulders or designated stopping areas may be necessary to allow PAVs to pull over safely when they encounter situations beyond their operational design domain, such as unexpected obstacles, severe weather, or construction zones. These modifications would help ensure that road infrastructure supports the unique operational needs of PAVs. Additionally, PAVs' reduced wheel wander and precise positioning could accelerate pavement rutting and deterioration, necessitating optimized pavement designs based on variables like speed limits, lane widths, and loading patterns [310], [311]. For example, in high-speed AV-only lanes, where vehicles are expected to maintain tight formations and precise positioning, pavement designs may need to be reinforced to withstand the concentrated loads and minimize rutting. Skid resistance requirements may also need re-evaluation to ensure safe AV operation [297]. In scenarios where PAVs need to perform emergency maneuvers or braking, higher skid resistance may be required to maintain control and avoid collisions, especially in adverse weather conditions [284], [288].

### 6.5.2. Other Infrastructural Elements

Intersections, junctions, and parking facilities pose considerable challenges for PAVs due to their complexity and the need for real-time interpretation of dynamic traffic situations [312]. For instance, intersections often become bottlenecks, as PAVs must navigate intricate scenarios involving multiple vehicles, pedestrians, and signals. This complexity creates a need to redesign these areas to better accommodate PAVs and enable smoother traffic flow. Possible solutions include implementing dedicated AV lanes or enhancing infrastructure connectivity to facilitate communication between PAVs, traffic signals, and surrounding vehicles [313]. Additionally, parking facilities may need upgrades to support autonomous valet parking, allowing PAVs to independently drop off and pick up passengers without requiring human input [313]. These adaptations would help integrate PAVs more effectively into urban environments.

### 6.6. Conclusions and the Way Forward

In the future, all critical infrastructure elements, including bridges, tunnels [314], drainage systems, road lighting, roadside equipment, and facilities for vulnerable road users will have to be assessed thoroughly and upgraded to accommodate the needs of PAVs [315], [316]. Based on the data presented in Section 4 and the consideration offered in Section 7, bridges and tunnels will require enhanced connectivity and safety features to ensure that PAVs can navigate them reliably. These structures often have limited GPS and sensor functionality due to their design, which can hinder PAVs' ability to maintain positioning and detect obstacles accurately, aggravating the detection problems reviewed in Section 3: for example, roadside equipment and facilities for pedestrians and cyclists will have be designed or upgraded to ensure safe interactions with PAVs [318]. Also, drainage systems must be regularly monitored to prevent water accumulation that could affect AV sensors and road lighting improvements will be necessary to ensure PAVs environment perception operate safely at night [317]. In Section 7 we will argue that a key intervention for improving navigation and safety for PAVs consists in installing dedicated PAV communication systems, such as vehicle-to-infrastructure (V2I) technology or enhanced lane markings and lighting, could [314].

As discussed in Section 11, transportation agencies and policymakers, in collaboration with the manufacturers, will be able to address these key aspects of road infrastructure only if they systematically coordinate to proactively plan and implement the necessary changes, upgrades, and standardizations. This coordinate planning will ensure a smooth transition to a future with widespread deployment of Level 4 and Level 5 autonomous vehicles while maintaining safety, efficiency, and accessibility for all road users.

## 7. Connected Autonomous Vehicles and Traffic Management

### 7.1. Introduction and Methodology

The advancement of PAVs technology is driven not only by innovation in AI and sensors but also by progress in telecommunication, mobile networking, and roadside infrastructure. These advancements enable intelligent traffic control and management systems. The new era of autonomous driving presents opportunities for both centralised and distributed traffic control systems, each with its own set of advantages and challenges. These challenges include various technical limitations and a trade-off between autonomous decision-making and system-wide optimization. With recent developments in communication technology, PAVs are now interconnected, which introduces new complexities in traffic management. For example, it involves a more complex version of the "Vehicle Routing problem" (VRP), a class of problems that require designing routes for a fleet of vehicles to serve customers. The VRP for PAVs is a particularly complex subclass of problems involving multiple autonomous vehicles entering and exiting a network and designing the best routes for them to bring the passengers to their destination. Solving this problem means optimizing the traffic flow coordinating the movements of PAVs to maximize efficiency and safety for each vehicle involved. This problem can be addressed using a variety of computer science models, including static routing game models, evolutionary dynamics of repeated games models, queuing models, and online routing game models [346, 347, 348]. Note that VRP differs from path planning (Sections 3 and 4) because the former involves finding the path, from origin to destination in the network, that is optimal for a flux of coordinated vehicles, while the latter is about calculating the best trajectory for a vehicle's movement in each particular manoeuvre.

As mentioned in Section 6, traditional traffic control systems (traffic lights, stop signs, and roundabouts) may soon become outdated, necessitating their replacement with smart roadside infrastructure. A related but distinct issue is that intelligent traffic control will be essential for the widespread deployment of PAVs. This new era of transportation calls for novel technologies and theories supporting smart roadside infrastructure and effective traffic control, both of which require robust systems for communication and coordination among vehicles. These solutions can generally be implemented in two ways: centralized control and distributed decision-making [327]. In a centralized system, a central entity manages traffic, potentially outperforming decentralized systems, where individual vehicles make autonomous decisions through automated negotiations with each other and with roadside infrastructure. Balancing these approaches presents additional challenges for government authorities and road users. In this section, we will briefly discuss on current development and future challenges with the technologies of Connected Autonomous Vehicles (CAVs).

### 7.2. Connected Autonomous Vehicles: the Adaptivity Challenge

CAVs technologies combine autonomous driving capabilities with advanced connectivity, offering substantial potential to improve road safety by reducing accidents through real-time data exchange and predictive algorithms. CAV communication can be categorized into three main types: Vehicle-to-Vehicle (V2V), Vehicle-to-Infrastructure (V2I), and Vehicle-to-Everything (V2X) [334, 335, 336]:

- *Vehicle-to-Vehicle (V2V)* – This communication allows vehicles to directly exchange information on speed, location, direction, and other critical data to prevent collisions, enhance safety, and improve traffic flow. V2V enables collision avoidance, real-time traffic information sharing, and cooperative adaptive cruise control.
- *Vehicle-to-Infrastructure (V2I)* – V2I communication connects vehicles with infrastructure elements such as traffic lights, road signs, and parking meters. This helps optimize traffic flow, reduce congestion, and provide timely information on road conditions and traffic signals. Applications include smart traffic signal systems, real-time parking information, toll collection, and hazard warnings.
- *Vehicle-to-Everything (V2X)* – V2X is an umbrella term that *includes* V2V and V2I, as well as other communication forms like Vehicle-to-Pedestrian (V2P) and Vehicle-to-Network (V2N). V2X aims to establish a fully connected transportation ecosystem where vehicles, pedestrians, network services, and other entities can interact seamlessly, enhancing safety and efficiency. Beyond V2V and V2I, V2X includes communication with mobile devices for pedestrian safety, network-based data processing, and interaction with emergency services and bicycles.

While V2V, V2I, and V2X communication technologies enable seamless connectivity among vehicles and infrastructure, fully harnessing the potential of CAVs to improve traffic efficiency and road safety remains challenging. Integrating these communication systems into autonomous driving is essential for smarter traffic management and roadside assistance. These efforts are vital for achieving widespread adoption and effective implementation of CAV technologies [319, 321, 323, 330, 331]. Additionally, the deployment of CAV technologies requires a robust regulatory and legal framework to address critical issues such as liability, data privacy, and cybersecurity (Section 7.6).

As we have mentioned earlier, the challenges to traffic management introduce by autonomous vehicles on roads largely depend on the scarce adaptivity of these systems to varying conditions. New theories for better understanding of the new era of transportation and new technologies for smart roadside infrastructures and intelligent traffic control are crucial for development and deployment of autonomous vehicles with the necessary adaptive qualities. As discussed in Section 6.1, adaptivity involves up-to-date knowledge of changing road regulation and real-time sensitivity to unexpected traffic conditions. Thanks to the constant flux of information they share with the network and with other vehicles, CAVs represent a new approach to meet these requirements and thus achieve smarter traffic control. In the following sections, we will discuss the related issues and challenges.

### 7.3. The Intelligent Intersection Management Challenge

The direct application for CAV technologies is smarter intersection control. Intersections are critical points in traffic control systems, often causing significant delays and accidents. Modern traffic signal systems have evolved from fixed timing signals to those that adjust based on various parameters, including time of day and traffic density. Common intersection traffic management strategies include Cybercars-2, Intersafe-2, Autonomous Intersection Management [329, 334], and the Intelligent and Cooperative Intersection Collision Avoidance System [350]. Vision-based systems utilize sensors to monitor traffic flow, but they typically manage individual intersections independently.

However, current road infrastructures and traffic control technologies are designed for human drivers, and even self-driving cars are trained to recognize human-oriented traffic signals, which may not be necessary for future CAV traffic management. New traffic management protocols have been proposed in the literature specifically for traffic with autonomous vehicles, such as first-come first-served (FCFS), virtual traffic light, virtual roundabout and priority-based protocols, and so on [337, 338, 339]. Beyond these protocols, more sophisticated intersection management systems are being developed that integrate V2X communication. This approach can significantly enhance traffic flow and safety by enabling coordinated manoeuvres and dynamic adjustment to dynamic traffic conditions [331]. Designing and developing traffic management protocols for fully autonomous vehicles is more challenging than anticipated, primarily due to the limited number of such vehicles currently on the roads. To address this issue, simulators are employed to generate data and enable comparisons with existing traffic management systems. Commonly used simulators include aimsun [487], AIM4 [488], SUMO [489], and CARLA [490].

### 7.4. The Traffic Control Challenge: Balancing Centralised and Decentralised Control

Optimising the efficiency of complex traffic systems ideally involves centralised control. However, implementing centralised control over human-driven vehicles is challenging, as enforcing real-time decisions on individual drivers is nearly impossible. With connected vehicles, centralised traffic control becomes feasible in certain contexts, such as at intersections, though it requires additional investments in communication and infrastructure [349, 350]. Alternatively, each connected and autonomous vehicle (CAV) can function as an intelligent agent, capable of making safe driving decisions, predicting traffic flow, and selecting optimal routes. In this setup, traffic involving CAVs can be modeled as a multi-agent system, allowing traffic analysis through equilibrium calculations [333].

Centralised control generally provides better efficiency in terms of minimizing traffic delays [320]. The challenge lies in determining when centralised control should be applied versus decentralized, agent-driven control. Roughgarden proposed addressing this question with the concept of the Price of Anarchy (PoA) [342]. The PoA is defined as the ratio of system cost under decentralised control to that under centralized control, serving as an index to identify which traffic situations allow for self-determined vehicle decisions and which require centralised control. When the PoA is low, the difference in system cost between decentralized and centralized control is minimal, allowing vehicles to make independent decisions. Conversely, when the PoA is high, self-determination becomes inefficient, and centralized control can achieve better traffic flow.

The PoA concept can be extended by incorporating real-time data and adaptive control strategies. By continuously monitoring traffic conditions and calculating the PoA, traffic management systems can dynamically switch between centralised and decentralised control modes. This adaptive approach enhances traffic flow, reduces congestion, and improves overall travel times.

### 7.5. Automated Negotiation Challenges and Potential Solutions

One of the most challenging issues for PAVs and AV traffic control is enabling PAVs to navigate roads with unknown traffic rules or situations they were never trained for. Currently, no PAVs can operate under such conditions autonomously. Developing PAVs capable of handling unknown environments requires automated negotiation between vehicles or between vehicles and roadside infrastructure. Automated negotiation is essential for agents in a multi-agent system to collaborate and reach mutually beneficial solutions [322]. This method uses computational approaches to facilitate decision-making and coordination in scenarios involving PAVs, traffic management, and logistics. Agents can propose solutions that are either accepted or rejected, with accepted proposals associated with utility values. Automated negotiation offers a promising solution, enabling PAVs to communicate, coordinate, and make collective real-time decisions. Beyond traffic control, automated negotiation is also applicable in logistics [324, 325, 326, 328].

However, current technology does not yet support PAVs traveling on completely unknown roads. New approaches are required. A similar problem has been explored by researchers in General Game Playing (GGP), which addresses how a computer player can engage in a game without knowing the rules until the game begins [340]. In GGP, any game is described in Game Description Language (GDL), allowing a general game player to interpret the rules and devise effective strategies autonomously. Extensions of GDL [341, 342, 343, 344, 345] have successfully been applied in various areas, including financial markets [341], auctions [343], and automated negotiation [344]. It is feasible to extend GDL further to describe road segments and traffic control protocols, supporting automated negotiation mechanisms for PAVs. Research on automated negotiation for PAVs can be divided into three main directions:

### 7.5.1. Description of Road Networks

One approach to modeling traffic information involves representing road segments as grids and expressing traffic flow through spatiotemporal relationships. By extending GDL to include these elements, we can enable PAVs to interpret diverse traffic situations. This spatiotemporal extension could also support the description of traffic management protocols in terms of negotiation mechanisms.

### 7.5.2. V2I Negotiation

With connected PAV technology, vehicles can negotiate travel permissions with roadside infrastructure when centralized or semi-centralized control is enforced. This is particularly useful in complex scenarios, such as multi-way intersections, roadwork zones, emergencies, or unusual weather conditions. A GDL-based description of road configurations and negotiation protocols could facilitate negotiation between PAVs and roadside infrastructure.

### 7.5.3. V2V Negotiation

Vehicle-to-vehicle (V2V) negotiation can be advantageous in unsignaled intersections or rural areas. In simpler traffic scenarios, V2V negotiation is straightforward to implement. PAVs could automatically generate negotiation mechanism descriptions from predefined templates and share these with nearby vehicles [349].

The GGP-based approach enables PAVs to interpret and engage in negotiation mechanisms without predefined scenario-specific facilities. When a vehicle approaches a road segment or intersection, it could automatically request a traffic management description from roadside infrastructure, if available. Alternatively, it could generate a traffic management description to initiate negotiation with other vehicles. A GDL parser and state-machine solver would enable an PAV to interpret any valid negotiation mechanism and devise its own strategy for traffic agreement negotiation.

### 7.6. Data Privacy and Cybersecurity

Cybersecurity vulnerabilities pose critical and multifaceted threats to the safe and reliable operation of these vehicles. Various cybersecurity challenges face CAVs, including attack vectors, vulnerability types, and emerging solutions. As smart vehicles increasingly leverage interconnectivity—through vehicle-to-vehicle (V2V), vehicle-to-infrastructure (V2I), and vehicle-to-everything (V2X) communications—the attack surface expands, creating both novel cybersecurity threats and amplifying existing ones [414]: CAVs rely heavily on a complex network of sensors (e.g., LiDAR, radar, cameras) and high-performance computing platforms to make safety-critical decisions [415]; in addition, communication with centralized data centers or roadside units (RSUs) for real-time traffic information, software updates, and maintenance tasks further complicates the cybersecurity landscape. Hence, safeguarding CAVs from malicious actors and sophisticated cyberattacks has become a priority for researchers because without robust cybersecurity frameworks for data confidentiality, integrity, and availability the technological promise of CAVs could be severely [416].

### 7.6.1. Attack Vectors and Vulnerabilities

Attacks on autonomous and connected vehicles can target hardware, software, or the communication channels linking these systems. Hardware-oriented attacks exploit vulnerabilities in electronic control units (ECUs), onboard diagnostic (OBD-II) ports, or external interfaces that lack rigorous authorization checks. A study by Zhang & colleagues [417] highlights the risk posed by side-channel attacks, where attackers monitor power usage or electromagnetic emissions to compromise cryptographic keys. On the software side, researchers have identified vulnerabilities in the operating systems, firmware, and various application-level modules that govern the vehicle's functionality [418]. These weaknesses can include poor coding practices, unpatched software libraries, and misconfigurations in safety-critical functions (e.g., brake or steering control). For instance, an often-cited study by Miller & Valasek [419] demonstrated how attackers could remotely seize control of a vehicle's steering and braking systems by exploiting the vehicle's telematics control unit. Meanwhile, research on advanced driver assistance systems reveals that deep learning frameworks used for object detection and sensor fusion can be susceptible to adversarial inputs. Attackers can craft perturbations in sensor data—sometimes imperceptible to the human eye — to induce misclassification in neural networks [420].

An equally significant concern is the communication network itself. The V2X ecosystem relies on DSRC, cellular networks, or emerging technologies such as 5G to transmit data. However, wireless protocols are inherently susceptible to eavesdropping, spoofing, Denial of Service (DoS) attacks, and data injection [421]. Some research points out that multi-hop communications, where data travels through intermediate nodes, further expand the attack surface, enabling attackers to intercept or manipulate data in transit [422]. Cyberattacks on CAVs can have immediate, severe consequences. In the realm of personal vehicles, compromised steering or braking systems pose direct risks to the safety of passengers, pedestrians, and other road users. For instance, a successful DoS attack could freeze critical sensor data or disrupt vehicle decision-making at high speed, leading to collisions and potentially catastrophic results [423].

The interplay between physical safety and cybersecurity becomes even more pressing in the commercial sector, particularly for autonomous trucks. A cyberattack on a fleet of trucks can produce large-scale disruptions to supply chains, cause severe road accidents, or inflict considerable economic damage [424]. Beyond physical harm, there are major concerns about the privacy of user data. Autonomous and connected vehicles generate and process large amounts of sensitive information, such as routes, destinations, user preferences, and even biometric data for driver authentication [433]. Such data can be exploited to track individuals, commit fraud, or execute identity theft if adequate encryption and data-protection measures are not in place. Additionally, as more ACVs integrate with smart city infrastructures, a breach in the transportation network could propagate to other critical systems, including emergency services, power grids, and telecommunication networks.

### 7.6.2. Key Security Challenges

One of the most commonly cited challenges in the literature is the resource-constrained environment of some automotive components, which limits the feasibility of strong cryptographic algorithms. Embedded systems in vehicles often operate under tight constraints regarding memory, processing power, and energy consumption [416]. Sophisticated encryption or frequent key rotations may exceed system capacity or negatively affect latency—both of which can degrade real-time vehicular performance. Another substantial challenge is the automotive supply chain's fragmented nature. Components originate from diverse manufacturers, and software often integrates proprietary and open-source code. According to Kaiwartya & colleagues [425], the lack of standardized security protocols and interoperability frameworks creates patchwork solutions that hackers can exploit.

Moreover, automotive systems are expected to have extended lifespans (ten to fifteen years or more), which complicates *over-the-air* updates and security patch management. There is also a skills gap in cybersecurity for the automotive sector. As vehicles transition from mechanical machines to interconnected computing platforms, a specialized workforce with expertise in both automotive engineering and cybersecurity is needed. Industry reports and scholarly investigations alike emphasize the urgent necessity to educate and train a new generation of professionals who can apply advanced cryptographic, machine learning, and systems engineering knowledge to secure ACVs [426].

### 7.6.3. Mitigation Strategies and Proposed Solutions

Peer-reviewed literature presents a broad range of solutions, from hardware-based defenses to sophisticated *intrusion detection systems* leveraging artificial intelligence. *Hardware security modules* integrated into the ECU can store cryptographic keys securely and enable hardware-based trust anchors, preventing unauthorized tampering [414]. On the communication front, secure key management schemes using lightweight cryptography, such as *elliptic-curve cryptography*, have been explored to achieve strong security without overburdening system resources [427]. Meanwhile, *machine learning* (ML) and *deep learning* (DL) techniques are increasingly proposed for real-time anomaly detection. Researchers highlight the promise of deep neural networks for identifying unusual behaviors in sensor readings or CAN (Controller Area Network) bus traffic [428]. Similarly, blockchain-based frameworks have been considered for ensuring trustworthy data exchange in V2X systems [429]. These decentralized ledgers allow for tamper-proof transaction records, authenticating data broadcasted across multiple vehicles and infrastructure nodes [429]. Moreover, the adoption of formal verification methods has garnered interest to rigorously validate the correctness of safety-critical software. While formal methods can be computationally intensive, some studies demonstrate their use in checking system-level properties, such as ensuring that messages or software tasks remain within trusted boundaries [430]. These strategies, combined with frequent security audits and multi-layer defense architectures, form the foundation of a holistic approach to protecting autonomous and connected vehicles.

As we will see in Section 11, security measures in ACVs extend beyond technical solutions. Policy and regulatory frameworks, such as the ISO/SAE 21434 "Road Vehicles—Cybersecurity Engineering" standard, outline requirements for vehicle manufacturers and suppliers to establish secure software development life cycles and risk management practices [431]. Additionally, institutions like the National Highway Traffic Safety Administration (NHTSA) and the European Union Agency for Cybersecurity (ENISA) have published guidelines recommending layered security designs, third-party testing, and mandatory reporting of cyber incidents. Standardization remains a work in progress, however. Given the different communication protocols, vehicle architectures, and regional regulations, the automotive industry faces challenges in converging on a unified set of cybersecurity standards. Still, there is consensus among stakeholders that government and industry collaboration is key to achieving a robust cybersecurity posture across brands and borders [432].

### 7.6.4. Future Directions in Cybersecurity

The body of scientific literature underscores that cybersecurity for autonomous and connected vehicles is inherently complex, driven by the interplay of hardware limitations, sophisticated software stacks, and multi-channel communications. Attack vectors can encompass everything from physical tampering with onboard components to adversarial manipulation of sensor inputs and malicious exploitation of wireless networks. The consequences, from direct threats to passenger safety to large-scale disruptions of commercial fleets, highlight the critical nature of securing ACVs against cyberattacks.

Emerging research provides a raft of potential mitigation strategies, including hardware security modules, advanced cryptographic protocols, AI-driven intrusion detection, and blockchain-based frameworks. Nevertheless, further work is essential, especially in standardizing security practices, ensuring the sustainability and longevity of security measures, and cultivating the necessary expertise in this interdisciplinary area. As regulatory bodies, manufacturers, and researchers converge on robust cybersecurity solutions, the ultimate goal remains the same: realizing the full promise of autonomous and connected vehicles while preserving public safety, privacy, and trust. Only through a concerted, collaborative effort can the transportation sector fully harness the transformative benefits of ACVs without succumbing to the cyber threats that currently loom over this promising technology.

### 7.7 Conclusions and the Way Forward

By facilitating real-time communication and coordination among PAVs, automated negotiation can reduce traffic bottlenecks, enhance responses to road hazards, promote energy-efficient driving, and ensure equitable road use. However, successful implementation requires overcoming technical challenges, such as developing robust negotiation algorithms that can handle complex, dynamic traffic scenarios and process large data volumes in real-time. Furthermore, public acceptance is crucial for widespread adoption of these technologies (see Section 9). Future research should prioritize refining coordination and negotiation algorithms and integrating them seamlessly into existing transportation infrastructures (as seen in Section 6). To build trust and acceptance among the public is equally important, as we will see in Sections 9: this endeavour involves not only technological advancements but also addressing socio-cultural, ethical, and regulatory concerns (as discussed in Sections 9-11) Continued exploration and development of these systems can help pave the way for safer, more efficient, and equitable transportation.

## 8. Human Factors and Interfaces

### 8.1. Introduction and Methodology

The road to PAVs is developing into an engineering reality. However, until fully autonomous vehicles are developed, they are expected to be limited in their capabilities, requiring humans to attend to unexpected traffic situations beyond the capabilities of PAVs (i.e., deteriorated weather conditions, roadworks, disorderly parking lots, etc.). Human factors, a multidisciplinary field focusing on understanding the interactions between humans and other system elements, plays an important role in the design of safe, user-centric and effective automated vehicle technology [351, 352]. The transition from human-driven to autonomous vehicles presents a complex array of human factors challenges, as the technology is not yet 100% reliable and safe[353, 354]. The integration of human factors knowledge is crucial when developing safety-critical systems, such as automated vehicles.

This Section provides a brief review of published literature on some key human factors issues and challenges associated with the transition from manually driven to autonomous driving, including ways to address these challenges: cognitive workload; situational awareness; the Human-Machine Interface (HMI) challenge. The review considered the Human Factors & Ergonomics literature of the last 10 years, including peer-reviewed research publications. Keywords like "cognitive workload", "situational awareness", "Human-Machine Interface" have been chosen to identify the most relevant advancements in the space of human factors and autonomous vehicles. Three databases were interrogated: 'PubMed', 'Web of Science', 'Google Scholar'.

### 8.2. The Cognitive Workload Challenge: Human Performance Issues

Cognitive load refers to the amount of mental effort required to process information and perform tasks and can impair performance and decision-making if it is too high or too low [356].

### 8.2.1. Cognitive Underload

High levels of automation can reduce the driver's cognitive load, potentially leading to complacency and decreased vigilance. If the workload is low during automated driving, the driver may experience passive fatigue or lack of direct control over the task [357]. Previous research focusing mainly on SAE Level 3 equipped vehicles has highlighted that increased vehicle automation is associated with reduced driver vigilance as shown by increased braking and steering reaction times in response to a sudden critical event [358]. Recent studies have also shown that, because of low workload in periods of automated driving, drivers engage in secondary tasks as opposed to monitoring and supervising autonomous driving [359]. These studies suggest that drivers can be more vulnerable to distractions and inattention during periods of driving automation, compromising their ability to suddenly regain control of the vehicle when needed [360].

### 8.2.2. Cognitive Overload

In contrast, the transition between automated and manual control, particularly in SAE Level 3 vehicles, can significantly increase cognitive load. Drivers need to quickly reorient themselves to the driving environment, when requested by the automation, and make immediate decisions, which can be challenging if they have been disengaged during automated driving. This sudden increase in cognitive load can lead to performance deficits [361]. In addition, the requirement to supervise automation can increase cognitive load [362].

### 8.2.3. The Interface Design Challenge

The design complexity of HMIs can either mitigate or exacerbate cognitive load. Generally, poorly designed interfaces that require significant mental effort to understand and operate can overwhelm drivers, while well-designed interfaces can help distribute cognitive load more effectively [363]. Developing adaptive interfaces that adjust the amount and type of information based on the driver's current cognitive state and situational context can help manage cognitive load. For example, the interface could provide more detailed information in complex situations and reduce information flow during routine operations [360]. Implementing real-time feedback mechanisms that alert drivers to changes in the driving environment or system status can help maintain situational awareness and manage cognitive load. Providing timely support and guidance can ensure that drivers are prepared to take back control of a vehicle when necessary [365]. Finally, providing comprehensive training and opportunities for users to become familiar with automated systems can help them manage cognitive load more effectively. Training programs should focus on helping users understand how to interact with the system and what to expect during automated and manual driving phases [366, 354].

### 8.3. The Situational Awareness Challenge

Situational awareness refers to the perception of environmental elements, comprehension of their meaning, and projection of their future status. High levels of situational awareness are crucial for making informed decisions and responding effectively to dynamic driving situations [367].

When operating PAVs, Kyriakidis et al. [369] suggested determining the individual capabilities of human drivers (i.e., level of situation awareness and reaction times, time to ensure safety while maintaining changing driving modes). It would be unrealistic to expect human drivers to constantly monitor the automation system when operating PAVs at Level 3. According to Reed [369], although it is feasible to deploy conditionally automated vehicles (i.e. SAE Level 3), the expectation that a human driver can remain alert and rapidly regain situation awareness following a request by the system is unrealistic.

In addition, the authors advised that it is important to define the minimum time requirements for human drivers to return in the control loop, for several driving scenarios. It would be important to determine the type and frequency of information that human drivers should be receiving to facilitate and maintain their situation awareness, primarily when they are not engaged in the driving task [369]. Merat and colleagues [369] suggested that, for the next 5-10 years, the research should be focusing on providing solutions for maintaining human drivers' situation awareness, when they are engaged in the driving task. Importantly, improvements in the design and performance of HMIs are required to establish the type and amount of information that drivers should receive to cope with any unexpected situation [370]. The most common issues associated with situational awareness in automated vehicles may include loss of situational awareness, appropriate attention allocation, mode confusion, and transition periods between automated and manual control.

**8.3.1. Reduced situational awareness** is a serious risk for automated systems, as drivers may become overly reliant on automation and disengage from the driving task. This can result in drivers being unaware of critical changes in the driving environment, which is particularly problematic in situations where manual intervention is required [367].

**8.3.2. Appropriate allocation of attention** is necessary to maintain an adequate level of situational awareness. Automated systems that demand too much or too little attention can disrupt this balance. Drivers may either become overloaded with information, or when cognitive resources are underutilized, both of which can impair their ability to maintain situational awareness [365].

**8.3.3. Mode confusion** occurs when drivers are uncertain about the current state of the automated system, such as whether the vehicle is in manual or automated mode; or in which automated mode it is in. This confusion can lead to incorrect assumptions about the vehicle's behavior and reduce situational awareness [368].

**8.3.4. Transitioning between automated and manual control** is a critical period for situational awareness. Drivers need to rapidly reacquire situational awareness, which can be challenging if they have been disengaged from the driving task during automation. This can lead to delays in recognizing and responding to hazards [361].

### 8.4. The Human-Machine Interface (HMI) Challenge: Design and Usability Issues

From the human factors perspective, the design of HMIs plays a significant role in ensuring effective interaction between drivers and automated systems [379]. Effective HMI design is crucial for ensuring that users can easily and intuitively control and monitor operations, maintain situational awareness, and respond to system alerts and notifications [362].

### 8.4.1. Lack of clarity and simplicity

Clear, intuitive displays and controls are essential for conveying critical information about the status of the vehicle automation and any required driver actions. Poorly designed interfaces can lead to confusion, errors, and delayed response times [362, 363]. The HMI must provide clear and unambiguous information about the vehicle's status, intended actions, and any required human interventions. Therefore, HMI design must prioritize user-centered principles to enhance usability and safety [30]. Poorly designed HMIs can increase cognitive load by overwhelming users with excessive information, not presenting information required or presenting information in a confusing manner. Complex interfaces can hinder user comprehension and decision-making, leading to errors and reduced trust in the technology [362]. For example, in recent years, the designers of smart cars' interfaces have transferred to touchscreens several basic functions that traditionally were entrusted to physical buttons and levers (e.g., signals and wipers). However, safety concerns were recently raised in relation to this trend, with experts emphasizing that screen-based controls significantly increase the risk of distraction and recommending the return of levers and physical buttons to reduce partially or completely the driver's reliance on screens [411].

### 8.4.2. Lack of consistency

Furthermore, lack of consistency across HMIs in different automated vehicle models or manufacturers can confuse users and increase the learning curve. Inconsistent placement of controls, varying visual styles, and different interaction methods can make it challenging for users to transfer their knowledge and skills between different automated vehicle systems [364]. In addition, transitioning between automated and manual driving modes can pose usability challenges. Users must quickly adapt to changes in interface behavior and understand when manual intervention is required. Poorly managed transitions can disrupt driving tasks and reduce overall system usability [377]. For partially automated vehicles (SAE Levels 1-3), issues here include which driving functions should be allocated, why, when and who is responsible for the allocation [372]. The appropriate allocation of function to human drivers and automation is also a fundamental issue in HMI design: humans and machines have different capabilities, strengths and limitations and automated driving systems, at least for partially automated vehicles (SAE Levels 2 and 3) should be designed to optimize the performances of both human and machine. Hence, more research is needed in fully immersive PAV simulation and traffic flow simulations to predict road safety effects of PAVs [369].

Ways forward in improving HMI design and usability include consistency and standardization, simplification of information presentation, and a user-centered design approach which caters for all users, including the fitness impaired [381] and people with disabilities [382]. Establishing consistency and standardization in HMI design elements across different PAV models and manufacturers improves usability. Standardized icons, control placements, and interaction methods help users quickly learn and operate PAV systems, reducing confusion and enhancing user confidence [364]. Adopting a user-centered design approach ensures that HMIs are designed based on user needs, preferences, and cognitive capabilities, and there are various human factors considerations that need to be taken into account in scientifically testing and evaluating automated driving systems [376]. Involving users in the design process through iterative testing and feedback helps identify usability issues early and refine interface designs accordingly [362].

### 8.5. Human Factors Challenges in SEA Levels 4 and 5

Human factors will not cease to be relevant when, and if, the entire vehicle fleet becomes entirely autonomous. Building human trust in their ability to operate and interact safely with other vehicles and road users will continue to be a major challenge. The HMI through which people summon, interact, communicate and are physically accommodated by them will need to be designed for easy and comfortable interaction for a wide range of able and not so able users, both of whom may require some level of training in how to safely and comfortably interact with them. Occupants will need to be alert, and understand how to respond, to unexpected events (e.g., emergencies and breakdowns), which may also require some level of training.

Likewise, vehicles will need to be trained to manage unexpected events within the vehicle cabin (e.g., sick occupants). Humans will still be required to monitor the movements of PAVs along the road network, as "ground traffic controllers", and in some circumstances (e.g., if they breakdown and cannot park themselves in a safe location) pilot them remotely to safety, like drones; and, of course, humans will continue to be required for some time to design the very algorithms that drive fully automated vehicles. Performance of these new human activities will bring with them a whole new range of human factors issues and challenges.

### 8.6. Conclusions and the Way Forward

The shifting role of the human driver from one in which they are in total control to one in which they are responsible primarily for monitoring and supervising the driving task performed by automation may lead to problems of inattention, reduced situational awareness and manual skill degradation. By incorporating human factors into the design, testing, and deployment of highly automated vehicles, stakeholders can create systems that not only enhance transportation efficiency but also foster a positive user experience and societal acceptance.

A significant challenge remains with SAE Level 3 vehicles, where human drivers must resume manual control if the automated driving system fails or reaches the limits of its competence. As technology advances, resolving human factors considerations becomes increasingly critical [383, 384]. Seamless transition between human and autonomous control is essential. This involves clear communication about when and why humans should take over. Sudden requests for human intervention can be problematic if the driver is not adequately prepared [385]. The road to fully autonomous driving is complex, and for various reasons slower than predicted. Without a comprehensive understanding of human factors and socio-cultural attitudes towards autonomy (Section 9), it will not be possible to navigate these challenges and realize the potential benefits of this transformative technology.

## 9. Attitudes and Perceptions

### 9.1. Introduction and Methodology

This section investigates the attitudes and perceptions of the public in relation to PAVs, as widespread opinions, perceptions, and feelings about autonomous technologies are likely to influence the propensity of the prospective users to adopt self-driving cars and rely on them. Despite their potential advantages, the adoption of PAVs has been slow [146] and experts whether this is caused by the negative psychological socio-cultural attitudes of the public, more than the objective limitations of the current technologies. Some of the attitudes studied by social robotics and Human-Robot Interaction scholars are particularly relevant in this context as they directly impact on the adopt propensity of people: the most important constructs to be considered in this context are Trust, Acceptance, Tolerance [412].

Other attitudes may play a secondary role as they indirectly shape the propensity of people to adopt autonomous cars and rely on them: these tendencies are studied under the label of "anthropomorphism" and concern the attribution of human-like qualities, like intelligence or empathy, to machines [413]. Attitudes and perceptions may vary significantly based on the demographical factors like age, sex, cultural background, etc. For instance, a survey conducted by the Australian Automobile Association (AAA) [147] revealed that 60% of respondents were open to using PAVs as an alternative to public transportation. However, 86% of American drivers expressed fear about riding in fully autonomous vehicles, demonstrating regional differences in attitudes towards PAVs. These regional differences suggest that cultural values, technological familiarity, and trust in institutional frameworks play a significant role in shaping public attitudes toward PAVs.

To address these challenges in the adoption of PAVs, it is critical to understand the interaction between public attitudes, external narratives, and policy considerations. Narratives delivered through media and public discourse can shape perceptions of PAVs, either positively or negatively, thereby influencing trust and acceptance of the technology. For example, media narratives often utilize framing effects to highlight either benefits, such as reduced traffic accidents, or risks, such as data security concerns, directly affecting public perception. Moreover, policy and regulatory frameworks play a vital role in establishing safety and performance standards, which directly impact public confidence in PAVs. These aspects must be considered together with factors such as affordability, safety concerns, privacy risks, and cultural influences, as they collectively shape public acceptance. In this context, societal and cultural factors further influence adoption. For example, in collectivist societies like China, social influence and peer approval significantly impacts individuals' attitudes toward new technologies [155]. Conversely, in individualistic societies, personal experiences and trust in institutions may play a more prominent role [159]. Finally, individual factors such as age and education level also influence adoption rates, with younger and more highly educated individuals generally being more receptive to PAVs.

This section adopts a comprehensive methodological framework to investigate the factors contributing to low public trust and limited PAV adoption. The analysis integrates findings from surveys and studies conducted across diverse geographic and cultural contexts, focusing on public perceptions, attitudes, and behavioral intentions. The data sources include responses from thousands of participants in countries such as China, India, Japan, the US, the UK, Australia, and Israel, as well as insights from global studies spanning over 100 countries. By combining quantitative analyses of large-scale survey data with qualitative insights into psychological and socio-cultural dimensions, such as trust, acceptance, and anthropomorphism, this framework provides a comprehensive understanding of the key factors shaping public attitudes and adoption of PAVs. This approach not only identifies universal trends but also highlights context-specific challenges, providing valuable insights for stakeholders to build public trust in PAVs.

## 9.2. The Trust Challenge: Overreliance and Under-reliance

Trust is a pivotal element in the adoption and acceptance of autonomous driving technology [359]. Trust in the context of automated and autonomous vehicles refers to the degree of confidence that users place in the vehicle's ability to perform driving tasks safely and reliably [352]. Trust is a critical factor influencing the acceptance of automated and autonomous vehicles, impacting both individual users and broader societal adoption. Human trust in automation is not entirely calibrated: sometimes it is too low (under-trust), and sometimes too high (overreliance). Overreliance occurs when a driver does not question the performance of automation and insufficiently counterchecks the automation status [371]. Overreliance results in drivers ignoring system limitations and failing to take over control when necessary [352, 372]. In contrast, under-trust occurs when drivers do not trust the automated system sufficiently, leading them to intervene unnecessarily or avoid using the technology altogether. This behavior can disrupt the vehicle's operations and reduce the perceived benefits of automation [373]. Hence, only a well-balanced amount of trust, motivated by knowledge, can produce the optimal diffusion of a new technology.

Trust is not static; it evolves based on users' experiences with technology. Positive experiences can build trust [374]. Trust issues indeed pose significant barriers to the rapid deployment of automated and autonomous vehicles. Users are more likely to adopt these technologies if they trust them to be safe and reliable. Conversely, mistrust can lead to resistance and slow adoption rates [354]. This challenge is particularly pronounced at the current stage of development, where the vehicles are not yet fully autonomous and require occasional human intervention (SAE Level 3). Trust issues in fully autonomous vehicles (SAE Level 5) will be critical but different in nature, as there will be no driver controlling the vehicle. Trust must be calibrated appropriately to the system's capabilities and limitations. Mismatches between user expectations and system performance can lead to dissatisfaction and misuse of the technology [375].

Hancock and colleagues indicated that transparent communication about the system's capabilities and limitations would be key to fostering appropriate levels of trust [376]. In addition to communication, Hoff and Bashir highlighted that education is important in ensuring that users know when and how autonomous vehicles make decisions [374]. Fagnant and Kockelman pointed out the important role of regulatory bodies' in ensuring that PAVs of Level 4 and 5 meet stringent safety criteria before they are widely deployed [377]. Burke outlines a range of human factors considerations in preparing policy and regulation for automated vehicles [378]. Establishing clear regulations and industry standards for AV performance and safety can help build public trust. It is human nature that a driver, who is relieved even briefly from their driving task, will engage in other distracting tasks. From a liability standpoint, the industry will not introduce such a distraction-inducing system unless the automation can bring the vehicle to a minimal risk condition if no driver response is detected. Fisher et al. [352] suggested that little opportunity to 'communicate with Level 5 vehicle' might lead to feelings of 'dread risk' that can undermine trust. Hence more research is needed on public acceptance and trust in automation, the interaction of the PAVs with other vehicles and road users, and the amount and type of information that the human drivers shall be receiving by the automated system.

### 9.3. Scarce Familiarity Problem

Lack of trust towards PAVs might derive from scarce familiarity with automation and poor knowledge of the technology, as only a very tiny percentage of the populace has had any direct experience of autonomous drive to date. This lack of direct experience is often filled up by narratives and opinions that are not necessarily accurate. The problem of limited familiarity and inadequate knowledge about PAVs significantly affects their acceptance and adoption. Due to the lack of direct experience, individuals often rely on external narratives (many of which are inaccurate or misleading) to form their perceptions. Schoettle and Sivak [148] conducted a survey to investigate public opinion on PAVs across six countries: China, India, Japan, the US, the UK, and Australia. While most respondents expressed positive attitudes toward PAVs, the study reflected a significant issue of limited familiarity with PAVs across these countries. For instance, 87% of respondents in China had heard of PAVs prior to the survey, compared to only 57.4% in Japan, illustrating a considerable disparity in awareness. In addition, knowledge of automation levels was notably lacking. Among Japanese respondents, 50.6% reported using vehicles without any automation (Level 0), and 10.8% were uncertain about the level of automation in their vehicle's. This limited familiarity and practical exposure suggest that public opinions are often shaped by external narratives and media coverage rather than firsthand experience, which could lead to misunderstanding or suspicion toward PAVs.

Kyriakidis et al. [149] conducted a global survey involving 5,000 participants from 109 countries to explore public opinion on PAVs. The findings revealed that unfamiliarity with PAV technology significantly impacts its acceptance. For example, although most of respondents found the idea of fully automated driving fascinating, only 52.2% were aware of Google's Driverless Car, reflecting significant gaps in knowledge about autonomous technologies. Furthermore, participants expressed a preference for manual driving as the most enjoyable mode (mean value = 4.04 on a 1-5 scale), indicating limited awareness of the potential benefits of PAVs. The study further revealed that many respondents lacked understanding of existing automation levels in vehicles, and 22% indicated that they would not pay for fully automated driving systems. Haboucha et al. [150] found that limited familiarity with PAVs has a significant impact on adoption rates.

A survey of 721 participants from Israel and North America revealed that 44% consistently chose traditional vehicles in all scenarios, demonstrating hesitations resulting from a lack of knowledge. The study further revealed that individuals with lower education levels, less technological exposure, or limited understanding of PAV capabilities were less likely to adopt PAVs. Furthermore, 25% of respondents indicated they would refuse to use shared autonomous vehicles even if provided free of charge, revealing a lack of trust rooted in unfamiliarity. Lee et al. [151] conducted a large-scale survey of 1,765 participants in the United States to explore how age and technological experience influence perceptions of and attitudes toward PAVs. The study found that limited familiarity with technology significantly hinders the acceptance of PAVs, especially among older generations. For instance, older respondents reported lower general trust in technology (mean = 2.93 on a 1-5 scale) and less confidence in learning and using new technologies compared to Millennials (mean = 3.15 vs. 3.78 on a 1-5 scale). Furthermore, older adults were more likely to view PAVs as less useful, reliable, and compatible with their lifestyles.

In conclusion, the limited public familiarity and knowledge of PAVs represent significant barriers to their acceptance and adoption. Addressing these challenges requires a comprehensive strategy, including public education activities and interactive demonstrations. By providing opportunities for hands-on experiences with autonomous technologies and delivering accurate, transparent information, particularly on safety, reliability, and potential benefits, these efforts can effectively increase public familiarity, clarify misconceptions, and build trust in PAVs.

### 9.4. Low Acceptance and Technology Resistance Problem

Besides the lack of trust, a low propensity to adopt PAVs may derive from widespread narratives and beliefs that are hostile to the introduction of autonomous technologies, for example when they are associated in public discourse with technological unemployment or threats to human uniqueness. Lee et al. [152] examined the influencing factors on PAVs through collecting responses from 459 South Koreans over 20 years of age. The survey results shown that factors directly related to drivers such as anxiety, carelessness, ease of driving and driving education influence the acceptance of partial autonomous vehicles, while external environmental factors such as extra expenses and infrastructure affect the acceptance of full autonomous vehicles.

Kaye et al. [153] found that individuals residing in France have greater intentions to use PAVs compared to individuals residing in Australia and Sweden. At the same time, Potoglou et al. [154] investigated the consumers' intentions to pay for both autonomous and alternative-fuel vehicles through performing an experiment in six countries, Germany, India,

Japan, Sweden, the UK and the US, and found significant heterogeneity both within and across the samples. In particular, consumers in Japan are willing to pay for PAVs, while consumers in most European countries need to be compensated for automation. As regards samples from the same country, the consumers most enthusiastic about PAVs usually have a university degree and are more interested in novel technologies.

Man et al. [155] applied a technology acceptance model to identify the factors influence the acceptance of PAVs among Hong Kong drivers. They found that trust and perceived usefulness positively determine the attitudes and attentions to use PAVs. Zhang et al. [156] reported an investigation of the automated vehicle acceptance in China from the perspectives of social influence and initial trust. They conclude that both social influence and initial trust play important role in determining users' intention to use PAVs. First, due to the influence of collectivist culture, the individual's decision is likely to be influenced by other people's opinion because of face saving and group conformity. Hence, social influence has a stronger impact on technology acceptance behavior in Chinese culture than it in western culture. Moreover, users with an openness to new experiences are more likely to accept PAVs and have a higher intention to trust them.

### 9.5. Cultural Prejudices Problem

Due to different value-systems and practices, different cultures may have different attitudes and perceptions towards the adoption of emerging technologies, with certain cultural traditions more inclined to trust innovative and disruptive solutions than others. This might be an issue for establishing international standards and policies. In 2021, Huang and Qian [157] performed a nation-wide survey in China to investigate the influence of reasoning process on consumer's attitude and intentions towards PAVs. They found that one of the Chinese cultural values, "face consciousness", positively influence the adoption of PAVs from dual perspectives because of the competing perception on the desirability of adopting PAVs: "face" represents an individual's desire to gain, maintain, and avoid losing fac$^e$ in relation to others in social activities, and refers to a sense of favorable social self-worth that an individual desires others to have of him or her in a relational and network context [158].

More specifically, PAVs are priced with high premiums and equipped with rich applications of novel technologies (e.g., vision-based driver assist features), and will be a symbol of trendy technological product. Under this condition, the feeling of pride, dignity, and vanity derived from self-driving vehicles may drive consumers to adopt the technology. At the same time, PAVs are still considered as a risk-taking choice which are connected with legal and ethical doubt. Therefore, face consciousness may lead consumers to choose more mature and widely accepted vehicles.

Escandon et al. [159] investigated the influence of the indulgence dimension on the relationship between risk perception (e.g., financial, psychological, and time) and purchase intention in autonomous vehicles in Vietnam and Colombia. As a psychological metric, indulgence refers to the cultural dimension that emphasizes the gratification of human desires and enjoyment of life: it is associated with societies that prioritize leisure, pleasure, and the pursuit of happiness, contrasting with cultures that value restraint and self-control. The study collected questionnaires from 800 Colombian and Vietnamese car drivers aged 18 or over and found that indulgence directly affect the adoption of autonomous vehicles. In low indulgence country, Vietnam, consumers tend to pay more attention to financial and psychological risks. In high indulgence country, Colombia, irrational emotion (e.g., fulfilling desire) is the decisive factor for purchase intention. However, for the time risk, the influence of indulgence exists in both countries.

Yun et al. [160] investigated the relationship between culture difference and public opinion on PAVs in China, India, Japan, the US, the UK and Australia. The investigation demonstrated that cultural difference plays an important role in the acceptance of PAVs. Specifically, more individualized societies are less willing to pay for autonomous vehicles, and more indulgent and less hierarchical societies are less willing to pay for and less concerned about PAVs. However, the uncertainty avoidance that refers to the degree to which the individuals of a society feel uncomfortable with uncertainty and ambiguity is insignificant for willingness to pay and levels of concern about PAVs.

Gopinath and Narayanamurthy [146] indicated that the adoption of PAVs is moderated by the level of automation, vehicle ownership, and culture. Taniguchi et al. [145] investigated the acceptance of PAVs in Japan, the UK and German. They found that cultural differences have an important influence on the attitude towards PAV in these three countries. In particular, the Japanese participants are broadly positive, the British participants are broadly neutral, and the Germany participants are broadly negative. The result suggests that participants from a more hierarchical and more "masculine" nation are more likely to accept PAVs (according to Hofstede's cultural dimensions theory, a high masculinity society is a society in which social gender roles are clearly distinct and people tend to hold strong opinions about the occupation of men and women).

### 9.6. Conclusions and the Way Forward

This section explores public attitudes and perceptions toward PAVs, focusing on trust, familiarity, acceptance, and the cultural factors influencing adoption rates. While PAVs offer substantial social and economic benefits, such as reducing road accidents, improving mobility, and minimizing environmental impact, their widespread adoption faces challenges rooted in technical, psychological, and socio-cultural factors.

Trust plays a pivotal role in PAV adoption, influenced by both over-reliance and over-reliance on PAV systems. Studies emphasize the importance of calibrated trust, transparent system functionality, and robust safety regulations to build confidence in PAV technology. Limited exposure to PAV technology and insufficient understanding of automation levels deepens public scepticisms, often leading individuals to rely on inaccurate narratives. Cultural factors also impact public attitudes. For example, in collectivist societies like China, social norms strongly influence acceptance, whereas individualistic cultures, such as those in the United States, show greater resistance to additional costs associated with PAVs. Variations in cultural values, including face consciousness and indulgence, complicate the development of universal standards and policies. Furthermore, public narratives emphasizing either benefits (e.g., reduced traffic accidents) or risks (e.g., data security concerns) significantly shape trust and perceptions of PAVs.

In conclusion, to address these challenges, governments and technological firms will need to cooperate to design education programs tailored to the public's needs, community engagement activities, and collaborative efforts among researchers working in the public and in the private sector. As suggested in Section 8, providing hands-on experiences and transparent information about safety, reliability, and potential benefits, is a necessary measure to build trust, address public concerns, and pave the way for widespread PAV adoption.

## 10. Ethical Dilemmas

### 10.1. Introduction and Methodology

This section considers the ethical challenges and moral problems relating to the development and implementation of autonomous personal vehicles. Following the characterisation of Chella *et al* [387], ethics pertains to creating and deploying AI that is "safe, lawful, just, respectful of the rights of users, culturally sensitive, non-discriminatory" and promotes the wellbeing of individuals and society. We are interested in the ethical complexities when designing or implementing PAVs – in particular, situations where it is not immediately clear what the right or best course of action might be. Such problems are pertinent not only for direct users of PAVs, but for governments, policymakers and the broader public, as well as the manufacturers, engineers and designers behind PAVs.

The main ethical problems regarding PAVs today are addressing trolley problems, social acceptability and the question of accountability. The trolley problem is a thought experiment initially conceived by Philippa Foot [393], where a moral decision must be made between different combinations of lives being saved and sacrificed [400]. With respect to PAVs, trolley problems are discussed as moral dilemmas which arise from extreme traffic scenarios (e.g. when an accident is unavoidable, whether swerve to hit two elderly people or sacrifice a sole younger passenger). Social acceptability is similarly an ethical issue as reluctance from the public will slow down PAV implementation timelines, even if such vehicles are proven to be safer than human drivers. Considering accountability also presents an ethical challenge – who will be held responsible for accidents caused by PAVs in the future?

Published works discussed in this section were chosen according to the following criteria: published since 2019, peer reviewed, journal impact factor greater than one. The keywords searched were: "ethical challenges"; "moral issues"; "autonomous vehicles"; "trolley problem"; "accountability". A few exceptions were made to include landmark papers in the field, such as the Moral Machines experiment [386].

### 10.2. The Trolley Problem

Let us first consider the perennial and oft-discussed issue for autonomous personal vehicles – the trolley problem. Although initially an ethical thought experiment, the trolley problem has very real-world implications for autonomous technologies. In a PAV context, a trolley problem is an extreme traffic event where some sort of accident or collision is likely unavoidable. The PAV system must reconcile how to prioritise the safety of passengers and any number of pedestrians involved [409]. We can appreciate the infinite number of complex extreme traffic scenarios that could arise and necessarily warrant some sort of ethical decision-making from a PAV system (involving different combinations of vehicles, surrounding infrastructure, passengers and pedestrians).

Cuneen *et al.* describe how the current state of PAV technology operates with a "shared responsibility" between the driver and the PAV system [388, p. 66]. This is expected to remain the case until PAVs are able to perceive their environment and navigate as effectively and safely as humans. We are particularly interested in this ethically complex future where agency will inevitably shift from a human driver to the PAV system.

How will a PAV make decisions in complex traffic scenarios? One approach will be for PAVs to act in accordance with some sort of preprogrammed ethical decision-making calculus. Several different approaches have been debated for this programming. These include selection of normative theories (e.g. utilitarian or deontological) or the aggregation of societal preferences. So, how do we choose between these approaches? This was the question asked by Awad *et al.* in the "Moral Machine (MM) experiment" – a piece of research that aimed to determine collective ethical preferences to inform AV system design and has greatly influenced the field of PAV ethics research. Although Awad *et al*. found an overall preference for PAVs sparing *more lives* and sparing *young lives*, there were significant variations across cultures. Many scholars agree that no single normative theory is a sufficient or appropriate decision-making calculus for a PAV but are also wary of deferring to majority public opinion [391, 392, 386]. A combination of these two approaches is likely the best way forward. Recent methods such as the Ethical Valence Theory [392] and integrated path-planning that considers future actions [410] aim to more effectively automate trolley-like decision making – but there is little consensus on the best course of action.

Further, who gets to ultimately choose the decision-making calculus of a PAV? Whether this choice lies with the owner, the car itself, the manufacturer, the government or another authority will be influenced by the conversation around autonomy and liability (see Section 10.4 and Section 11.3 for discussion). We may reach a point in the future where PAV systems will make ethical decisions based on their own free reflection – i.e. decision-making in real time. This would of course require a level of system sophistication that would go beyond mere pre-programming of a PAV (requiring a sufficient amount of *common sense reasoning*, as per Section 5).

PAV trolley problems have also come under a host of scrutiny, with suggestions that they are unrealistic and have been unhelpfully framed by the media and academics alike [388, 397, 395, 389, 400]. De Freitas *et al.* argue that trolley problems are unlikely to occur on real roads, difficult to detect, and too hard to control or respond to reliably [389]. Cunneen *et al.* concur, advocating for more realistic framing of PAV-related ethical issues as determined by current technological progress [388]. For example, the notion that PAVs will be sophisticated enough to classify pedestrians and other agents (i.e. as young/old or male/female) will rely on vehicle-to-vehicle and vehicle-to-infrastructure sharing of data, which remains decade/s in the future [388]. PAVs today, by contrast, are still refining accurate perception of the environment [394]. Königs (2023) similarly makes the case that trolley problems are too unlikely to give any meaningful moral guidance, however believes there are methodologically important as they reveal that our instincts/prejudices can lead us astray [399].

It should also be noted that there is comparatively little discussion of trolley problems in PAV industry reports [400]. In their comprehensive document review of PAV manufacturer reports, Martinho *et al.* (2021) found no mentioned of trolley problems – rather, ethically complex extreme traffic scenarios were often mentioned as *edge cases* (including by General Motors, Nvidia, and AutoX) [400]. Some manufacturers offer more transparency on this topic than others. For example, in PAV manufacturer Nuro's *Delivering Safety: Nuro's Approach* report [404] the company stipulates that its vehicles can self-sacrifice: "in the unlikely case of a vehicle ever encountering an unavoidable collision scenario, a driverless, passengerless vehicle also has the unique opportunity to prioritize the safety of humans, other road users, and occupied vehicles over its contents" (p. 9).

Despite academic critique of sensationalised trolley problems, discussion remains ongoing: the nature of such problems is of poignant interest to the public and poses genuine pragmatic challenges for policymakers and manufacturers alike. Rather than waiting for consensus on a PAV ethical decision-making process, however, manufacturers are likely to proceed with basic measures such as optimising radars, speed reduction and avoidance of vulnerable agents (i.e. cyclists, pedestrians) to deal with crash scenarios [400, 408]. Mobileye, for instance, asserts that a PAV "should adjust its speed such that if a child emerge from behind some [occluded] object, then there would be no accident" [402, p.14]. Additionally, many manufacturers, including Toyota [406], Mobileye [402], and Mando [401], describe their vision of a PAV future which has no accidents at all [400].

**10.3. The Value Alignment Problem**

Let us discuss the role that public opinion plays in PAV development and deployment, and how this can be considered its own ethical challenge. While accidents will not be avoided altogether, it is widely accepted that adoption of PAVs will make roads significantly safer than they are today [392, 408, 400]. Once PAVs become safer than human drivers, are we morally obligated to roll them out? At this milestone, in the interest of saving lives, we could even make the case for relaxing PAV regulations, lowering manufacturer liability or phasing out human driving altogether – which would also make traffic more sustainable, safer and more inclusive [391].

Robinson *et al.* [405] point out that unless PAVs are embraced by the public, they will not realise their full potential or be successfully integrated. Nevertheless, we should not underestimate the challenge of aligning PAVs with a diversity of social values. Not only must PAV development be sensitive to cultural differences, as demonstrated in the MM experiment, it must also be distanced from human bias, prejudice, ignorance and egoism [392]. One of the major cultural differences observed is the way those from more individualist cultures prioritise protecting the greatest number of people, while those from collectivist nations believe elderly people are due a certain respect [386]. In addition, many prospective PAV users would prefer others to ride in utilitarian PAVs which protect as many lives as possible, while others would prefer to ride in PAVs that prioritize themselves [408].

The over-reporting or incorrect framing of trolley-like problems, together with high-profile fatalities such as that of an Arizonian pedestrian during Uber testing in 2018, do have large effects on public trust and will decelerate or even stall PAV adoption [395]. It has also been shown that we have higher standards for AV driving than human driving, which similarly contribute to delayed implementation [397]. Given the significant impact that social acceptance of PAVs will have, it is important that this technology is explainable [388] and communicated in an effective way. Several scholars [407, 396] advocate for Value Sensitive Design –integrating the values of various stakeholders – throughout the design and development of PAVs. In addition to explainability, Umbrello and Yampolskiy argue that PAV systems must have "traceable lines of decision-making logic that are…repeatable and consistent in order to be *verifiable*" [407, p. 316]. Value Sensitive Design is particularly important as we envision PAVs of the future that may be able to exercise common sense. Until then, we continue to maintain a level of meaningful human control over the decision-making systems employed by artificial systems – in order to ensure a level of value alignment, and, importantly, safety.

Regardless, at some point we will need to more honestly reflect on the point at which delaying PAV roll-out in the name of safety becomes just the opposite. Evans *et al*. describe this challenge as "striking a balance" between public acceptability and the moral requirement to introduce PAVs [392].

**10.4. The Accountability Problem: Apportionment of Responsibilities**

The final conflict we will examine is that which emerges around accountability. The evolution of autonomous driving technology brings forth the imperative of accountability, as AI-powered systems increasingly assume control over critical driving functions. We are particularly interested in accountability in relation to questions of control, blame and compensation (note that the term accountability will be used interchangeably with responsibility here). Present day testing of PAVs still relies on a human passenger as a fallback system to take control if need be. However, when PAVs transition from SAE Levels 0-2, where control remains with the driver, to Levels 4-5, where the vehicle is making decisions [403], who is responsible? We see what de Jong (2020) and others call a *responsibility-gap* arise here; the answer is not ultimately clear [390].

While the most obvious and most likely candidate for accountability is the vehicle manufacturer, other important players to remember are the road system facilitators (e.g. vehicle communication and guidance systems), policymakers and the AI built into the PAV – plus even pivoting to instead view AV traffic accidents as natural accidents like one would a cyclone or tsunami [397, 403]. Assigning responsibility also has very important consequences in a legal sense. In complex or trolley-like situations, what may be a more ethical course of action for a PAV (e.g. more lives saved) will not always have lower associated liability or be easier to defend in a lawsuit [409]. We will see tensions emerge between ensuring PAVs align with social values and dealing with situations fairly in our legal systems. There is clearly a conflict of interest between user desire for AV safety and manufacturers' vested interest in strategies with the lowest liability [405, 409]. Regulators will need to be proactive as PAV technology evolves. A lack of explicit policy will exacerbate uncertainty in this space, leaving important technological decisions to approaches that are market-driven and socially palatable [403].

Wu suggests that the law may even need to be altered to provide manufacturers immunity from liability for PAVs carrying out an approved ethical decision (2020). Future directions for enhancing accountability will involve the formulation these new legal and ethical frameworks that delineate the responsibilities of autonomous vehicle manufacturers, developers, and operators based on AI system autonomy levels. The integration of AI and ethical frameworks will enable vehicles to tackle moral dilemmas, making decisions that align with societal values and prioritise human safety. It is important to remember, however, that assigning blame is perhaps not as important as victim compensation or taking steps to prevent or reduce the risk of similar accidents in the future [397]. For now, preserving a level of meaningful human oversight fosters accountability through shared decision-making and the preservation of human agency. This allows users of PAVs to intervene in driving manoeuvrers, thereby sharing responsibility and enabling a dynamic collaboration between AI and human drivers.

**10.5. Conclusions and the Way Forward**

This section has reviewed three key ethical challenges to designing and implementing PAVs: reconciling AV trolley problems (Section 10.2), social palatability of the AV technology (Section 10.3) and the question of accountability (Section 10.4). However, the ethics of PAVs will need to address also other kinds of considerations. For example, as is typical of safety-critical technology, various concerns emerge from the understandable interest in maintaining a level of *meaningful human control and oversight* – highlighting our general wariness to defer decision-making power to a PAV [407]. We must be careful, however, not to reduce ethical concerns to risk management, or reduce risk management to safety considerations, as overemphasizing safety might, paradoxically, slow down the deployment of a technology that has the potential to make roads much safer in the future.

How will these challenges be addressed moving forward? With respect to trolley-like problems, research will continue to emerge where PAVs must exercise caution in dealing with increasingly complex situations (e.g. predicting pedestrian and cyclist intentions) [398]. We will also see AV technical solutions focus on combatting accountability concerns, such as with responsibility-focused decision-making algorithms or "black-boxes" akin to those on flights [400]. Further, linked to the challenge of accountability is the growing concern around AV data ownership, privacy and potential misuse of information, given the large volume of person-related geospatial data that will be available [397, 391, 409]. As we will see in Section 11, conversations around PAVs require a more sophisticated legal perspective that can incorporate liability as well as these security and data concerns. Many scholars call for greater collaboration across the board in the field of PAV ethics, in particular between academia, manufacturers and policymakers [408, 386, 388]. Efforts to roll out electric vehicles represent a significant term of comparison, as they have highlighted how the importance of public awareness, attitudes, and education cannot be understated, and a one-size-fits-all solution is not to be expected.

# 11. Policy and Regulation

## 11.1. Introduction and Methodology

The development and deployment of PAVs challenges policymakers internationally to formulate regulations, standards, and governance frameworks that maximize the anticipated benefits of the rapidly developing technology while guarding against potential negative impacts [434], [435]. This section reviews the literature on the main emerging policy issues to date, considers the extent to which each issue is putting breaks on the roll-out of PAVs, and highlights existing solutions. It concludes by identifying potential ways forward that could help policymakers design and adopt policies and standards to unlock the potential of automated driving technology and facilitate its widespread adoption. The main policy challenges identified in the literature can be grouped into the following three categories: safety and licensing, liability and insurance, and cybersecurity and privacy [436]. These issues

"represent substantial barriers to widespread [autonomous vehicle] technology implementation… [and] should be addressed to give manufacturers and investors more certainty in development" [437, p.16]. Focusing on these three issues is not intended to downplay the other challenges that policymakers will be required to solve, such as those related to the costs, infrastructural requirements, and sustainability of PAVs [438], [436]. However, those three issues have been selected for this review as they are the priorities that must be addressed rapidly to enable PAVs to move successfully from the current testing phase to widespread commercial deployment [439], [440].

When considering each of these issues, it is also necessary to be cognizant of the overarching challenges of policymaking in this field, which include the following: the insufficiency of existing regulations as distinctions between vehicle and driver become increasingly blurred [441]; the speed at which the technology is developing and outpacing regulatory efforts [442], [403], [440]; the challenge of regulating for an uncertain future [443], [444]; the complexity of harmonizing regulations across different jurisdictions [445]; and the cross-over between issues affecting PAVs and rapidly developing policy responses in related fields, most notably the AI systems upon which PAVs are almost entirely dependent [446]. We will return to each of those cross-cutting issues when reviewing the literature on the specific policy challenges and solutions.

The publications reviewed in the following sections were identified through searches of online databases, including Google Scholar and JSTOR, focusing on the years 2013–2024. Keywords searched included "autonomous vehicles" OR "self-driving car" AND "policy" OR "regulation." The articles selected were all published in journals with impact factors equal to or over 3.5. However, given the nascency of the field, the search was extended beyond such articles to include books, consultancy reports, and government publications to ensure that wide-ranging, relevant contributions were included.

## 11.2. Safety and Licensing

Unsurprisingly, the safety of PAVs' users is policymakers' key priority, and failing to address it satisfactorily would render all other policy developments irrelevant [436]. Safety concerns will potentially put major breaks on the roll-out and commercialization of SAVs if they undermine public trust in the technology and deter regulators from granting licenses for its usage [447]. Permissive regulations may appear to be in the industry's interests, but that would only be in the short term if they lead to safety issues that destroy public confidence in PAVs, which explains why the stricter safety regimes of jurisdictions like the EU may give them advantages over the relatively permissive US in terms of developing PAVs that consumers trust [448].

However, overly cautious safety regulations, especially in the testing phase, risk undermining the innovation that is essential if PAVs' positive potential is to be realized [445]. The challenge here is balance, but achieving it is complicated by the multi-level regulatory environment in which PAV manufacturers must operate. Shladover and Nowakowski [439] identified the two main elements of automotive safety as mechanical faults and drivers' competency. But in the US technological aspects of vehicles are regulated at the federal level by the National Highway Traffic Safety Administration (NHTSA) whereas the training of drivers, licensing, evaluation, and vehicle registration are the responsibility of the fifty state governments' departments of motor vehicles [449], [450]. The complexity of this situation with regard to PAVs—in which the technology increasingly *is* the driver—is clear.

That complexity is increased by the differing safety policies issued at different regulatory levels. Most national governments that have issued policies have made them non-binding [451], such as the NHTSA's guidelines to manufacturers and states [452]. State-level regulations meanwhile range on a continuum from California, which has set its own, binding, regulations on automated vehicle testing and deployment, to states such as Delaware and Indiana which have made no regulations [453]. Those risks creating what Anderson et al. [450, p. xxiv] called "a crazy quilt of different, and perhaps incompatible, requirements." This lack of harmonization could create unnecessary overlaps and regulatory uncertainty for the manufacturers of PAVs [454], limiting their abilities to develop, commercialize, and roll out the technology. Nevertheless, citing a lack of demonstrated problems with automated vehicles, Anderson et al. [450] argue in favor of holding off from attempting to impose federal standards across the country as such standards may both stifle innovation and become rapidly outdated as the technology develops.

In terms of solutions, Koopman [448] has called for increased regulatory oversight to ensure the safety of automated vehicles and create the levels of public trust required to support their deployment. Koopman [448] argues that policies to improve safety regulations for automated vehicles should have two elements: requiring the industry to self-certify conformity with its own safety engineering standards (an approach already proposed by the US Department of Transportation with its automated vehicle framework) and setting data reporting standards regarding accidents. The transparent sharing of such data increases accountability and allows potential risks to be identified and mitigated [436], [435].

Another solution is highlighted by Wansley's [455] study of the NHTSA's "novel regulatory strategy," which it has applied since 2021, which balances ordering daily crash reporting with the threat of rapid recalls in cases of technological failure. Wansley [455] argues that the strategy can be transformed into effective regulation by increasing the flow of safety data and enforcing the threat of recalls if manufacturers' vehicles create unacceptable levels of risk. Such an approach would pressure manufacturers to prioritize safety during development as a means of decreasing the costs and reputational damage associated with recalls. Other recommended safety solutions being applied in various jurisdictions include the following. California requires safety drivers and remote monitors to undergo additional training and stipulates that shifts from using the former to the latter require the issuance of new licenses with extra provisions [436]. Australia requires testing companies to develop comprehensive safety management plans which are monitored by recording trials, failures to comply with safety rules are also punishable with substantial fines and sanctions [436].

## 11.3. Liability and Insurance

Dentons [453] describes liability as one of the "twin pillars" that regulators of PAVs must address but argues it has been thus far neglected in favor of a primary focus on the other pillar: safety. That position is supported by international analysis showing that, despite much discussion, jurisdictions have been slow to adopt new liability provisions that specifically address the cases of autonomous vehicles [456]. As indicated above, the question of liability in the case of automation that blurs the distinction between vehicle and driver becomes increasingly complex as the level of automation increases [436], [446], especially because the potentially liable parties include not only the vehicle's manufacturer and its occupants but also other third parties along the supply chain [451].

Failing to create clarity in such cases or applying liability requirements that manufacturers deem overly threatening could restrict the commercialization of automated vehicles. Although one of the main arguments in favor of automation is that it will be much safer than human driving, Fagnant and Kockelman [437] have argued that holding PAVs to a much higher standard than humans when it comes to accident liability will increase the costs of the technology to such an extent that it will only be affordable by a select few. For that reason, Gless et al. [457] argued in favor of limiting manufacturers' liability provided they undertake to implement reasonable risk-control measures. Relatedly, Contissa [458] has argued that standards that limit manufacturers' design options can help to reduce their liability and thus encourage them to invest in PAV production and bring attractively priced vehicles to market. However, imposing such limitations comes with an obvious trade-off in terms of limiting innovation. To allay manufacturers' fears regarding liability while avoiding that trade-off, policymakers could consider placing limits on the amount of damages victims of accidents could claim due to PAVs' faults and governments could incentivize insurance companies to provide attractive policies for the makers of automated vehicles [452].

Koopman [448] makes the case that the lobbying efforts of manufacturers attempting to reduce their liability in the case of accidents are largely responsible for the patchwork of different approaches to questions of liability that can be found from state to state in the US. He argues that placing computer drivers under explicit duties of care at a federal level, with the manufacturer held responsible for any breaches of that duty, would resolve questions of liability. Rather than limiting the spread of automated vehicles, such approaches could enhance it by providing potential purchasers of such cars with clarity and reassurance regarding situations in which the manufacturer will be found liable for problems. In terms of specific solutions being implemented, California stipulates that testing permits will only be granted to companies that have substantial insurance [436]. Further, requirements to install black boxes can help to determine liability in the event of an accident [440]. Dentons [453] identifies Germany's three-pillar liability model (manufacturer, owner, driver) as an example of a balanced distribution of risk from which other jurisdictions could learn.

## 11.4. Data Privacy and Cybersecurity

The third main policy issue related to the widespread deployment of PAVs concerns the related challenges of data privacy and cybersecurity [436]. Autonomous vehicles are dependent upon gathering and analyzing data to complete their journeys. However, that results in the collection of a substantial quantity of personal data, primarily, but not exclusively, related to the location of the vehicle's users. Such data collection and storage create the potential for violations of individuals' rights regarding data privacy [459] and could facilitate a range of unwanted violations from targeted advertising to unlawful monitoring and stalking [437]. This raises questions about how to apply existing laws in this field, such as the European Union's General Data Protection Regulation, to automated vehicles [446]. Given that the AI systems upon which automated vehicles are reliant need data to function, learn, and improve, any restrictions placed on the gathering and use of such data to comply with regulation risks undermining innovation in the sector.

However, failure to regulate appropriately around issues of data privacy could generate resistance to the adoption of PAVs [445]. Sever and Contissa's [440] analysis of countries' acts related to self-driving vehicles shows that they do not include rules that explicitly address the protection of users' data. Exceptions to that include Germany and Australia, which has more actively engaged with the public regarding data privacy issues in the context of automated vehicles and formulated recommendations [451]. But perhaps the general lack of action to explicitly address such issues reflects the fact that questions of personal data are of less concern during the testing phase [436]; nevertheless, that gap will need to be filled as PAVs move toward more widespread deployment.

Relatedly, the possession of valuable personal data incentivizes hackers to attempt cybersecurity breaches [415]. Given the cybersecurity risks related to data, Lee [460] argues for national governments to take responsibility for creating regulatory frameworks that address cybersecurity and data privacy issues in the specific context of autonomous vehicles. Cybersecurity threats could also, potentially, involve hackers forcing automated vehicles to crash [437]. Koopman [448] claims that, despite such risks, the industry has underinvested in addressing cybersecurity and needs to be required to conform to its own cybersecurity standard, which is set out in ISO/SAE 21434. Taeihagh and Lim's [451] analysis of governmental responses to cybersecurity issues internationally reveals the familiar mix of depending on non-automated-vehicle-specific regulations, further researching the issues through working groups, and offering non-binding principles to manufacturers. Such reticence to take decisive action reflects another familiar response to the policy challenges associated with PAVs:

| Country/Region | Key Recent Devs | Distinctive/Notable Elements | Safety & Licensing | Data Privacy & Cybersecurity | Liability & Insurance |
|---|---|---|---|---|---|
| USA | - **2021**: *Automated Vehicles Comprehensive Plan* (USDOT) [462] - **2021**: *NHTSA Standing General Order 2021-01* for crash reporting [463] - **2022**: *AV TEST Initiative Updates* (NHTSA) [464] | - **Highly Decentralized**: State vs. federal roles remain distinct, creating both flexibility and inconsistencies. - **Emphasis on Industry-Led Innovation**: Voluntary guidelines rather than rigid federal regulations. - **Enhanced Data Collection**: AV TEST Initiative fosters real-time data sharing and transparency on AV performance. | - **Federal Strategy**: The *Automated Vehicles Comprehensive Plan* outlines collaborative safety goals and interagency coordination. - **Crash Reporting Requirements**: Under *Standing General Order 2021-01*, manufacturers and operators must report certain ADS/ADAS-related crashes to NHTSA. - **State-Level Testing Permits**: Most states allow AV testing with public reporting of disengagements (CA DMV regulations). Others (e.g., Arizona) have executive orders allowing broader testing. - **Voluntary Tech Standards**: NHTSA encourages industry-led safety assessments and publishes recommended best practices. | - **No Overarching Federal Data Privacy Statute**: AV data is generally subject to a patchwork of sectoral and state privacy laws (e.g., California Consumer Privacy Act). - **Cybersecurity Guidance**: NHTSA encourages manufacturers to develop vulnerability reporting policies, secure over-the-air updates, and integrate cybersecurity risk management from design to deployment. [462][463] - **Crash Data Access**: Under *Standing General Order 2021-01*, NHTSA can collect and analyze event data to identify safety-critical flaws. | - **Patchwork Liability**: Varies by state, but many jurisdictions treat the AV system manufacturer as liable when in fully automated mode (if proven defective or unsafe). [462][463] - **Insurance Requirements**: States generally require minimum motor vehicle insurance; with some (e.g., Michigan, Florida) explicitly defining coverage rules for AV operations/sectors. - **Incident Reporting & Investigations**: NHTSA can request detailed data post-crash to determine if system malfunctions or manufacturer faults contributed to an incident (enhanced by 2021–2022 NHTSA orders). |
| China | - **2021**: *Guiding Opinions on Network Security for ICVs* (MIIT & CAC) [465] - **2022**: *Shenzhen ICV Regulations* [466] | - **Top-Down Leadership**: Rapid government-driven deployment in pilot zones. - **Integration with Smart Cities**: Heavy investment in roadside sensors, 5G networks, and IoT infrastructure. - **Shenzhen as a Test Bed**: Detailed local regulations that could serve as a national template. | - **National Testing Framework**: Central guidelines for AV road tests require obtaining licenses from local authorities; pilot programs in Beijing and Shanghai have expanded from testing to commercial robotaxi services. - **Shenzhen Regulations (2022)**: Allows certain fully autonomous operation (under specific conditions) and sets out detailed requirements for safety drivers and remote monitoring centers. - **Infrastructure Focus**: Emphasis on V2X technologies and dedicated lanes in pilot zones to boost safety and reduce congestion. | - **Mandated Data Localization**: Under the *Guiding Opinions* and other data security laws, certain critical AV data must be stored on servers located in China. - **Cybersecurity Requirements**: Stringent obligations for hardware and software testing, encryption, and real-time risk monitoring. - **Personal Data Protection Law (2021)**: Requires explicit consent for collecting personal information within vehicles (e.g., biometrics for user authentication). | - **Strict Operator/Owner Liability**: Generally, the vehicle owner or operator is responsible for damages, but Shenzhen's new rules allow assigning liability to the developer of the autonomous system if its system is proven to be at fault. [466] - **Compulsory Insurance**: China requires motor vehicle insurance, and large insurers are developing specialized policies for autonomous fleets. - **Local Variations**: Some municipalities (e.g., Beijing) hold testing entities responsible for accidents during trials; others refine liability laws as commercial deployment expands. |
| EU | - **2021**: *Proposal for an Artificial Intelligence Act* [467] - **2022**: *Commission Implementing Regulation (EU) 2022/1426* on type-approval for advanced systems [468] | - **"Brussels Effect"**: EU tends to set stringent data protection and product liability standards that influence global practices. - **AI Act**: If passed in its current or revised form, it could impose risk-based obligations for AVs as high-risk AI systems, imposing strict oversight and compliance obligations. - **Focus on Consumer Protection**: balancing innovation incentives with robust safety, privacy, and liability frameworks. | - **Type-Approval Updates**: *Regulation (EU) 2022/1426* sets technical requirements for Automated Lane Keeping Systems (ALKS) and other advanced vehicle features. - **Safety Monitoring**: EU bodies are considering requiring real-time data recording and standardized testing scenarios. - **Harmonized Testing**: Member States coordinate pilot AV tests and share data, but local rules can add layers (e.g., Germany's Road Traffic Act amendments for Level 4 zones). | - **GDPR-Compliant**: All processing of personal data in AVs (e.g., driver monitoring cameras, telematics) must follow GDPR principles: consent, minimization, lawful basis for processing. - **AI Act (Proposed)**: Implementing *Regulation (EU) 2022/1426* also integrates cybersecurity requirements for AV software approval. The proposed AI Act will further detail risk-based obligations for systems that could pose "high risk" to safety. - **E-Privacy Considerations**: Data transmission from vehicles to infrastructure or OEM back-ends may be regulated under ePrivacy rules if they involve personal data. | - **Liability Under Product Liability Directive**: Manufacturers remain liable for defects in AV software/hardware. Drivers may be liable if they fail to take control when prompted. The EU is reviewing these directives to clarify fault in high-level automation scenarios. - **Motor Insurance Directive**: AVs are already covered under insurance directives as mandatory coverage for each Member States (France, Germany) have introduced or proposed clarifications about an insurer's right of recourse against the AV manufacturer if a defect caused an accident. |
| UK | - **2022**: *Law Commission Joint Report on Automated Vehicles* [469] - **2022**: *Future of Transport: Regulatory Review—National AV Data Standards (DfT)* [470] | - **Regulatory Clarity**: A new legislative framework is being shaped to clearly delineate liability and responsibilities among drivers, manufacturers, and service operators. - **Insurance Certainty**: UK's approach aims to assure that automated vehicle accidents are covered without protracted legal disputes. | - **Transition to Full Legislation**: The Law Commission's recommendations (2022) call for a new "Automated Vehicles Act" to replace or implement the *Automated and Electric Vehicles Act 2018*. - **"Listed Automated Vehicles"**: The Secretary of State can designate specific vehicle models as "self-driving," subjecting them to specialized safety requirements. - **Real-World Trials**: Ongoing pilot projects in towns and cities (e.g., Milton Keynes autonomous driverless pods and shuttles, governed by updated codes of practice). | - **Regulatory Clarity**: A new legislative framework is being shaped to clarify demarcate liability and responsibility between driver and AV system, and service operators. - **Insurance Certainty**: UK's approach aims to assure that automated vehicle accidents are covered without protracted legal disputes. | - **Proposed "No Fault" Insurance Model**: If a vehicle is in automated mode, the insurer covers accident liabilities and claims against the operator/owner, even if the AV software caused the accident. - **Civil vs. Criminal Liability**: The Law Commission (2022) recommends that the "user-in-charge" not be held criminally responsible for driving offences when the AV is in autonomous mode (unless they fail to respond to a handover request). - **Insurance Gaps**: Standard third-party motor insurance is mandatory, but specialized AV coverage is not well-defined, leading to legal ambiguity in accidents involving automated features. |
| India | - **2021**: *NITI Aayog Discussion Paper on Autonomous Vehicles & Future Mobility* [471] - **2022**: *MoRTH Draft Guidelines for ITS Deployment* [474] | - **Regulatory Clarity & Social Concerns**: Policymakers remain cautious about the impact on driver jobs, so AV technology is not aggressively promoted. - **Incremental Approach**: Priority is improving conventional road safety and infrastructure. - **Potential for Future Regulations**: NITI Aayog's proposals suggest that India may adopt a dedicated AV framework once pilot results are more conclusive. | - **Limited National AV Framework**: No dedicated AV licensing regime exists; pilot projects remain small-scale and often private or state-driven. - **ITS Focus**: The 2022 draft guidelines discuss partial autonomy and connected infrastructure, but do not create binding standards for high-level automation. - **NITI Aayog Recommendations**: Suggest establishing a regulatory sandbox for AV pilots, emphasizing scenario testing and road safety metrics. | - **Data Protection Bill Pending**: India's Personal Data Protection Bill (under parliamentary consideration) may impose data handling rules but is not yet law; AV-specific provisions are not defined. - **IT Act (2000)**: General cybersecurity clauses apply, but no AV-specific mandates. - **Draft ITS Guidelines**: Encourage data sharing among city agencies and private operators but do not specify robust cybersecurity frameworks for AV software. | - **Existing MV Act Usage**: The Motor Vehicles (Amendment) Act 2019 does not explicitly address autonomous vehicles, so liability defaults to drivers/operators or owners. - **Uncertain Manufacturer Liability**: In the absence of specific legislation, any claims against system developers must rely on general product liability and consumer protection laws. - **Insurance Gaps**: Standard motor insurance is mandatory, but specialized third-party motor insurance is not well-defined; AV trial applicants must prove they have adequate insurance coverage for potential damages. |
| Australia | - **2022**: *NTC National Enforcement Guidelines for Automated Vehicles in C-ITS & AV Data* [475] - **2023**: *NTC Regulating Government Access to C-ITS & AV Data—Final Recommendations* [476] | - **Collaborative Federalism**: The NTC attempts to harmonize states and territories to create a cohesive national market for AVs. - **Real-World AV Pilots**: Several state-based trials protocols. - **Focus on Data Governance**: Recent NTC recommendations aim to balance road safety benefits with privacy protections. | - **Trial-Focused**: States and territories follow the *NTC Guidelines for Trials of Automated Vehicles* (first published 2017, updated periodically). - **Enforcement Guidelines (2022)**: Clarify how police and road authorities should manage AV incidents for vehicles with different levels of automation. - **Working Toward National Law**: The NTC is developing a Uniform Regulatory Framework for commercial deployment of higher-level AVs, aiming to avoid inconsistent state rules. | - **Privacy Act 1988 (Cth)**: Governs personal data, but no dedicated AV privacy rules exist yet. - **NTC 2023 Recommendations**: Suggest limiting government access to AV/C-ITS data unless necessary for road safety, law enforcement, or public interest. - **Cybersecurity Discussions**: Australia is reviewing global standards (UN R155) for cybersecurity type approval, with potential future adoption in the national framework. | - **Existing Insurance Schemes**: Australia's compulsory third-party (CTP) motor injury schemes differ by state. Some (e.g., Victoria) are exploring how to handle accident claims. - **Liability Shifts/Roles**: The NTC's policy work (2022–2023) suggests adopting a "Automated Driving System Entity" concept, making the system provider liable for system-caused incidents. - **Trial Requirements**: AV trial applicants must prove they have adequate insurance coverage for potential damages. |

fear of curtailing that development or being left with regulations that fail to meet the capacity of technology. According to Mateo Sanguino et al. [461] increased investments in fighting these forms of cybercrime are required alongside improved collaboration between manufacturers, regulators, and researchers.

### 11.5. Conclusions and the Way Forward

The current capabilities of PAVs must balance automation with human inputs. Similarly, striking the right balance between safety and security, on the one hand, and innovation and cost, on the other, is essential when it comes to addressing policy and regulatory challenges in the sector in ways that will permit the successful commercialization of the technology and the realization of its potential. This section has reviewed both the overarching issues that present policy challenges in the PAV field (such as the speed of technological development and the need for regulatory harmonization across jurisdictions) and the three main specific policy challenges: safety and licensing, liability and insurance, and data privacy and cybersecurity. In addition to presenting the challenges in each area, solutions that are already being applied in some jurisdictions have also been explored. However, even the most developed solutions still tend to be relatively nascent and do not fully address the safety, liability, and cybersecurity risks that are likely to intensify as the technology develops further and moves from testing to large-scale commercial deployment. Solving those challenges requires policymakers to adopt approaches that address both the overarching challenges and the specific policy issues.

In terms of the overarching challenges, policymakers need to look for ways to harmonize regulations without stifling innovation or trying to impose one-size-fits-all solutions in places where they do not fit. One way of approaching that would be to establish an authoritative body that can coordinate the work of various government agencies, such as the UK's Centre for Connected and Autonomous Vehicles and Singapore's Committee on Autonomous Road Transport [456]. Further, any regulations that are adopted should have sufficient flexibility to enable changes in response to technological developments. In terms of specific challenges, policymakers should recognize that stringent safety controls are not the enemy of the spread of PAVs, in fact, they will contribute to it by helping to gain public trust which is essential if the technology is to be widely adopted [448]. Regulations on safety, liability, and insurance will work best when linked to existing legislation and industry standards but with amendments that recognize the unique circumstances of PAVs. Such approaches can be informed by previous studies in this area, such as Australia's National Transport Committee's Regulation Impact Statement process [456].

Finally, the challenges of policymaking in this area are best addressed by collaborative efforts, not only within and across jurisdictions but also through the involvement of industry, consumer groups, and researchers who can identify current solutions and predict the potential future challenges that contemporary policies must take into account. Such collaboration is the best way to ensure that the positive potential of PAVs is realized through the effective and safe commercialization and deployment of the technology.

## 12. Conclusive remarks: The Long Road Ahead

Due to the continuously changing and still partly undefined direction of the automotive industry world-wide, we still cannot provide definitive answers to the five research questions (Q1-Q5) formulated at the beginning of section 1.2. However, our review puts us in a better position to address those questions, having established the preconditions for formulating high-level evidence-based considerations of a systematic and strategic nature.

In addressing Q2, the review has highlighted that the most persistent obstacles faced by developers and makers cannot be solved with a 'techno solutionist' approach because they rarely are of an exclusively technological nature: while sensor technologies and machine-learning techniques become more powerful, robust, and reliable, the need to make appropriate decisions perfectly tailored to very specific, diverse, fluid, and complex sets of circumstances does not disappear but, on the contrary, becomes a growing challenge for PAVs, which need to deploy expert protocols not only to recognize and categorize different possible scenarios (including the most unlikely ones), but also to adaptively deal with them or avoid them altogether, anticipating the other vehicles' behaviour and predicting the consequences of their own behaviour so to reduce risks and prevent situations that could limit their mobility or present them with significant liabilities or impossible ethical decisions.

In addressing Q3, the review suggests that the greatest challenge associated with PAVs is the ambition to support, complement or even recreate and replace human intelligence, for example reproducing distinctively human capabilities like commonsense reasoning, situational awareness, and foresight: if PAVs need making decisions informed by a human-like understanding of their consequences then even a very simple task, if truly "intelligent", needs to carry the burden of the infinite interpretations of human experience and judgement. Unlike other problems and issues that appear localized and appear solvable with a mechanistic approach, the challenges related to intelligence and skill seem structural in nature (deeply rooted in the essential demands and expectations bestowed on autonomous technologies), extraordinarily persistent (requiring a long-term perspective) and requiring a holistic approach (an understanding of PAVs as integrated whole).

In addressing Q4, the review has shown the largely interconnected nature of the problems and issues affecting PAVs, as it has revealed unobvious links between technological limitations, human-centric challenges, and normative (ethical, legal) concerns: this suggests that, while the problems and issues affecting PAVs undoubtedly require specialist expertise to be understood and addressed correctly, they cannot be treated in isolation with the classical "silo" approach. As they are not irrelated, self-contained puzzles, problems can be recognized and addressed only if they are properly contextualized within a holistic framework that allows to appreciate their mutual dependencies with other issues and problems. Delineating the foundational or methodological premises of such a framework is beyond this review's scope.

However, we can confidently argue that such a framework has to be informed by human expertise and judgment, and comprehensive and flexible enough to treat systemically, as part of an interconnected whole, all the issues and problems reviewed so far, despite their different nature and origin: technological limitations, implementational defects, conceptualization & design errors, human-centric specificities and preferences, institutional and legal requirements, societal expectations, economic viability. It is worth mentioning that second order cybernetics and 4-E cognition, combined with human factors & ergonomics, have already provided the general theoretical background to establish holistic and experientially aware frameworks of this kind to orient research & development in the fields of autonomy.

In addressing Q5, the review has highlighted a great variety of approaches and best practices tailored to address both the specificities and the interconnectedness of the issues and problems considered. This justifies different expectations and hopes towards different categories of problems in the short or rather in the long term by means of focalized interventions or more sweeping strategies. On the one hand, many of the local issues that still affect PAVs are likely to be significantly mitigated or entirely solved in the short or medium period, based on the ongoing progress being made in the relevant areas and thanks to the availability of multiple solutions that are mechanistically identifiable, technologically feasible, and economically viable without requiring any major perspective or paradigm change (for example the algorithms for managing intersection traffic and for optimizing the coordination of multiple connected PAVs).

On the other hand, some of the problems discussed in this review could not be solved by a simple fix or some additional refinement of existing technologies, because they require either a global change in perspective/approach (for example building a new sensitivity and attitude towards PAVs by the investors, adopters, final users, and the societies at large) or an ideational effort to advance a ground-breaking techno-scientific framework (like the one required to produce AGI and overcome the theoretical challenges of commonsense reasoning). Finally, a solution to the most structural, long-term problems may require some radical decisions and ambitious top-down interventions from the leaderships of our countries, decisions involving large-scale coordinated efforts by various sectors of the society, massive investments, a courageous vision, and firm governance (for example, long-term plans to systematically redesign road infrastructure and to deeply reconfigure the economy underlying the automative and transportation industries). Such energetic interventions would be particularly necessary if it was proven that the global markets do not always spontaneously promote innovation and that technological progress is not a destiny that necessarily tends to fulfill itself but an uncertain outcome that requires coordinated efforts and risky decisions.

Taken together, these considerations suggest what needs to be done to make PAVs real and accessible to anybody, thus providing a still tentative but consistent strategic to answer to Q1. However, the concrete significance and practical meaning of the suggestions presented here strongly depends on the stakes associated with PAVs and the ambitions or needs that we intend to fulfill by means of autonomous cars. (i) If one thinks of PAVs as instruments to achieve a utilitarian benefit, such as making mobility more efficient and safer, then the steps to be done concern primarily the resolution of uncertainties concerning current sensing and navigation technologies to achieve a human-like or better-than-human level of performance. (ii) If one thinks of PAVs as a commercial product that looks more or less appealing to consumers based on its trustworthiness, perceived intelligence, and comfort, then the steps to be done will focus on the human-factors and user's attitudes to match the product's qualities with the customers' expectations. (iii) If one thinks of PAVs as an epochal step towards the redefinition of logistics and mobility in our civilization, with complex political implications that reach to the emancipation of individuals or the engineering of socio-cultural aspects of the collective life, then the priority would be the construction of an extensive, cohesive infrastructure – physical and normative at once – that sustains the coming of PAVs while assigning to it a precise socio-economical and cultural role.

# References

[1] P. Viswanath, K. Chitnis, P. Swami, M. Mody, S. Shivalingappa, S. Nagori, M. Mathew, K. Desappan, S. Jagannathan, D. Poddar et al., "A diverse low cost high performance platform for advanced driver assistance system (ADAS) applications," in Proc. CVPR Workshops, 2016, pp. 1–9.

[2] E. Yurtsever, J. Lambert, A. Carballo, and K. Takeda, "A survey of autonomous driving: Common practices and emerging technologies," IEEE Access, vol. 8, pp. 58 443–58 469, 2020.

[3] Y. Li and J. Ibanez-Guzman, "LiDAR for autonomous driving: The principles, challenges, and trends for automotive LiDAR and perception systems," IEEE Signal Process. Mag., vol. 37, no. 4, pp. 50–61, 2020.

[4] K. Muhammad, A. Ullah, J. Lloret, J. Del Ser, and V. H. C. de Albuquerque, "Deep learning for safe autonomous driving: Current challenges and future directions," IEEE Trans. Intell. Transp. Syst., vol. 22, no. 7, pp. 4316–4336, 2020.

[5] J. Nidamanuri, C. Nibhanupudi, R. Assfalg, and H. Venkataraman, "A progressive review: Emerging technologies for ADAS driven solu- tions," IEEE Trans. Intell. Veh., vol. 7, no. 2, pp. 326–341, 2021.

[6] F. Manfio Barbosa and F. Santos Oso´rio, "Camera-radar perception for autonomous vehicles and ADAS: Concepts, datasets and metrics," arXiv e-prints, pp. arXiv–2303, 2023.

[7] T. S. of Automotive Engineers (SAE), "Taxonomy and definitions for terms related to driving automation systems for on-road motor vehicles," https://www.sae.org/standards/content/j3016 202104/.

[8] E. Arnold, O. Y. Al-Jarrah, M. Dianati, S. Fallah, D. Oxtoby, and A. Mouzakitis, "A survey on 3D object detection methods for autonomous driving applications," IEEE Trans. Intell. Transp. Syst., vol. 20, no. 10, pp. 3782–3795, 2019.

[9] Y. Zhang, Z. Lu, X. Zhang, J.-H. Xue, and Q. Liao, "Deep learning in lane marking detection: A survey," IEEE Trans. Intell. Transp. Syst., vol. 23, no. 7, pp. 5976–5992, 2021.

[10] S. Sivaraman and M. M. Trivedi, "Looking at vehicles on the road: A survey of vision-based vehicle detection, tracking, and behavior analysis," IEEE Trans. Intell. Transp. Syst., vol. 14, no. 4, pp. 1773– 1795, 2013.

[11] A. Mukhtar, L. Xia, and T. B. Tang, "Vehicle detection techniques for collision avoidance systems: A review," IEEE Trans. Intell. Transp. Syst., vol. 16, no. 5, pp. 2318–2338, 2015.

[12] Z. Wang, J. Zhan, C. Duan, X. Guan, P. Lu, and K. Yang, "A review of vehicle detection techniques for intelligent vehicles," IEEE Trans. Neural Netw. Learn. Syst., 2022.

[13] B. Ranft and C. Stiller, "The role of machine vision for intelligent vehicles," IEEE Trans. Intell. Veh., vol. 1, no. 1, pp. 8–19, 2016.

[14] A. Bar Hillel, R. Lerner, D. Levi, and G. Raz, "Recent progress in road and lane detection: a survey," Mach. Vis. Appl., vol. 25, no. 3, pp. 727–745, 2014.

[15] H. Yin and C. Berger, "When to use what data set for your self-driving car algorithm: An overview of publicly available driving datasets," in Proc. ITSC. IEEE, 2017, pp. 1–8.

[16] J. Van Brummelen, M. O'Brien, D. Gruyer, and H. Najjaran, "Autonomous vehicle perception: The technology of today and tomorrow," Transp. Res. Part C Emerg., vol. 89, pp. 384–406, 2018.

[17] N. Ragesh and R. Rajesh, "Pedestrian detection in automotive safety: Understanding state-of-the-art," IEEE Access, vol. 7, pp. 47 864– 47 890, 2019.

[18] S. Grigorescu, B. Trasnea, T. Cocias, and G. Macesanu, "A survey of deep learning techniques for autonomous driving," J. Field Robot., vol. 37, no. 3, pp. 362–386, 2020.

[19] L. Chen, S. Lin, X. Lu, D. Cao, H. Wu, C. Guo, C. Liu, and F.-Y. Wang, "Deep neural network based vehicle and pedestrian detection for autonomous driving: A survey," IEEE Trans. Intell. Transp. Syst., vol. 22, no. 6, pp. 3234–3246, 2021.

[20] J. E. Espinosa, S. A. Velast´ın, and J. W. Branch, "Detection of motorcycles in urban traffic using video analysis: A review," IEEE Trans. Intell. Transp. Syst., vol. 22, no. 10, pp. 6115–6130, 2020.

[21] Wiki, "Ieee xplore," https://en.wikipedia.org/wiki/IEEE Xplore.

[22] A. G. Howard, M. Zhu, B. Chen, D. Kalenichenko, W. Wang, T. Weyand, M. Andreetto, and H. Adam, "MobileNets: Efficient convolutional neural networks for mobile vision applications," arXiv preprint arXiv:1704.04861, 2017.

[23] I. Alhashim and P. Wonka, "High quality monocular depth estimation via transfer learning," arXiv preprint arXiv:1812.11941, 2018.

[24] J. Redmon and A. Farhadi, "YOLOv3: An incremental improvement," arXiv preprint arXiv:1804.02767, 2018.

[25] F. Yu, W. Xian, Y. Chen, F. Liu, M. Liao, V. Madhavan, and T. Darrell, "BDD100K: A diverse driving video database with scalable annotation tooling," arXiv preprint arXiv:1805.04687, vol. 2, no. 5, p. 6, 2018.

[26] Z. Che, G. Li, T. Li, B. Jiang, X. Shi, X. Zhang, Y. Lu, G. Wu, Y. Liu, and J. Ye, "D²-City: A large-scale dashcam video dataset of diverse traffic scenarios," arXiv preprint arXiv:1904.01975, 2019.

[27] A. Bochkovskiy, C.-Y. Wang, and H.-Y. M. Liao, "YOLOv4: Optimal speed and accuracy of object detection," arXiv preprint arXiv:2004.10934, 2020.

[28] J. Geyer, Y. Kassahun, M. Mahmudi, X. Ricou, R. Durgesh, A. S. Chung, L. Hauswald, V. H. Pham, M. Mu¨hlegg, S. Dorn et al., "A2D2: Audi autonomous driving dataset," arXiv preprint arXiv:2004.06320, 2020.

[29] L. Wang, M. Famouri, and A. Wong, "DepthNet Nano: A highly com- pact self-normalizing neural network for monocular depth estimation," arXiv preprint arXiv:2004.08008, 2020.

[30] L. Ewecker, L. Ohnemus, R. Schwager, S. Roos, and S. Sarala- jew, "Combining visual saliency methods and sparse keypoint an- notations to providently detect vehicles at night," arXiv preprint arXiv:2204.11535, 2022.

[31] H. Jayarathne, T. Samarakoon, H. Koralege, A. Divisekara, R. Ro- drigo, and P. Jayasekara, "DualCam: A novel benchmark dataset for fine-grained real-time traffic light detection," arXiv preprint arXiv:2209.01357, 2022.

[32] M. Alibeigi, W. Ljungbergh, A. Tonderski, G. Hess, A. Lilja, C. Lind- strom, D. Motorniuk, J. Fu, J. Widahl, and C. Petersson, "Zenseact Open Dataset: A large-scale and diverse multimodal dataset for au- tonomous driving," arXiv preprint arXiv:2305.02008, 2023.

[33] R. D. Staff, "NavLab: The self-driving car of the '80s," https://www. rediscoverthe80s.com/2016/11/navlab-the-selfdriving-car-of-the-80s. html, 2016.

[34] E. D. Dickmanns, Dynamic vision for perception and control of motion. Springer Science & Business Media, 2007.

[35] Wiki, "Tesla, Inc." https://en.wikipedia.org/wiki/Tesla, Inc.

[36] M. Joire, "We test out the hands-free Mercedes-Benz Drive Pilot level 3 system," https://www.motor1.com/news/656719/ mercedes-drive-pilot-level-3/.

[37] J. S. Choksey, "What is Volvo Ride Pilot?" https://www.jdpower.com/ cars/shopping-guides/what-is-volvo-ride-pilot.

[38] AutoMuse, "Audi can't upgrade A8 to level 3 au- tonomous driving," https://www.automuse.co.nz/news/ audi-cant-upgrade-a8-to-level-3-autonomous-driving, 2023.

[39] Audi, "Audi driver assistance systems overview," https://www. audibellevue.com/research/audi-driver-assistance.htm.

[40] Subaru, "Subaru's preventative safety: How does it work?" https:// www.quantrellsubaru.com/subaru-eyesight-system/.

[41] Audi, "Audi driver assistance systems overview," https:// audibellevue.com/research/audi-driver-assistance.htm.

[42] BMW, "Overview of the main driver assistance systems," https://www. bmw.com/en/innovation/the-main-driver-assistance-systems.html.

[43] Fiat, "Advanced safety systems for full driving control," https://www. fiatcamper.com/en/product/safety.

[44] Ford, "Ford Co-Pilot360 Technology," https://www.ford.com/ technology/driver-assist-technology/#1.

[45] Honda, "Honda sensing," https://www.honda.com.au/buy/safety/ honda-sensing.

[46] Hyundai, "Hyundai SmartSense our network of advanced safety and convenienbc tech," https://www.hyundaiusa.com/us/en/safety.

[47] Kia, "Kia Drive Wise" advanced driver assistance systems," https:// www.kiaworldcar.com/wc-kia-drive-wise-adas/.

[48] L. Rover, "Advanced driver assistance systems (adas)," https://www. landrover.com/ownership/incontrol/driver-assistance.html.

[49] Lexus, "LEXUS SAFETY SYSTEM+ features, operation, limi- tations and precautions," https://www.lexus.com/content/dam/lexus/ documents/safety/2023-LSS-Document-Final.pdf.

[50] MAZDA, "I-ACTIVSENSE Safety technologies to support your safe driving," https://www.mazda.com/en/archives/safety2/i-activsense/.

[51] Mercedes-Benz, "The front runner in automated driving and safety technologies," https://group.mercedes-benz.com/innovation/case/ autonomous/drive-pilot-2.html.

[52] L. Kim, "What is Mercedes-Benz Drive Pilot?" https://www.jdpower. com/cars/shopping-guides/what-is-mercedes-benz-drive-pilot.

[53] Mitsubishi, "MiTEC Mitsubishi Motors Intuitive Technology," https: //www.mitsubishi-motors.com.au/buying-tools/mitec.html.

[54] Nissian, "ProPILOT," https://www.nissan-global.com/EN/ INNOVATION/TECHNOLOGY/ARCHIVE/PROPILOT/.

[55] ——, "ProPILOT 2.0," https://www.nissan-global.com/EN/ INNOVATION/TECHNOLOGY/ARCHIVE/AD2/.

[56] Subaru, "Subaru Eyesight System: How does it work?" https://www. subaru.com/subaru-eyesight-system/.

[57] Tesla, "Autopilot and full self-driving capability," https://www.tesla. com/en au/support/autopilot.

[58] TOYOTA, "Toyota safety sense," https://www.toyota.com.au/ toyota-safety-sense.

[59] Volkswagen, "IQ.DRIVE our suite of advanced driver assistance sys- tems," https://www.volkswagen.com.au/en/technology/safety/iq-drive. html.

[60] E. Marti, M. A. De Miguel, F. Garcia, and J. Perez, "A review of sensor technologies for perception in automated driving," IEEE Intell. Transp. Syst. Mag., vol. 11, no. 4, pp. 94–108, 2019.

[61] B. Krzanich, "Data is the new oil in the future of automated driving," in Intel Newsroom, 2016.

[62] Y. Zhang, A. Carballo, H. Yang, and K. Takeda, "Perception and sensing for autonomous vehicles under adverse weather conditions: A survey," ISPRS J. Photogramm. Remote Sens., vol. 196, pp. 146– 177, 2023.

[63] L. Liu, S. Lu, R. Zhong, B. Wu, Y. Yao, Q. Zhang, and W. Shi, "Computing systems for autonomous driving: State of the art and challenges," IEEE Internet Things J., vol. 8, no. 8, pp. 6469–6486, 2020.

[64] A. Geiger, P. Lenz, C. Stiller, and R. Urtasun, "Vision meets robotics: The KITTI dataset," Int. J. Robot. Res., vol. 32, no. 11, pp. 1231–1237, 2013.

[65] M. B. Jensen, M. P. Philipsen, A. Møgelmose, T. B. Moeslund, and M. M. Trivedi, "Vision for looking at traffic lights: Issues, survey, and perspectives," IEEE Trans. Intell. Transp. Syst., vol. 17, no. 7, pp. 1800– 1815, 2016.

[66] Z. Zhu, D. Liang, S. Zhang, X. Huang, B. Li, and S. Hu, "Traffic-sign detection and classification in the wild," in Proc. CVPR, 2016, pp. 2110– 2118.

[67] K. Behrendt, L. Novak, and R. Botros, "A deep learning approach to traffic lights: Detection, tracking, and classification," in Proc. ICRA. IEEE, 2017, pp. 1370–1377.

[68] Y. Choi, N. Kim, S. Hwang, K. Park, J. S. Yoon, K. An, and I. S. Kweon, "KAIST multi-spectral day/night data set for autonomous and assisted driving," IEEE Trans. Intell. Transp. Syst., vol. 19, no. 3, pp. 934–948, 2018.

[69] L. Neumann, M. Karg, S. Zhang, C. Scharfenberger, E. Piegert, S. Mistr, O. Prokofyeva, R. Thiel, A. Vedaldi, A. Zisserman et al., "NightOwls: A pedestrians at night dataset," in Proc. ACCV, 2019, pp. 691–705.

[70] Y. Feng, D. Kong, P. Wei, H. Sun, and N. Zheng, "A benchmark dataset and multi-scale attention network for semantic traffic light detection," in Proc. ITSC. IEEE, 2019, pp. 1–8.

[71] Q.-H. Pham, P. Sevestre, R. S. Pahwa, H. Zhan, C. H. Pang, Y. Chen, A. Mustafa, V. Chandrasekhar, and J. Lin, "A 3D dataset: Towards autonomous driving in challenging environments," in Proc. ICRA. IEEE, 2020, pp. 2267–2273.

[72] C. Ertler, J. Mislej, T. Ollmann, L. Porzi, G. Neuhold, and Y. Kuang, "The mapillary traffic sign dataset for detection and classification on a global scale," in Proc. ECCV, 2020, pp. 68–84.

[73] W. Zhou, J. S. Berrio, C. De Alvis, M. Shan, S. Worrall, J. Ward, and E. Nebot, "Developing and testing robust autonomy: The University of Sydney campus data set," IEEE Intell. Transp. Syst. Mag., vol. 12, no. 4, pp. 23–40, 2020.

[74] S. Saralajew, L. Ohnemus, L. Ewecker, E. Asan, S. Isele, and S. Roos, "A dataset for provident vehicle detection at night," in Proc. IROS. IEEE, 2021, pp. 9750–9757.

[75] X. Zhang, Z. Li, Y. Gong, D. Jin, J. Li, L. Wang, Y. Zhu, and H. Liu, "OpenMPD: An open multimodal perception dataset for autonomous driving," IEEE Trans. Veh. Technol., vol. 71, no. 3, pp. 2437–2447, 2022.

[76] O. Jayasinghe, S. Hemachandra, D. Anhettigama, S. Kariyawasam, T. Wickremasinghe, C. Ekanayake, R. Rodrigo, and P. Jayasekara, "Towards real-time traffic sign and traffic light detection on embedded systems," in Proc. IV. IEEE, 2022, pp. 723–728.

[77] Y. Liao, J. Xie, and A. Geiger, "KITTI-360: A novel dataset and benchmarks for urban scene understanding in 2D and 3D," IEEE Trans. Pattern Anal. Mach. Intell., 2022.

[78] D.-H. Paek, S.-H. Kong, and K. T. Wijaya, "K-Lane: Lidar lane dataset and benchmark for urban roads and highways," in Proc. CVPR, 2022, pp. 4450–4459.

[79] X. Yang, J. Yan, W. Liao, X. Yang, J. Tang, and T. He, "SCRDet++: Detecting small, cluttered and rotated objects via instance-level feature denoising and rotation loss smoothing," IEEE Trans. Pattern Anal. Mach. Intell., vol. 45, no. 2, pp. 2384–2399, 2022.

[80] X. Dong, M. A. Garratt, S. G. Anavatti, and H. A. Abbass, "Towards real-time monocular depth estimation for robotics: A survey," IEEE Trans. Intell. Transp. Syst., vol. 23, no. 10, pp. 16 940–16 961, 2022.

[81] H. Laga, L. V. Jospin, F. Boussaid, and M. Bennamoun, "A survey on deep learning techniques for stereo-based depth estimation," IEEE Trans. Pattern Anal. Mach. Intell., vol. 44, no. 4, pp. 1738–1764, 2020.

[82] K. Wang and S. Shen, "MVDepthNet: Real-time multiview depth estimation neural network," in Proc. Int. Conf. 3D Vis. (3DV). IEEE, 2018, pp. 248–257.

[83] F. Tosi, F. Aleotti, M. Poggi, and S. Mattoccia, "Learning monocular depth estimation infusing traditional stereo knowledge," in Proc. CVPR, 2019, pp. 9799–9809.

[84] Y. Wang, Z. Lai, G. Huang, B. H. Wang, L. Van Der Maaten, M. Campbell, and K. Q. Weinberger, "Anytime stereo image depth estimation on mobile devices," in Proc. ICRA. IEEE, 2019, pp. 5893–5900.

[85] Z. Li, X. Liu, N. Drenkow, A. Ding, F. X. Creighton, R. H. Taylor, and M. Unberath, "Revisiting stereo depth estimation from a sequence-to-sequence perspective with transformers," in Proc. ICCV, 2021, pp. 6197–6206.

[86] Y. LeCun, Y. Bengio, and G. Hinton, "Deep learning," Nature, vol. 521, no. 7553, pp. 436–444, 2015.

[87] D. Eigen, C. Puhrsch, and R. Fergus, "Depth map prediction from a single image using a multi-scale deep network," Proc. NIPS, vol. 27, 2014.

[88] S. F. Bhat, I. Alhashim, and P. Wonka, "AdaBins: Depth estimation using adaptive bins," in Proc. CVPR, 2021, pp. 4009–4018.

[89] D. Wofk, F. Ma, T.-J. Yang, S. Karaman, and V. Sze, "FastDepth: Fast monocular depth estimation on embedded systems," in Proc. ICRA. IEEE, 2019, pp. 6101–6108.

[90] X. Dong, M. A. Garratt, S. G. Anavatti, and H. A. Abbass, "Mo-bileXNet: An efficient convolutional neural network for monocular depth

[91] X. Dong, M. A. Garratt, S. G. Anavatti, H. A. Abbass, and J. Dong, "Lightweight monocular depth estimation with an edge guided net- work," in Proc. ICARCV, 2022, pp. 204–210.

[92] L. Liu, W. Ouyang, X. Wang, P. Fieguth, J. Chen, X. Liu, and M. Pietika¨inen, "Deep learning for generic object detection: A survey," Int. J. Comput. Vis., vol. 128, pp. 261–318, 2020.

[93] R. Girshick, J. Donahue, T. Darrell, and J. Malik, "Rich feature hierarchies for accurate object detection and semantic segmentation," in Proc. CVPR, 2014, pp. 580–587.

[94] R. Girshick, "Fast R-CNN," in Proc. ICCV, 2015, pp. 1440–1448.

[95] S. Ren, K. He, R. Girshick, and J. Sun, "Faster R-CNN: Towards real-time object detection with region proposal networks," Proc. NIPS, vol. 28, 2015.

[96] J. R. Uijlings, K. E. Van De Sande, T. Gevers, and A. W. Smeulders, "Selective search for object recognition," Int. J. Comput. Vis., vol. 104, pp. 154–171, 2013.

[97] K. He, X. Zhang, S. Ren, and J. Sun, "Spatial pyramid pooling in deep convolutional networks for visual recognition," IEEE Trans. Pattern Anal. Mach. Intell., vol. 37, no. 9, pp. 1904–1916, 2015.

[98] J. Redmon, S. Divvala, R. Girshick, and A. Farhadi, "You only look once: Unified, real-time object detection," in Proc. CVPR, 2016, pp. 779–788.

[99] W. Liu, D. Anguelov, D. Erhan, C. Szegedy, S. Reed, C.-Y. Fu, and A. C. Berg, "SSD: Single shot multibox detector," in Proc. ECCV. Springer, 2016, pp. 21–37.

[100] J. Redmon and A. Farhadi, "YOLO9000: better, faster, stronger," in Proc. CVPR, 2017, pp. 7263–7271.

[101] C.-Y. Wang, A. Bochkovskiy, and H.-Y. M. Liao, "Yolov7: Trainable bag-of-freebies sets new state-of-the-art for real-time object detectors," in Proc. CVPR, 2023, pp. 7464–7475.

[102] Z. Zou, K. Chen, Z. Shi, Y. Guo, and J. Ye, "Object detection in 20 years: A survey," Proc. IEEE, 2023.

[103] J. Zhang, L. Lin, J. Zhu, Y. Li, Y.-c. Chen, Y. Hu, and S. C. Hoi, "Attribute-aware pedestrian detection in a crowd," IEEE Trans. Multimedia., vol. 23, pp. 3085–3097, 2020.

[104] Wikipedia, "Automotive night vision," https://en.wikipedia.org/wiki/ Automotive night vision.

[105] N. Dalal and B. Triggs, "Histograms of oriented gradients for human detection," in Proc. CVPR, vol. 1. IEEE, 2005, pp. 886–893.

[106] S. Zhang, J. Yang, and B. Schiele, "Occluded pedestrian detection through guided attention in CNNs," in Proc. CVPR, 2018, pp. 6995–7003.

[107] G. Li, Y. Yang, and X. Qu, "Deep learning approaches on pedestrian detection in hazy weather," IEEE Trans. Ind. Electron., vol. 67, no. 10, pp. 8889–8899, 2019.

[108] A. Nowosielski, K. Małecki, P. Forczman´ski, A. Smolin´ski, and K. Krzywicki, "Embedded night-vision system for pedestrian detection," IEEE Sens., vol. 20, no. 16, pp. 9293–9304, 2020.

[109] J. U. Kim, S. Park, and Y. M. Ro, "Uncertainty-guided cross-modal learning for robust multispectral pedestrian detection," IEEE Trans. Circuits Syst., vol. 32, no. 3, pp. 1510–1523, 2021.

[110] K. Dasgupta, A. Das, S. Das, U. Bhattacharya, and S. Yogamani, "Spatio-contextual deep network-based multimodal pedestrian detection for autonomous driving," IEEE Trans. Intell. Transp. Syst., vol. 23, no. 9, pp. 15 940–15 950, 2022.

[111] W. Tian and M. Lauer, "Fast cyclist detection by cascaded detector and geometric constraint," in Proc. ITSC, 2015, pp. 1286–1291.

[112] X. Li, F. Flohr, Y. Yang, H. Xiong, M. Braun, S. Pan, K. Li, and D. M. Gavrila, "A new benchmark for vision-based cyclist detection," in Proc. IV. IEEE, 2016, pp. 1028–1033.

[113] X. Li, L. Li, F. Flohr, J. Wang, H. Xiong, M. Bernhard, S. Pan, D. M. Gavrila, and K. Li, "A unified framework for concurrent pedestrian and cyclist detection," IEEE Trans. Intell. Transp. Syst., vol. 18, no. 2, pp. 269–281, 2016.

[114] K. Wang and W. Zhou, "Pedestrian and cyclist detection based on deep neural network fast R-CNN," International Journal of Advanced Robotic Systems, vol. 16, no. 2, p. 1729881419829651, 2019.

[115] N. Annapareddy, E. Sahin, S. Abraham, M. M. Islam, M. DePiro, and T. Iqbal, "A robust pedestrian and cyclist detection method using thermal images," in 2021 Systems and Information Engineering Design Symposium (SIEDS). IEEE, 2021, pp. 1–6.

[116] F. Garcia, D. Martin, A. De La Escalera, and J. M. Armingol, "Sensor fusion methodology for vehicle detection," IEEE Intell. Transp. Syst. Mag., vol. 9, no. 1, pp. 123–133, 2017.

[117] Z. Yang, J. Li, and H. Li, "Real-time pedestrian and vehicle detection for autonomous driving," in Proc. IV, 2018, pp. 179–184.

[118] H. Wang, Y. Yu, Y. Cai, X. Chen, L. Chen, and Q. Liu, "A com- parative study of state-of-the-art deep learning algorithms for vehicle detection," IEEE Intell. Transp. Syst. Mag., vol. 11, no. 2, pp. 82–95, 2019.

[119] J. Dai, Y. Li, K. He, and J. Sun, "R-FCN: Object detection via region-based fully convolutional networks," Proc. NIPS, vol. 29, 2016.

[120] T.-Y. Lin, P. Goyal, R. Girshick, K. He, and P. Dolla´r, "Focal loss for dense object detection," in Proc. ICCV, 2017, pp. 2980–2988.

[121] B. Wu, F. Iandola, P. H. Jin, and K. Keutzer, "SqueezeDet: Unified, small, low power fully convolutional neural networks for real-time object detection for autonomous driving," in Proc. CVPR Workshops, 2017, pp. 129–137.

[122] L. Chen, Q. Ding, Q. Zou, Z. Chen, and L. Li, "DenseLightNet: A light-weight vehicle detection network for autonomous driving," IEEE Trans. Ind. Electron., vol. 67, no. 12, pp. 10 600–10 609, 2020.

[123] C. B. Murthy, M. F. Hashmi, and A. G. Keskar, "EfficientLiteDet: a real-time pedestrian and vehicle detection algorithm," Mach. Vis. Appl., vol. 33, no. 3, p. 47, 2022.

[124] L. Ewecker, E. Asan, L. Ohnemus, and S. Saralajew, "Provident vehicle detection at night for advanced driver assistance systems," Auton. Robots, vol. 47, no. 3, pp. 313–335, 2023.

[125] E. Oldenziel, L. Ohnemus, and S. Saralajew, "Provident detection of vehicles at night," in Proc. IV, 2020, pp. 472–479.

[126] L. Ewecker, E. Asan, and S. Roos, "Detecting vehicles in the dark in urban environments-a human benchmark," in Proc. IV, 2022, pp. 1145–1151.

[127] P. Sevekar and S. Dhonde, "Nighttime vehicle detection for intelligent headlight control: A review," in 2016 2nd International Conference on Applied and Theoretical Computing and Communication Technology (iCATccT). IEEE, 2016, pp. 188–190.

[128] J. Tang, S. Li, and P. Liu, "A review of lane detection methods based on deep learning," Pattern Recognit., vol. 111, p. 107623, 2021.

[129] J. Kim and M. Lee, "Robust lane detection based on convolutional neural network and random sample consensus," in Neural Information Processing: 21st International Conference, ICONIP 2014, Kuching, Malaysia, November 3-6, 2014. Proceedings, Part I 21. Springer, 2014, pp. 454–461.

[130] A. Gurghian, T. Koduri, S. V. Bailur, K. J. Carey, and V. N. Murali, "DeepLanes: End-to-end lane position estimation using deep neural networksa," in Proc. CVPR workshops, 2016, pp. 38–45.

[131] D. Neven, B. De Brabandere, S. Georgoulis, M. Proesmans, and L. Van Gool, "Towards end-to-end lane detection: an instance seg-mentation approach," in Proc. IV, 2018, pp. 286–291.

[132] L. Tabelini, R. Berriel, T. M. Paixao, C. Badue, A. F. De Souza, and T. Oliveira-Santos, "Keep your eyes on the lane: Real-time attention-guided lane detection," in Proc. CVPR, 2021, pp. 294–302.

[133] X. Li, J. Li, X. Hu, and J. Yang, "Line-CNN: End-to-end traffic line detection with line proposal unit," IEEE Trans. Intell. Transp. Syst., vol. 21, no. 1, pp. 248–258, 2019.

[134] A. Hata and D. Wolf, "Road marking detection using LIDAR reflective intensity data and its application to vehicle localization," in Proc. ITSC. IEEE, 2014, pp. 584–589.

[135] L. Caltagirone, S. Scheidegger, L. Svensson, and M. Wahde, "Fast LIDAR-based road detection using fully convolutional neural net- works," in Proc. IV. IEEE, 2017, pp. 1019–1024.

[136] M. Bai, G. Mattyus, N. Homayounfar, S. Wang, S. K. Lakshmikanth, and R. Urtasun, "Deep multi-sensor lane detection," in Proc. IROS. IEEE, 2018, pp. 3102–3109.

[137] X. Zhang, Z. Li, X. Gao, D. Jin, and J. Li, "Channel attention in LiDAR-camera fusion for lane line segmentation," Pattern Recognit., vol. 118, p. 108020, 2021.

[138] J. Stallkamp, M. Schlipsing, J. Salmen, and C. Igel, "The German traffic sign recognition benchmark: a multi-class classification compe- tition," in Proc. IJCNN. IEEE, 2011, pp. 1453–1460.

[139] S. Houben, J. Stallkamp, J. Salmen, M. Schlipsing, and C. Igel, "Detection of traffic signs in real-world images: The german traffic sign detection benchmark," in Proc. IJCNN. Ieee, 2013, pp. 1–8.

[140] H. Luo, Y. Yang, B. Tong, F. Wu, and B. Fan, "Traffic sign recognition using a multi-task convolutional neural network," IEEE Trans. Intell. Transp. Syst., vol. 19, no. 4, pp. 1100–1111, 2017.

[141] H. S. Lee and K. Kim, "Simultaneous traffic sign detection and boundary estimation using convolutional neural network," IEEE Trans. Intell. Transp. Syst., vol. 19, no. 5, pp. 1652–1663, 2018.

[142] J. Li and Z. Wang, "Real-time traffic sign recognition based on efficient CNNs in the wild," IEEE Trans. Intell. Transp. Syst., vol. 20, no. 3, pp. 975–984, 2018.

[143] U. Kamal, T. I. Tonmoy, S. Das, and M. K. Hasan, "Automatic traffic sign detection and recognition using SegU-Net and a modified Tversky loss function with L1-constraint," IEEE Trans. Intell. Transp. Syst., vol. 21, no. 4, pp. 1467–1479, 2019.

[144] W. Min, R. Liu, D. He, Q. Han, Q. Wei, and Q. Wang, "Traffic sign recognition based on semantic scene understanding and structural traffic sign location," IEEE Trans. Intell. Transp. Syst., vol. 23, no. 9, pp. 15 794–15 807, 2022.

[145] A. Taniguchi, M. Enoch, A. Theofilatos, and P. Ieromonachou, "Un-derstanding acceptance of autonomous vehicles in Japan, UK, and Germany," Urban, Planning and Transport Research, vol. 10, no. 1, pp. 514–535, 2022.

[146] K. Gopinath and G. Narayanamurthy, "Early bird catches the worm! meta-analysis of autonomous vehicles adoption–moderating role of automation level, ownership and culture," Int. J. Inf. Manage., vol. 66, p. 102536, 2022.

[147] AAA, "AAA: Today's vehicle technology must walk so self-driving cars can run," https://newsroom.aaa.com/2021/02/ aaa-todays-vehicle-technology-must-walk-so-self-driving-cars-can-run/, 2021.

[148] B. Schoettle and M. Sivak, "Public opinion about self-driving vehicles in China, India, Japan, the US, the UK, and Australia," University of Michigan, Ann Arbor, Transportation Research Institute, Tech. Rep., 2014.

[149] M. Kyriakidis, R. Happee, and J. C. de Winter, "Public opinion on au-tomated driving: Results of an international questionnaire among 5000 respondents," Transportation Research Part F: Traffic Psychology and Behaviour, vol. 32, pp. 127–140, 2015.

[150] C. J. Haboucha, R. Ishaq, and Y. Shiftan, "User preferences regarding autonomous vehicles," Transportation Research Part C: Emerging Technologies, vol. 78, pp. 37–49, 2017.

[151] C. Lee, C. Ward, M. Raue, L. D'Ambrosio, and J. F. Coughlin, "Age differences in acceptance of self-driving cars: A survey of perceptions and attitudes," in Human Aspects of IT for the Aged Population. Aging, Design and User Experience: Third International Conference, ITAP 2017, Held as Part of HCI International 2017, Vancouver, BC, Canada, July 9-14, 2017, Proceedings, Part I 3. Springer, 2017, pp. 3–13.

[152] J. Lee, H. Chang, and Y. I. Park, "Influencing factors on social acceptance of autonomous vehicles and policy implications," in 2018 Portland International Conference on Management of Engineering and Technology (PICMET). IEEE, 2018, pp. 1–6.

[153] S.-A. Kaye, I. Lewis, S. Forward, and P. Delhomme, "A priori accep-tance of highly automated cars in Australia, France, and Sweden: A theoretically-informed investigation guided by the TPB and UTAUT," Accident Analysis & Prevention, vol. 137, p. 105441, 2020.

[154] D. Potoglou, C. Whittle, I. Tsouros, and L. Whitmarsh, "Consumer intentions for alternative fuelled and autonomous vehicles: A segmen-tation analysis across six countries," Transportation Research Part D: Transport and Environment, vol. 79, p. 102243, 2020.

[155] S. S. Man, W. Xiong, F. Chang, and A. H. S. Chan, "Critical factors influencing acceptance of automated vehicles by Hong Kong drivers," IEEE Access, vol. 8, pp. 109 845–109 856, 2020.

[156] T. Zhang, D. Tao, X. Qu, X. Zhang, J. Zeng, H. Zhu, and H. Zhu, "Automated vehicle acceptance in China: Social influence and initial trust are key determinants," Transportation research part C: emerging technologies, vol. 112, pp. 220–233, 2020.

[157] Y. Huang and L. Qian, "Understanding the potential adoption of autonomous vehicles in China: The perspective of behavioral reasoning theory," Psychology & Marketing, vol. 38, no. 4, pp. 669–690, 2021.

[158] Y. Bao, K. Z. Zhou, and C. Su, "Face consciousness and risk aversion: do they affect consumer decision-making?" Psychology & Marketing, vol. 20, no. 8, pp. 733–755, 2003.

[159] D. Escandon-Barbosa, J. Salas-Paramo, A. I. Meneses-Franco, and C. Giraldo-Gonzalez, "Adoption of new technologies in developing countries: The case of autonomous car between Vietnam and Colom- bia," Technology in Society, vol. 66, p. 101674, 2021.

[160] Y. Yun, H. Oh, and R. Myung, "Statistical modeling of cultural differences in adopting autonomous vehicles," Applied Sciences, vol. 11, no. 19, p. 9030, 2021.

[161] K. Laigo, "Masculine vs. Feminine Culture: Another layer of culture," https://witi.com/articles/1824/Masculine-vs.-Feminine-Culture:-Another-Layer-of-Culture/.

[162] L. Yahiaoui, M. U˝ric´a´˝r, A. Das, and S. Yogamani, "Let the sunshine in: Sun glare detection on automotive surround-view cameras," Electronic Imaging, vol. 2020, no. 16, pp. 80–1, 2020.

[163] J. Zhang and K. B. Letaief, "Mobile edge intelligence and computing for the internet of vehicles," Proc. IEEE, vol. 108, no. 2, pp. 246–261, 2019.

[164] S.-C. Lin, Y. Zhang, C.-H. Hsu, M. Skach, M. E. Haque, L. Tang, and J. Mars, "The architectural implications of autonomous driving: Constraints and acceleration," in Proceedings of the Twenty-Third International Conference on Architectural Support for Programming Languages and Operating Systems, 2018, pp. 751–766.

[165] S. LeVine, "What it really costs to turn a car into a self-driving vehicle," https://qz.com/924212/ what-it-really-costs-to-turn-a-car-into-a-self-driving-vehicle, 2017.

[166] K. Othman, "Exploring the implications of autonomous vehicles: A comprehensive review," Innovative Infrastructure Solutions, vol. 7, no. 2, p. 165, 2022.

[167] Bloomberg News, "Tesla is testing self-driving cars on California roads," 2017. [Online]. Available: https://www.iotworldtoday.com/transportation-logistics/tesla-is-testing-self-driving-cars-on-california-roads

[168] T. May, "Mercedes-benz to test level 4 autonomous vehicles in Beijing," 2024. [Online]. Available: https://futuretransport-news.com/mercedes-benz-to-test-level-4-autonomous-vehicles-in-beijing/

[169] I. Wood, "RTA and cruise begin AV trial on dubai roads," 2023. [Online]. Available: https://www.automotivetestingtechnologyinternational.com/news/adas-cavs/rta-and-cruise-begin-av-trial-on-dubai-roads.html

[170] "L3Pilot: Joint European effort boosts automated driving," Connected Automated Driving, 2024. [Online]. Available: https://www.connectedautomateddriving.eu/blog/l3pilot-joint-european-effort-boosts-automated-driving/

[171] M. Saines, "Autonomous trucks begin testing on Japanese expressway," 2023. [Online]. Available: https://www.gpsworld.com/autonomous-trucks-begin-testing-on-japanese-expressway/

[172] "Robotaxis get green light for airport run," People's Daily, 2023. [Online]. Available: https://english.beijing.gov.cn/beijinginfo/sci/latesttrend/latesttrendlist/202403/t20240301_3577675.html

[173] A. Look, "Dubai to start robotaxi trials next month in major autonomous push," 2023. [Online]. Available: https://edition.cnn.com/2023/09/27/tech/dubai-rolls-out-robotaxis-mime-intl/index.html

[174] J. Elias, "Waymo opens robotaxi service to all san francisco users," 2024. [Online]. Available: https://www.cnbc.com/2024/06/25/waymo-opens-robotaxi-service-to-all-san-francisco-users.html

[175] "TIER IV to launch robotaxi business in Tokyo, november 2024: Starting with limited operational domains," Tier IV, 2024. [Online]. Available: https://tier4.jp/en/media/detail/sys id=3zc5Jf8KEJ9DOqewaAiUM&category=NEWS

[176] M. Naiseh, J. Clark, T. Akarsu, Y. Hanoch, M. Brito, M. Wald, T. Webster, and P. Shukla, "Trust, risk perception, and intention to use autonomous vehicles: an interdisciplinary bibliometric review," AI & SOCIETY, pp. 1–21, 2024.

[177] T. Krisher, "Tractor-trailers with no one aboard? the future is near for self-driving trucks on us roads," 2024. [Online]. Available: https://safety21.cmu.edu/2024/06/12/tractor-trailers-with-no-one-aboard-the-future-is-near-for-self-driving-trucks-on-us-roads/

[178] M. A. Khan, H. E. Sayed, S. Malik, T. Zia, J. Khan, N. Alkaabi, and H. Ignatious, "Level-5 autonomous driving—are we there yet? a review of research literature," ACM Computing Surveys (CSUR), vol. 55, no. 2, pp. 1–38, 2022.

[179] Abbass, H. A., Scholz, J., & Reid, D. (Eds.). (2018). Foundations of trusted autonomy. Cham, Switzerland: Springer International Publishing. https://doi.org/10.1007/978-3-319-64816-3

[180] J. Janai, F. Güney, A. Behl, A. Geiger et al., "Computer vision for autonomous vehicles: Problems, datasets and state of the art," Foundations and Trends® in Computer Graphics and Vision, vol. 12, no. 1–3, pp. 1–308, 2020.

[181] Z. Wang, Y. Wu, and Q. Niu, "Multi-sensor fusion in automated driving: A survey," IEEE Access, vol. 8, pp. 2847–2868, 2019.

[182] S. Yao, R. Guan, X. Huang, Z. Li, X. Sha, Y. Yue, E. G. Lim, H. Seo, K. L. Man, X. Zhu et al., "Radar-camera fusion for object detection and semantic segmentation in autonomous driving: A comprehensive review," IEEE Transactions on Intelligent Vehicles, 2023.

[183] Ultralytics, "Ultralytics YOLOv8," 2023. [Online]. Available: https://docs.ultralytics.com/models/yolov8/

[184] K. Burnett, D. J. Yoon, Y. Wu, A. Z. Li, H. Zhang, S. Lu, J. Qian, W. K. Tseng, A. Lambert, K. Y. Leung et al., "Boreas: A multi-season autonomous driving dataset," The International Journal of Robotics Research, vol. 42, no. 1-2, pp. 33–42, 2023.

[185] W. Maddern, G. Pascoe, C. Linegar, and P. Newman, "1 year, 1000 km: The Oxford RobotCar dataset," Int. J. Robot. Res., vol. 36, no. 1, pp. 3–15, 2017.

[186] M. Chikaraishi, D. Khan, B. Yasuda, and A. Fujiwara, "Risk perception and social acceptability of autonomous vehicles: A case study in Hiroshima, Japan," Transport Policy, vol. 98, pp. 105–115, 2020.

[187] C. Cui, Y. Ma, X. Cao, W. Ye, Y. Zhou, K. Liang, J. Chen, J. Lu, Z. Yang, K.-D. Liao et al., "A survey on multimodal large language models for autonomous driving," in Proceedings of the IEEE/CVF Winter Conference on Applications of Computer Vision, 2024, pp. 958–979.

[188] S. Liu, L. Liu, J. Tang, B. Yu, Y. Wang, and W. Shi, "Edge computing for autonomous driving: Opportunities and challenges," Proceedings of the IEEE, vol. 107, no. 8, pp. 1697–1716, 2019.

[189] J. Gu, Z. Wang, J. Kuen, L. Ma, A. Shahroudy, B. Shuai, T. Liu, X. Wang, G. Wang, J. Cai et al., "Recent advances in convolutional neural networks," Pattern recognition, vol. 77, pp. 354–377, 2018.

[190] H. Shu, T. Liu, X. Mu, and D. Cao, "Driving Tasks Transfer Using Deep Reinforcement Learning for Decision-Making of Autonomous Vehicles in Unsignalized Intersection," IEEE Trans. Veh. Technol., vol. 71, no. 1, pp. 41-52, Jan. 2022, doi: 10.1109/TVT.2021.3121985.

[191] T. P. Lillicrap, "Continuous control with deep reinforcement learning," arXiv preprint arXiv:1509.02971, 2015.

[192] C. Liu, S. Lee, S. Varnhagen, and H. E. Tseng, "Path planning for autonomous vehicles using model predictive control," in Proc. IEEE Intell. Vehicles Symp. (IV), 2017, pp. 174-179, doi: 10.1109/IVS.2017.7995808.

[193] B. R. Kiran et al., "Deep reinforcement learning for autonomous driving: A survey," IEEE Trans. Intell. Transp. Syst., vol. 23, no. 6, pp. 4909-4926, 2021, doi: 10.1109/TITS.2021.3057874.

[194] A. Baheri, "Safe reinforcement learning with mixture density network, with application to autonomous driving," Results Control Optim., vol. 6, 2022, Art. no. 100095, doi: 10.1016/j.rico.2022.100095.

[195] C. E. Garcia, D. M. Prett, and M. Morari, "Model predictive control: Theory and practice—A survey," Automatica, vol. 25, no. 3, pp. 335-348, 1989.

[196] U. Rosolia and F. Borrelli, "Learning model predictive control for iterative tasks: A data-driven control framework," IEEE Trans. Autom. Control, vol. 63, no. 7, pp. 1883-1896, 2017, doi: 10.1109/TAC.2017.2732058.

[197] Z. Chen and X. Huang, "End-to-end learning for lane keeping of self-driving cars," in Proc. IEEE Intell. Vehicles Symp. (IV), 2017, pp. 1856-1860, doi: 10.1109/IVS.2017.7995968.

[198] S. Azam et al., "N 2 C: Neural network controller design using behavioral cloning," IEEE Trans. Intell. Transp. Syst., vol. 22, no. 7, pp. 4744-4756, 2021, doi: 10.1109/TITS.2020.3018899.

[199] M. Zanon and S. Gros, "Safe reinforcement learning using robust MPC," IEEE Trans. Autom. Control, vol. 66, no. 8, pp. 3638-3652, 2020, doi: 10.1109/TAC.2020.2990363.

[200] M. Babu et al., "Model predictive control for autonomous driving considering actuator dynamics," in Proc. Amer. Control Conf. (ACC), 2019, pp. 1983-1989, doi: 10.23919/ACC.2019.8815314.

[201] F. Codevilla et al., "End-to-end driving via conditional imitation learning," in Proc. IEEE Int. Conf. Robot. Autom. (ICRA), 2018, pp. 4693-4700, doi: 10.1109/ICRA.2018.8460487.

[202] S. Abhiman et al., "Towards autonomous driving system using behavioral cloning approach," in Proc. IEEE Glob. Conf. Adv. Technol. (GCAT), Bangalore, India, 2023, pp. 1-7.

[203] N. Albarella et al., "A hybrid deep reinforcement learning and optimal control architecture for autonomous highway driving," Energies, vol. 16, no. 8, Art. no. 3490, 2023, doi: 10.3390/en16083490.

[204] Z. Bai et al., "Hybrid reinforcement learning-based eco-driving strategy for connected and automated vehicles at signalized intersections," IEEE Trans. Intell. Transp. Syst., vol. 23, no. 9, pp. 15850-15863, 2022, doi: 10.1109/TITS.2021.3074823.

[205] L. Wang et al., "A combined reinforcement learning and model predictive control for car-following maneuver of autonomous vehicles," Chin. J. Mech. Eng., vol. 36, no. 1, Art. no. 80, 2023, doi: 10.1186/s10033-023-00779-7.

[206] A. Joglekar et al., "Hybrid reinforcement learning-based controller for autonomous navigation," in Proc. IEEE Veh. Technol. Conf. (VTC-Spring), 2022, pp. 1-6, doi: 10.1109/VTC2022-Spring54318.2022.9860857.

[207] B. Paden et al., "A survey of motion planning and control techniques for self-driving urban vehicles," IEEE Trans. Intell. Vehicles, vol. 1, no. 1, pp. 33-55, 2016, doi: 10.1109/TIV.2016.2578705.

[208] C. Hubmann et al., "Decision making for automated driving using probabilistic maneuver intention estimation," IEEE Robot. Autom. Lett., vol. 5, no. 4, pp. 6509-6516, 2020, doi: 10.1109/LRA.2020.3016657.

[209] C. Hubmann et al., "Automated driving in uncertain environments: Planning with interaction and uncertain maneuver prediction," IEEE Trans. Intell. Vehicles, vol. 3, no. 1, pp. 5-17, 2018, doi: 10.1109/TIV.2017.2788199.

[210] Z. Wang, Y. Wu, and Q. Niu, "Multi-sensor fusion in automated driving: A survey," IEEE Access, vol. 8, pp. 2847-2868, 2019, doi: 10.1109/ACCESS.2019.2961957.

[211] D. J. Yeong, G. Velasco-Hernandez, J. Barry, and J. Walsh, "Sensor and sensor fusion technology in autonomous vehicles: A review," Sensors, vol. 21, no. 6, p. 2140, 2021.

[212] C. Zhang et al., "Robust-FusionNet: Deep multimodal sensor fusion for 3-D object detection under severe weather conditions," IEEE Trans. Instrum. Meas., vol. 71, pp. 1-13, 2022, doi: 10.1109/TIM.2022.3162192.

[213] E. S. Lee and D. Kum, "Feature-based lateral position estimation of surrounding vehicles using stereo vision," in Proc. IEEE Intell. Vehicles Symp. (IV), 2017, pp. 779-784, doi: 10.1109/IVS.2017.7995811.

[214] C. Hubmann, J. Schulz, M. Becker, D. Althoff, and C. Stiller, "Automated driving in uncertain environments: Planning with interaction and uncertain maneuver prediction," IEEE Trans. Intell. Vehicles, vol. 3, no. 1, pp. 5-17, 2018, doi: 10.1109/TIV.2018.2788377.

[215] W. Song, G. Xiong, and H. Chen, "Intention-aware autonomous driving decision-making in an uncontrolled intersection," Math. Probl. Eng., vol. 2016, Art. no. 1025349, 2016, doi: 10.1155/2016/1025349.

[216] D. Sadigh, S. Sastry, S. A. Seshia, and A. D. Dragan, "Planning for autonomous cars that leverage effects on human actions," in Proc. Robot. Sci. Syst., vol. 2, 2016, pp. 1-9.

[217] W. Schwarting, J. Alonso-Mora, and D. Rus, "Planning and decision-making for autonomous vehicles," Annu. Rev. Control Robot. Auton. Syst., vol. 1, no. 1, pp. 187-210, 2018, doi: 10.1146/annurev-control-060117-105157.

[218] D. Foster, D. J. Foster, N. Golowich, and A. Rakhlin, "On the complexity of multi-agent decision making: From learning in games to partial monitoring," in Proc. 36th Annu. Conf. Learn. Theory, PMLR, 2023, pp. 2678-2792.

[219] M. Khonji et al., "A risk-aware architecture for autonomous vehicle operation under uncertainty," in Proc. IEEE Int. Symp. Safety, Security, Rescue Robot. (SSRR), 2020, pp. 311-317, doi: 10.1109/SSRR49932.2020.9274947.

[220] V. A. Vijayakumar, J. Shanthini, and S. Karthick, "Convolutional recurrent neural network framework for autonomous driving behavioral model," in Proc. Intell. Data Commun. Technol. Internet Things (ICICI), 2020, pp. 761-772, Springer Singapore.

[221] S. Mozaffari et al., "Deep learning-based vehicle behavior prediction for autonomous driving applications: A review," IEEE Trans. Intell. Transp. Syst., vol. 23, no. 1, pp. 33-47, 2020, doi: 10.1109/TITS.2019.2928675.

[222] R. Mur-Artal and J. D. Tardós, "ORB-SLAM2: An open-source SLAM system for monocular, stereo, and RGB-D cameras," IEEE Trans. Robot., vol. 33, no. 5, pp. 1255-1262, 2017, doi: 10.1109/TRO.2017.2705103.

[223] D. Cattaneo, M. Vaghi, and A. Valada, "LCDNet: Deep loop closure detection and point cloud registration for LiDAR SLAM," IEEE Trans. Robot., vol. 38, no. 4, pp. 2074-2093, 2022, doi: 10.1109/TRO.2022.3153156.

[224] C. Campos et al., "ORB-SLAM3: An accurate open-source library for visual, visual–inertial, and multimap SLAM," IEEE Trans. Robot., vol. 37, no. 6, pp. 1874-1890, 2021, doi: 10.1109/TRO.2021.3075642.

[225] S. Leutenegger et al., "Keyframe-based visual–inertial odometry using nonlinear optimization," Int. J. Robot. Res., vol. 34, no. 3, pp. 314-334, 2015, doi: 10.1177/0278364914554813.

[226] O. Dabeer et al., "An end-to-end system for crowdsourced 3D maps for autonomous vehicles: The mapping component," in Proc. IEEE/RSJ Int. Conf. Intell. Robots Syst. (IROS), 2017, pp. 634-641, doi: 10.1109/IROS.2017.8202181.

[227] Q. Liu, T. Han, J. Xie, and B. Kim, "Real-time dynamic map with crowdsourcing vehicles in edge computing," IEEE Trans. Intell. Vehicles, vol. 8, no. 4, pp. 2810-2820, 2022, doi: 10.1109/TIV.2022.3142175.

[228] Q. Liu, Y. Zhang, and H. Wang, "EdgeMap: Crowdsourcing high-definition map in automotive edge computing," in Proc. IEEE Int. Conf. Commun. (ICC), 2022, pp. 4300-4305, doi: 10.1109/ICC42927.2022.9877385.

[229] Q. Liu, T. Han, L. Xie, and B. Kim, "LiveMap: Real-time dynamic map in automotive edge computing," in Proc. IEEE INFOCOM, 2021, pp. 1-10, doi: 10.1109/INFOCOM42981.2021.9488749.

[230] A. G. Barto, "Reinforcement learning: An introduction," SIAM Rev., vol. 6, no. 2, p. 423, 2021, doi: 10.1137/SIAMR2021.000093.

[231] G. E. Moore, "A defence of common sense," in Contemporary British Philosophy, J. Muirhead, Ed. London, U.K.: George Allen Unwin Ltd., 1925, pp. 32–59.

[232] J. McCarthy, "Programs with common sense at the Wayback Machine," in Proc. Teddington Conf. Mechanization Thought Processes, Her Majesty's Stationery Office, 1959, pp. 75–91.

[233] B. Smith, "Formal ontology, common sense and cognitive science," Int. J. Hum.-Comput. Stud., vol. 43, pp. 641–667, 1995, doi: 10.1016/ijhcs.1995.07.001.

[234] H. J. Levesque, Ed., Common Sense, the Turing Test, and the Quest for Real AI. Cambridge, MA, USA: MIT Press, 2017.

[235] S. Bubeck et al., "Sparks of artificial general intelligence: Early experiments with GPT-4," arXiv, vol. abs/2303.12712, 2023.

[236] H. J. Levesque, E. Davis, and L. Morgenstern, "The Winograd schema challenge," in Proc. Int. Conf. Principles Knowl. Representation Reasoning, 2012, pp. 552–561.

[237] J. Browning and Y. LeCun, "Language, common sense, and the Winograd schema challenge," Artif. Intell., vol. 325, p. 104031, 2023, doi: 10.1016/j.artint.2023.104031.

[238] S. Harnad, "Symbol grounding problem," Physica D, vol. 42, pp. 335–346, 1990, doi: 10.1016/0167-2789(90)90087-6.

[239] D. C. Mollo and R. Milli`ere, "The vector grounding problem," arXiv, vol. abs/2304.01481, 2023.

[240] A. Vaswani et al., "Attention is all you need," in Adv. Neural Inf. Process. Syst., 2017, pp. 6000–6010.

[241] A. Dosovitskiy et al., "An image is worth 16x16 words: Transformers for image recognition at scale," in Proc. Int. Conf. Learn. Representations, 2021.

[242] A. Radford et al., "Learning transferable visual models from natural language supervision," in Proc. Mach. Learn. Res., 2021, pp. 8748–8763.

[243] C. Jia et al., "Scaling up visual and vision-language representation learning with noisy text supervision," in Proc. Mach. Learn. Res., 2021, pp. 4904–4916.

[244] M. Tsimpoukelli et al., "Multimodal few-shot learning with frozen language models," in Adv. Neural Inf. Process. Syst., 2021.

[245] P. Gao et al., "LLaMA-Adapter V2: Parameter-efficient visual instruction model," arXiv, vol. abs/2304.15010, 2023.

[246] D. Fu et al., "Drive like a human: Rethinking autonomous driving with large language models," arXiv, vol. abs/2307.07162, 2023.

[247] J. Mao et al., "GPT-driver: Learning to drive with GPT," arXiv, vol. abs/2310.01415, 2023.

[248] H.-N. Hu et al., "Joint monocular 3D vehicle detection and tracking," in Proc. IEEE Int. Conf. Comput. Vis., 2019, pp. 5390–5399.

[249] S. Yao et al., "ReAct: Synergizing reasoning and acting in language models," in Adv. Neural Inf. Process. Syst., 2022.

[250] Z. Yang et al., "The dawn of LMMs: Preliminary explorations with GPT-4V(ision)," arXiv, vol. abs/2309.17421, 2023.

[251] H. Liu et al., "Visual instruction tuning," in Adv. Neural Inf. Process. Syst., 2024, pp. 34892–34916.

[252] Z. Chen et al., "InternVL: Scaling up vision foundation models and aligning for generic visual-linguistic tasks," in Proc. IEEE Conf. Comput. Vis. Pattern Recognit., 2024, pp. 24185–24198.

[253] C. Cui et al., "A survey on multimodal large language models for autonomous driving," in Proc. IEEE/CVF Winter Conf. Appl. Comput. Vis., 2023, pp. 958–979.

[254] S. Luo et al., "Delving into multi-modal multi-task foundation models for road scene understanding: From learning paradigm perspectives," arXiv, vol. abs/2401.08045, 2024.

[255] X. Yan et al., "Forging vision foundation models for autonomous driving: Challenges, methodologies, and opportunities," arXiv, vol. abs/2401.08045, 2024.

[256] W. Han et al., "DME-Driver: Integrating human decision logic and 3D scene perception in autonomous driving," arXiv, vol. abs/2401.03641, 2024.

[257] A. Dosovitskiy et al., "CARLA: An open urban driving simulator," in Proc. 1st Annu. Conf. Robot Learn., 2017, pp. 1–16.

[258] Z. Xu et al., "DriveGPT4: Interpretable end-to-end autonomous driving via large language model," IEEE Robot. Autom. Lett., vol. 9, pp. 8186–8193, 2024, doi: 10.1109/LRA.2024.3145000.

[259] J. Kim et al., "Textual explanations for self-driving vehicles," in Proc. Eur. Conf. Comput. Vis., 2018, pp. 563–578.

[260] Y. Huang, J. Sansom, Z. Ma, F. Gervits, and J. Chai, "DriVLMe: Enhancing LLM-based autonomous driving agents with embodied and

social experiences," in Proc. IEEE Int. Conf. Comput. Vis. Pattern Recognit. (CVPR), 2024.

[261] Y. Ma, A. Abdelraouf, R. Gupta, Z. Wang, and K. Han, "Video token sparsification for efficient multimodal LLMs in autonomous driving," arXiv, vol. abs/2409.11182, 2024.

[262] S. Hu et al., "AgentsCoMerge: Large language model empowered collaborative decision making for ramp merging," arXiv, vol. abs/2408.03624, 2024.

[263] H. Liu, R. Yao, Z. Huang, S. Shen, and J. Ma, "LMMCoDrive: Cooperative driving with large multimodal model," arXiv, vol. abs/2401.08045, 2024.

[264] S. Sural and R. Rajkumar, "ContextVLM: Zero-shot and few-shot context understanding for autonomous driving using vision language models," arXiv, vol. abs/2409.00301, 2024.

[265] L. Wen et al., "On the road with GPT-4V(ision): Early explorations of visual-language model on autonomous driving," arXiv, vol. abs/2311.05332, 2023.

[266] K. Li et al., "CODA: A real-world road corner case dataset for object detection in autonomous driving," in Proc. Eur. Conf. Comput. Vis. (ECCV), 2022, pp. 406–423.

[267] K. Chen et al., "Automated evaluation of large vision-language models on self-driving corner cases," arXiv, vol. abs/2404.10595, 2024.

[268] Y. Xue et al., "The Greatest LVLM system: 1st place solution for ECCV 2024 corner case scene understanding challenge," in Proc. Eur. Conf. Comput. Vis. (ECCV), 2024.

[269] X. Han, Y. Huang, S. Tang, and X. Chu, "From regional to general: A vision-language model-based framework for corner cases comprehension in autonomous driving," in Proc. Eur. Conf. Comput. Vis. (ECCV), 2024.

[270] Y. Cheng, M.-H. Chen, and S.-H. Lai, "A lightweight vision-language model pipeline for corner-case scene understanding in autonomous driving," in Proc. Eur. Conf. Comput. Vis. (ECCV), 2024.

[271] B. Zhou et al., "TinyLLaVA: A framework of small-scale large multimodal models," arXiv, vol. abs/2402.14289, 2024.

[272] Y. Deng et al., "An analysis of adversarial attacks and defenses on autonomous driving models," in Proc. IEEE Int. Conf. Pervasive Comput. Commun., 2020.

[273] B. Badjie, J. Cecílio, and A. Casimiro, "Adversarial attacks and countermeasures on image classification-based deep learning models in autonomous driving systems: A systematic review," ACM Comput. Surveys, vol. 57, Art. no. 20, 2024.

[274] M. Aldeen et al., "An initial exploration of employing large multimodal models in defending against autonomous vehicles attacks," in Proc. IEEE Intell. Vehicles Symp. (IV), 2024, pp. 3334–3341.

[275] R. Song et al., "Enhancing LLM-based autonomous driving agents to mitigate perception attacks," arXiv, vol. abs/2409.14488, 2024.

[276] S. Zernetsch et al., "Trajectory forecasts with uncertainties of vulnerable road users by means of neural networks," in Proc. IEEE Intell. Vehicles Symp. (IV), 2019, pp. 810–815.

[277] A. Rasouli, I. Kotseruba, and J. K. Tsotsos, "Understanding pedestrian behavior in complex traffic scenes," IEEE Trans. Intell. Vehicles, vol. 3, pp. 61–70, 2018.

[278] F. Camara et al., "Pedestrian models for autonomous driving part II: High-level models of human behavior," IEEE Trans. Intell. Transp. Syst., vol. 22, pp. 5453–5472, 2021.

[279] J. Huang et al., "GPT-4V takes the wheel: Evaluating promise and challenges for pedestrian behavior prediction," arXiv, vol. abs/2311.14786, 2023.

[280] A. Plebe et al., "Human-inspired autonomous driving: A survey," Cogn. Syst. Res., vol. 83, Art. no. 101169, 2024. [Online]. Available: https://doi.org/10.1016/j.cogsys.2023.101169.

[281] B. Goertzel et al., Eds., Artificial General Intelligence. Berlin, Germany: Springer-Verlag, 2023.

[282] Z. Yuan, Z. Li, and L. Sun, "TinyGPT-V: Efficient multimodal large language with small backbones," arXiv, vol. abs/2312.16862, 2023.

[283] "Road Infrastructure Challenges Faced by Automated Driving: A Review," Accessed: Nov. 4, 2024. [Online]. Available: https://www.mdpi.com/2076-3417/12/7/3477.

[284] K. Coyner and J. Bittner, "Automated Vehicles and Infrastructure Enablers: Pavement Markings and Signs," SAE Tech. Paper EPR2022011, Warrendale, PA, USA, May 2022, doi: 10.4271/EPR2022011.

[285] K. Roy, "Modelling and Evaluating the Deployment of an Autonomous Vehicle Zone in Transportation Networks," M.S. thesis, Monash Univ., 2021, doi: 10.26180/15057804.v1.

[286] J. Khoury, K. Amine, and R. A. Saad, "An Initial Investigation of the Effects of a Fully Automated Vehicle Fleet on Geometric Design," J. Adv. Transp., vol. 2019, Art. no. 6126408, 2019, doi: 10.1155/2019/6126408.

[287] S. Wang, B. Yu, Y. Ma, J. Liu, and W. Zhou, "Impacts of Different Driving Automation Levels on Highway Geometric Design from the Perspective of Trucks," J. Adv. Transp., vol. 2021, Art. no. 5541878, 2021, doi: 10.1155/2021/5541878.

[288] A. Č. Ivančev, V. Dragčević, and T. Džambas, "Road infrastructure requirements to accommodate autonomous vehicles," in Proc. 7th Int. Conf. Road Rail Infrastruct., May 2024. Accessed: Nov. 4, 2024. [Online]. Available: https://cetra.grad.hr/ocs/index.php/cetra7/cetra2022/paper/view/1462.

[289] Y. Zhao, X. Ying, and J. Li, "Research on Geometric Design Standards for Freeways under a Fully Autonomous Driving Environment," Appl. Sci., vol. 12, no. 14, Art. no. 14, Jan. 2022, doi: 10.3390/app12147109.

[290] A. Amditis et al., "Road infrastructure taxonomy for connected and automated driving," in Cooperative Intell. Transp. Syst.: Towards High-Level Automated Driving, pp. 309–325, doi: 10.1049/PBTR025E_ch14.

[291] T. U. Saeed, "Road Infrastructure Readiness for Autonomous Vehicles," Ph.D. dissertation, Purdue Univ. Grad. Sch., 2019, doi: 10.25394/PGS.8949011.v1.

[292] M. Waqas and P. Ioannou, "Automatic Vehicle Following Under Safety, Comfort, and Road Geometry Constraints," IEEE Trans. Intell. Vehicles, vol. 8, no. 1, pp. 531–546, Jan. 2023, doi: 10.1109/TIV.2022.3177176.

[293] S. Wang et al., "Sight Distance of Automated Vehicles Considering Highway Vertical Alignments and Its Implications for Speed Limits," IEEE Intell. Transp. Syst. Mag., vol. 16, no. 4, pp. 45–61, Jul. 2024, doi: 10.1109/MITS.2023.3334769.

[294] I. Najeh et al., "Maintenance Strategy for the Road Infrastructure for the Autonomous Vehicle," presented at the ESREL'20 - PSAM 15, 30th Eur. Safety Rel. Conf. and 15th Probabilistic Safety Assess. Manag. Conf., 2020. Accessed: Nov. 4, 2024. [Online]. Available: https://trid.trb.org/View/1899368.

[295] O. Tengilimoglu, O. Carsten, and Z. Wadud, "Infrastructure-related challenges in implementing connected and automated vehicles on urban roads: Insights from experts and stakeholders," IET Intell. Transp. Syst., vol. 17, no. 12, pp. 2352–2368, 2023, doi: 10.1049/itr2.12413.

[296] F. Chen, R. Balieu, and N. Kringos, "Potential Influences on Long-Term Service Performance of Road Infrastructure by Automated Vehicles," Transp. Res. Rec., vol. 2550, no. 1, pp. 72–79, Jan. 2016, doi: 10.3141/2550-10.

[297] K. Othman, "Impact of Autonomous Vehicles on the Physical Infrastructure: Changes and Challenges," Designs, vol. 5, no. 3, Art. no. 3, Sep. 2021, doi: 10.3390/designs5030040.

[298] X. Lu, "Infrastructure Requirements for Automated Driving," 2018. Accessed: Nov. 4, 2024. [Online]. Available: https://www.semanticscholar.org/paper/Infrastructure-Requirements-for-Automated-Driving-Lu/8271a578fcff810fbd8491baffd5afe44e9f52d5.

[299] A. Biswas and H.-C. Wang, "Autonomous Vehicles Enabled by the Integration of IoT, Edge Intelligence, 5G, and Blockchain," Sensors, vol. 23, no. 4, Art. no. 1963, Feb. 2023, doi: 10.3390/s23041963.

[300] M. Sadaf et al., "Connected and Automated Vehicles: Infrastructure, Applications, Security, Critical Challenges, and Future Aspects," Technologies, vol. 11, no. 5, Art. no. 5, Oct. 2023, doi: 10.3390/technologies11050117.

[301] S. A. Yusuf, A. Khan, and R. Souissi, "Vehicle-to-everything (V2X) in the autonomous vehicles domain – A technical review of communication, sensor, and AI technologies for road user safety," Transp. Res. Interdiscip. Perspect., vol. 23, Art. no. 100980, Jan. 2024, doi: 10.1016/j.trip.2023.100980.

[302] M. J. Khan et al., "Augmenting CCAM Infrastructure for Creating Smart Roads and Enabling Autonomous Driving," Remote Sens., vol. 15, no. 4, Art. no. 4, Jan. 2023, doi: 10.3390/rs15040922.

[303] S. Wang and Z. Li, "Roadside Sensing Information Enabled Horizontal Curve Crash Avoidance System Based on Connected and Autonomous Vehicle Technology," Transp. Res. Rec., vol. 2673, no. 5, pp. 49–60, May 2019, doi: 10.1177/0361198119837957.

[304] Y. Cao et al., "Study on the Deviation Characteristics of Driving Trajectories for Autonomous Vehicles and the Design of Dedicated Lane Widths," Sustainability, vol. 16, no. 21, Art. no. 21, Jan. 2024, doi: 10.3390/su16219155.

[305] S. G. Machiani et al., "Implications of a Narrow Automated Vehicle-Exclusive Lane on Interstate 15 Express Lanes," J. Adv. Transp., vol. 2021, Art. no. 6617205, 2021, doi: 10.1155/2021/6617205.

[306] P. Aryal, "Optimization of geometric road design for autonomous vehicle," M.S. thesis, KTH, 2020. Accessed: Nov. 4, 2024. [Online]. Available: https://urn.kb.se/resolve?urn=urn:nbn:se:kth:diva-290030.

[307] S. Mecheri, F. Rosey, and R. Lobjois, "The effects of lane width, shoulder width, and road cross-sectional reallocation on drivers' behavioral adaptations," Accid. Anal. Prev., vol. 104, pp. 65–73, Jul. 2017, doi: 10.1016/j.aap.2017.04.019.

[308] D. P., "Preparing Infrastructure for Automated Vehicles," ITF. Accessed: Nov. 4, 2024. [Online]. Available: https://www.itf-oecd.org/preparing-infrastructure-automated-vehicles.

[309] Y. Rahmati et al., "Toward Human-Centered Design of Automated Vehicles: A Naturalistic Brake Policy," Front. Future Transp., vol. 2, Jun. 2021, doi: 10.3389/ffutr.2021.683223.

[310] A. Yeganeh, B. Vandoren, and A. Pirdavani, "Automated Trucks' Impact on Pavement Fatigue Damage," Appl. Sci., vol. 14, no. 13, Art. no. 13, Jan. 2024, doi: 10.3390/app14135552.

[311] M. M. Rana, "Effects of autonomous vehicles on pavement distress & road safety and pavement distress optimization," M.S. thesis, Memorial Univ. Newfoundland, 2021. Accessed: Nov. 4, 2024. [Online]. Available: https://research.library.mun.ca/15029/.

[312] Y. Mo et al., "Enhanced Perception for Autonomous Vehicles at Obstructed Intersections: An Implementation of Vehicle to Infrastructure (V2I) Collaboration," Sensors, vol. 24, no. 3, Art. no. 3, Jan. 2024, doi: 10.3390/s24030936.

[313] "Navigating Intersection Challenges: Enabling Emergency Responders and Autonomous Vehicles with LYT - LYT," Accessed: Nov. 4, 2024. [Online]. Available: https://lyt.ai/blog/navigating-intersection-challenges-with-avs/.

[314] U. Malik, "Autonomous Vehicles in Road Tunnels: A Risk Safety Perspective," M.S. thesis, Univ. Stavanger, 2023. Accessed: Nov. 4, 2024. [Online]. Available: https://uis.brage.unit.no/uis-xmlui/handle/11250/3089887.

[315] S. A. Bagloee et al., "Autonomous vehicles: Challenges, opportunities, and future implications for transportation policies," J. Mod. Transp., vol. 24, no. 4, pp. 284–303, Dec. 2016, doi: 10.1007/s40534-016-0117-3.

[316] P. India, "Autonomy Delay: Current Challenges and Bottlenecks," FutureBridge. Accessed: Nov. 4, 2024. [Online]. Available: https://www.futurebridge.com/blog/autonomy-delay-current-challenges-and-bottlenecks/.

[317] "How autonomous vehicles will change road designs | ASCE," Accessed: Nov. 4, 2024. [Online]. Available: https://www.asce.org/publications-and-news/civil-engineering-source/civil-engineering-magazine/article/2021/08/how-autonomous-vehicles-will-change-road-designs.

[318] P. Koopman and B. Osyk, "Safety Argument Considerations for Public Road Testing of Autonomous Vehicles," vol. 1, 2019, doi: 10.4271/2019-01-0123.

[319] N. Agatz, A. Erera, M. Savelsbergh, and X. Wang, "Optimization for dynamic ride-sharing: A review," Eur. J. Oper. Res., vol. 223, no. 2, pp. 295–303, 2012, doi: 10.1016/j.ejor.2012.05.028.

[320] A. Akella, S. Seshan, R. Karp, S. Shenker, and C. Papadimitriou, "Selfish behavior and stability of the internet: A game-theoretic analysis of TCP," ACM SIGCOMM Comput. Commun. Rev., vol. 32, no. 4, pp. 117–130, 2002, doi: 10.1145/964725.633049.

[321] B. M. Baker and M. Ayechew, "A genetic algorithm for the vehicle routing problem," Comput. Oper. Res., vol. 30, no. 5, pp. 787–800, 2003, doi: 10.1016/S0305-0548(02)00051-5.

[322] P. G. Balaji and D. Srinivasan, "An introduction to multi-agent systems," in Innovations in Multi-Agent Systems and Applications-1, pp. 1–27, 2010, doi: 10.1007/978-3-642-14435-6_1.

[323] R. Bellman, "On a routing problem," Q. Appl. Math., vol. 16, no. 1, pp. 87–90, 1958, doi: 10.1090/qam/99847.

[324] C. A. Boer et al., "Distributed e-services for road container transport simulation," in Proc. 15th Eur. Simul. Symp., pp. 541–550, 2003.

[325] D. de Jonge and C. Sierra, "Automated negotiation for package delivery," in Proc. IEEE Int. Conf. Self-Adaptive Self-Organizing Syst. Workshops, 2012, pp. 83–88, doi: 10.1109/SASO.2012.10.

[326] D. de Jonge and C. Sierra, "NB3: A multilateral negotiation algorithm for large, non-linear agreement spaces with limited time," Auton. Agents Multi-Agent Syst., vol. 29, no. 5, pp. 896–942, 2015, doi: 10.1007/s10458-015-9308-1.

[327] D. de Jonge and D. Zhang, "GDL as a unifying domain description language for declarative automated negotiation," Auton. Agents Multi-Agent Syst., vol. 35, no. 1, Art. no. 13, 2021, doi: 10.1007/s10458-020-09494-1.

[328] D. de Jonge, F. Bistaffa, and J. Levy, "A heuristic algorithm for multi-agent vehicle routing with automated negotiation," 2021.

[329] A. de La Fortelle, "Analysis of reservation algorithms for cooperative planning at intersections," in Proc. Int. IEEE Conf. Intell. Transp. Syst. (ITSC), Sep. 2010, pp. 445–449, doi: 10.1109/ITSC.2010.5625021.

[330] P. Toth and D. Vigo, The Vehicle Routing Problem. Philadelphia, PA, USA: SIAM, 2001, doi: 10.1137/1.9780898718515.

[331] J. Wang, Y. Shao, Y. Ge, and R. Yu, "A survey of vehicle-to-everything (V2X) testing," Sensors, vol. 19, no. 2, Art. no. 334, 2019, doi: 10.3390/s19020334.

[332] M. Wang, Z. Chen, L. Mu, and X. Zhang, "Road network structure and ride-sharing accessibility: A network science perspective," Comput. Environ. Urban Syst., vol. 80, Art. no. 101430, 2020, doi: 10.1016/j.compenvurbsys.2020.101430.

[333] J. G. Wardrop, "Road paper. Some theoretical aspects of road traffic research," Proc. Inst. Civ. Eng., vol. 1, no. 3, pp. 325–362, 1952, doi: 10.1680/ipeds.1952.11259.

[334] C. Wuthishuwong and A. Traechtler, "Vehicle-to-infrastructure based safe trajectory planning for autonomous intersection management," in Proc. IEEE Int. Conf. ITS Telecommun. (ITST), Nov. 2013, pp. 175–180, doi: 10.1109/ITST.2013.6685554.

[335] B. Xu et al., "V2I based cooperation between traffic signal and approaching automated vehicles," in Proc. IEEE Intell. Vehicles Symp. (IV), Jun. 2017, pp. 1658–1664, doi: 10.1109/IVS.2017.7995868.

[336] J. Yang, B. Pelletier, and B. Champagne, "Enhanced autonomous resource selection for LTE-based V2V communication," in Proc. IEEE Global Commun. Conf. (GLOBECOM), Dec. 2016, pp. 1–6, doi: 10.1109/GLOCOM.2016.7841881.

[337] J. Qiao, D. Zhang, and D. de Jonge, "Priority-based traffic management protocols for autonomous vehicles on road networks," in Proc. Australas. Joint Conf. Artif. Intell., 2022, pp. 240–253, doi: 10.1007/978-3-031-24667-3_19.

[338] J. Qiao et al., "A hybrid model of traffic assignment and control for autonomous vehicles," in Proc. PRIMA Int. Conf. Principles Practice Multi-Agent Syst., 2022, pp. 208–226, doi: 10.1007/978-3-031-21729-1_14.

[339] J. Qiao, D. Zhang, and D. de Jonge, "Graph representation of road and traffic for autonomous driving," in Proc. PRICAI Int. Conf. Trends Artif. Intell., 2019.

[340] M. R. Genesereth and M. Thielscher, General Game Playing. San Rafael, CA, USA: Morgan & Claypool, 2014.

[341] M. Thielscher and D. Zhang, "From general game descriptions to a market specification language for general trading agents," in Proc. Agent-Mediated Electron. Commerce, Lect. Notes Bus. Inf. Process., vol. 59, pp. 259–274, Springer, 2009.

[342] T. Roughgarden, "Selfish routing with atomic players," in Proc. ACM-SIAM Symp. Discrete Algorithms (SODA), 2005, pp. 1184–1185.

[343] M. Mittelmann, S. Bouveret, and L. Perrussel, "Representing and reasoning about auctions," Auton. Agents Multi-Agent Syst., vol. 36, no. 1, Art. no. 20, 2022, doi: 10.1007/s10458-021-09535-8.

[344] D. de Jonge and D. Zhang, "GDL as a Unifying Domain Description Language for Declarative Automated Negotiation," in Proc. 21st Int. Conf. Auton. Agents Multiagent Syst. (AAMAS), 2022, pp. 1935–1937.

[345] D. de Jonge and D. Zhang, "GDL as a unifying domain description language for declarative automated negotiation," Auton. Agents Multi-Agent Syst., vol. 35, no. 1, 2021.

[346] B. L. Golden, S. Raghavan, E. A. Wasil, et al., The Vehicle Routing Problem: Latest Advances and New Challenges, vol. 43. Springer, 2008.

[347] G. Laporte, "The vehicle routing problem: An overview of exact and approximate algorithms," Eur. J. Oper. Res., vol. 59, no. 3, pp. 345–358, 1992.

[348] T.-H. Nguyen and J. J. Jung, "Eco-based traffic routing method with automated negotiation for connected vehicles," Complex Intell. Syst., vol. 9, no. 1, pp. 625–636, 2023.

[349] H. Rakha and R. K. Kamalanathsharma, "Eco-driving at signalized intersections using V2I communication," in Proc. 14th Int. IEEE Conf. Intell. Transp. Syst. (ITSC), 2011, pp. 341–346.

[350] Z. Y. Rawashdeh and S. M. Mahmud, "Intersection collision avoidance system architecture," in Proc. 5th IEEE Consum. Commun. Netw. Conf., Jan. 2008, pp. 493–494.

[351] M. S. Sanders and E. J. McCormick, Human Factors in Engineering and Design. McGraw-Hill, 1993.

[352] D. L. Fisher, J. D. Lee, M. A. Regan, T. L. Dingus, and M. T. Young, Handbook of Human Factors for Automated, Connected, and Intelligent Vehicles. CRC Press, 2020.

[353] M. Martens and A. P. Beukel, "The road to automated driving: dual mode and human factors considerations," in Proc. 16th Int. IEEE Annu. Conf. Intell. Transp. Syst. (ITSC), Netherlands, 2013, pp. 2262–2267.

[354] M. Cunningham and M. Regan, "Autonomous vehicles: Human factors issues and future research," in Proc. 2015 Australas. Road Safety Conf., 2015.

[355] S. L. Schladover, "Automated vehicles and autonomous vehicles: Technical differences and policy implications," in Handbook of Human Factors for Automated, Connected, and Intelligent Vehicles, D. L. Fisher, W. J. Horrey, J. D. Lee, and M. A. Regan, Eds. Florida, USA: CRC Press, 2020, p. 190.

[356] J. Sweller, "Cognitive load during problem solving: Effects on learning," Cogn. Sci., vol. 12, no. 2, pp. 257–285, 1988.

[357] P. A. Desmond and P. A. Hancock, "Active and passive fatigue states," in Stress, Workload, and Fatigue. CRC Press, 2000, pp. 455–465.

[358] C. Neubauer, G. Matthews, L. Langheim, and D. Saxby, "Fatigue and voluntary utilization of automation in simulated driving," Hum. Factors, vol. 54, no. 5, pp. 734–746, 2012.

[359] N. Merat, A. H. Jamson, F. Lai, M. Daly, and O. Carsten, "Transition to manual: Driver behaviour when resuming control from a highly automated vehicle," Transp. Res. Part F: Traffic Psychol. Behav., vol. 26, pp. 1–9, 2014.

[360] M. S. Young and N. A. Stanton, "Back to the future: Brake reaction times for manual and automated vehicles," Ergonomics, vol. 50, no. 1, pp. 46–58, 2007.

[361] C. Gold, D. Damböck, L. Lorenz, and K. Bengler, "'Take over!' How long does it take to get the driver back into the loop?," in Proc. Hum. Factors Ergonom. Soc. Annu. Meeting, vol. 57, no. 1, pp. 1938–1942, SAGE Publications, 2013.

[362] C. D. Wickens, "Effort in human factors performance and decision making," Hum. Factors, vol. 56, pp. 1329–1336, 2014.

[363] D. A. Norman, The Design of Everyday Things: Revised and Expanded Edition. Basic Books, 2013.

[364] M. R. Endsley, "Autonomous driving systems: A preliminary naturalistic study of the Tesla Model S," J. Cogn. Eng. Decis. Making, vol. 11, no. 3, pp. 225–238, 2017.

[365] R. Parasuraman and D. H. Manzey, "Complacency and bias in human use of automation: Attentional integration," Hum. Factors, vol. 52, no. 3, pp. 381–410, 2010.

[366] M. S. Young and N. A. Stanton, "Back to the future: Brake reaction times for manual and automated vehicles," Ergonomics, vol. 50, no. 1, pp. 46–58, 2007.

[367] M. R. Endsley, "Toward a theory of situation awareness in dynamic systems," Hum. Factors, vol. 37, no. 1, pp. 32–64, 1995.

[368] N. B. Sarter and D. D. Woods, "How in the world did we ever get into that mode? Mode error and awareness in supervisory control," Hum. Factors, vol. 37, no. 1, pp. 5–19, 1995.

[369] M. Kyriakidis, J. de Winter, N. Stanton, T. Bellet, B. van Arem, K. Brookhuis, M. Martens, K. Bengler, J. Andersson, N. Merat, and N. Reed, "A human factors perspective on automated driving," Theor. Issues Ergon. Sci., vol. 20, no. 3, pp. 223–249, 2019.

[370] N. Merat and T. Louw, "Allocation of function to humans and the transfer of control," in Handbook of Human Factors for Automated, Connected, and Intelligent Vehicles, D. L. Fisher, W. J. Horrey, J. D. Lee, and M. A. Regan, Eds. Florida, USA: CRC Press, 2020, pp. 153–171.

[371] M. Saffarian, C. Happee, and J. C. F. De Winter, "Why do drivers have accidents in highly automated driving?," in Proc. Hum. Factors Ergonom. Soc. Annu. Meeting, vol. 56, no. 1, pp. 2296–2300, 2012.

[372] R. Parasuraman and V. Riley, "Humans and automation: Use, misuse, disuse, abuse," Hum. Factors, vol. 39, no. 2, pp. 230–253, 1997.

[373] J. D. Lee and K. A. See, "Trust in automation: Designing for appropriate reliance," Hum. Factors, vol. 46, no. 1, pp. 50–80, 2004.

[374] B. M. Muir, "Trust in automation: Part I. Theoretical issues in the study of trust and human intervention in automated systems," Ergonomics, vol. 37, no. 11, pp. 1905–1922, 1994.

[375] K. A. Hoff and M. Bashir, "Trust in automation: Integrating empirical evidence on factors that influence trust," Hum. Factors, vol. 57, no. 3, pp. 407–434, 2015.

[376] Russell and Grove, "Human factors considerations in preparing policy and regulation for automated vehicles," in Handbook of Human Factors for Automated, Connected, and Intelligent Vehicles, D. L. Fisher, W. J. Horrey, J. D. Lee, and M. A. Regan, Eds. Florida, USA: CRC Press, 2020, pp. 319–336.

[377] P. A. Hancock et al., "A meta-analysis of factors affecting trust in human-robot interaction," Hum. Factors, vol. 53, no. 5, pp. 517–527, 2011.

[378] M. Burke, "Human factors considerations in preparing policy and regulation for automated vehicles," in Handbook of Human Factors for Automated, Connected, and Intelligent Vehicles, Florida, USA: CRC Press, 2020, pp. 319–336.

[379] J. L. Campbell, L. Hoekstra-Atwood, J. Lee, and C. Richard, "HMI design for automated, connected, and intelligent vehicles," in Handbook of Human Factors for Automated, Connected, and Intelligent Vehicles, USA: CRC Press, 2020, pp. 337–358.

[380] M. A. Regan et al., "Education and training for drivers of assisted and automated vehicles," Austroads Res. Rep. AP-R616-20, Austroads, Sydney, Australia, 2020.

[381] J. Gaspar, "Human-machine interface design for fitness-impaired populations," in Handbook of Human Factors for Automated, Connected, and Intelligent Vehicles, D. L. Fisher, W. J. Horrey, J. D. Lee, and M. A. Regan, Eds. Florida, USA: CRC Press, 2020, pp. 359–376.

[382] R. A. Grier, "Automated vehicle design for people with disabilities," in Handbook of Human Factors for Automated, Connected, and Intelligent Vehicles, D. L. Fisher, W. J. Horrey, J. D. Lee, and M. A. Regan, Eds. Florida, USA: CRC Press, 2020, pp. 377–394.

[383] T. M. Gasser and D. Westhoff, BASt-study: Definitions of Automation and Legal Issues in Germany. Bundesanstalt für Straßenwesen, 2012.

[384] P. Lin, "Why ethics matters for autonomous cars," in Autonomes Fahren, M. Maurer, J. C. Gerdes, B. Lenz, and H. Winner, Eds. Springer Vieweg, 2016, pp. 69–85.

[385] I. Y. Noy, M. L. Shinar, and M. Horrey, "Automated driving: Safety blind spots," Safety Sci., vol. 102, pp. 68–78, 2018.

[386] E. Awad et al., "The Moral Machine Experiment," Nature, vol. 563, pp. 59–64, 2018.

[387] A. Chella et al., "Competent Moral Reasoning in Robot Applications: Inner Dialog as a Step Towards Artificial Phronesis," in Trolley Crash: Approaching Key Metrics for Ethical AI Practitioners, Researchers, and Policy Makers, P. Wu, M. Salpukas, X. Wu, and S. Ellsworth, Eds. London, United Kingdom; San Diego, CA, United States: Academic Press, an imprint of Elsevier, 2024.

[388] M. Cunneen et al., "Autonomous Vehicles and Avoiding the Trolley (Dilemma): Vehicle Perception, Classification, and the Challenges of Framing Decision Ethics," Cybernetics and Systems, vol. 51, no. 1, pp. 59–80, 2020.

[389] J. de Freitas et al., "Doubting Driverless Dilemmas," Perspectives on Psychological Science, vol. 15, no. 5, pp. 1284–1288, 2020.

[390] R. de Jong, "The Retribution-Gap and Responsibility-Loci Related to Robots and Automated Technologies: A Reply to Nyholm," Science and Engineering Ethics, vol. 26, pp. 727–735, 2020.

[391] H. Etienne, "When AI Ethics Goes Astray: A Case Study of Autonomous Vehicles," Social Science Computer Review, vol. 40, no. 1, pp. 236–246, 2022.

[392] K. Evans et al., "Ethical Decision Making in Autonomous Vehicles: The AV Ethics Project," Science and Engineering Ethics, vol. 26, pp. 3285–3312, 2020.

[393] P. Foot, "The Problem of Abortion and the Doctrine of the Double Effect," Oxford Review, vol. 5, pp. 5–15, 1967.

[394] L. G. Galvao et al., "Pedestrian and Vehicle Detection in Autonomous Vehicle Perception Systems—A Review," Sensors, vol. 21, 2021.

[395] T. Gill, "Ethical dilemmas are really important to potential adopters of autonomous vehicles," Ethics and Information Technology, vol. 23, pp. 657–673, 2021.

[396] R. Graubohm et al., "Value Sensitive Design in the Development of Driverless Vehicles: A Case Study on an Autonomous Family Vehicle," in International Design Conference - Design 2020, 2020.

[397] S. O. Hansson et al., "Self-Driving Vehicles—an Ethical Overview," Philosophy & Technology, vol. 34, pp. 1383–1408, 2021.

[398] G. Keeling, "Automated Vehicles and the Ethics of Classification," in Autonomous Vehicle Ethics: The Trolley Problem and Beyond, R. Jenkins, D. Černý, and T. Hribek, Eds. New York: Oxford Academic, 2022.

[399] P. Königs, "Of trolleys and self-driving cars: What machine ethicists can and cannot learn from trolleyology," Utilitas, vol. 35, pp. 70–87, 2023.

[400] A. Martinho et al., "Ethical issues in focus by the autonomous vehicles industry," Transport Reviews, vol. 41, no. 5, pp. 556–577, 2021.

[401] Mando, Mando Sustainability Report, 2018. [Online]. Available: https://www.responsibilityreports.com/Company/mando

[402] Mobileye, "Implementing the RSS model on NHTSA pre-crash scenarios," 2017. [Online]. Available: https://static.mobileye.com/website/corporate/rss/rss_on_nhtsa.pdf

[403] G. Mordue et al., "The looming challenges of regulating high-level autonomous vehicles," Transportation Research Part A: Policy and Practice, vol. 132, pp. 174–187, 2020.

[404] Nuro, Delivering Safety: Nuro's Approach, 2018. [Online]. Available: https://lindseyresearch.com/wp-content/uploads/2019/05/NHTSA-2019-0017-0023-delivering_safety_nuros_approach.pdf

[405] J. Robinson et al., "Ethical considerations and moral implications of autonomous vehicles and unavoidable collisions," Theoretical Issues in Ergonomics Science, vol. 23, no. 4, pp. 435–452, 2022.

[406] Toyota, Toyota Automated Driving, 2020. [Online]. Available: https://amrd.toyota.com/app/uploads/2022/02/ATwhitepaper.pdf

[407] S. Umbrello and R. Yampolskoy, "Designing AI for Explainability and Verifiability: A Value Sensitive Design Approach to Avoid Artificial Stupidity in Autonomous Vehicles," International Journal of Social Robotics, vol. 14, pp. 313–322, 2021.

[408] H. Wang et al., "Ethical Decision Making in Autonomous Vehicles: Challenges and Research Progress," IEEE Intelligent Transportation Systems Magazine, vol. 14, no. 1, pp. 6–17, 2022.

[409] S. S. Wu, "Autonomous vehicles, trolley problems, and the law," Ethics and Information Technology, vol. 22, pp. 1–13, 2020.

[410] Q. Yuan et al., "Decision-Making and Planning Methods for Autonomous Vehicles Based on Multistate Estimations and Game Theory," Advanced Intelligent Systems, vol. 5, no. 11, 2023.

[411] H. Nellen, "Car industry told to dial back use of touchscreens," Times, 3 March 2024. [Online]. Available: https://www.thetimes.com/uk/article/stop-making-dangerous-touchscreens-car-firms-told-xv3gmpdc6?region=global

[412] M. L. Cappuccio, J. C. Galliott, F. Eyssel, et al., "Autonomous Systems and Technology Resistance: New Tools for Monitoring Acceptance, Trust, and Tolerance," Int. J. Soc. Robot., vol. 16, pp. 1–25, 2024. doi: 10.1007/s12369-023-01065-2.

[413] M. L. Cappuccio, J. C. Galliott, and E. B. Sandoval, "Saving Private Robot: Risks and Advantages of Anthropomorphism in Agent-Soldier Teams," Int. J. Soc. Robot., vol. 14, no. 10, pp. 2135–2148, 2022. doi: 10.1007/s12369-021-00755-z.

[414] Y. Wang and N. Lu, "Survey on Security in Vehicular Ad Hoc Networks," IEEE Trans. Intell. Transp. Syst., vol. 20, no. 8, pp. 2892–2906, 2019. doi: 10.1109/TITS.2019.2900399.

[415] J. Petit and S. E. Shladover, "Potential Cyberattacks on Automated Vehicles," IEEE Trans. Intell. Transp. Syst., vol. 16, no. 2, pp. 546–556, 2015.

[416] M. Wolf, A. Weimerskirch, and C. Paar, "Security in Automotive Bus Systems: Cryptographic Solutions for CAN and FlexRay," in Embedded Security in Cars, 2nd ed., M. Wolf, Ed. Springer, 2020, pp. 83–101. doi: 10.1007/978-3-662-58434-6.

[417] R. Zhang, X. Wang, Q. Xu, and W. Zhuang, "Automotive Cybersecurity: Fundamentals, Current Challenges, and Future Directions," IEEE Commun. Surv. Tutor., vol. 23, no. 4, pp. 2465–2491, 2021. doi: 10.1109/COMST.2021.3108819.

[418] S. Checkoway et al., "Comprehensive Experimental Analyses of Automotive Attack Surfaces," in Proc. 20th USENIX Security Symp., 2011, pp. 77–92. [Online]. Available: https://www.usenix.org/conference/usenixsecurity11/technical-sessions/presentation/checkoway.

[419] C. Miller and C. Valasek, "Remote Exploitation of an Unaltered Passenger Vehicle," in Proc. Black Hat USA 2015 Conf., 2015.

[420] A. Kurakin, I. J. Goodfellow, and S. Bengio, "Adversarial Examples in the Physical World," in Artificial Intelligence Safety and Security, A. A. Suresh, Ed. Chapman & Hall/CRC, 2018, pp. 99–112. doi: 10.1201/9781351251389-6.

[421] N. Lu, N. Cheng, N. Zhang, X. Shen, and J. W. Mark, "Connected Vehicles in 5G and Beyond: Solutions and Challenges," IEEE Veh. Technol. Mag., vol. 14, no. 1, pp. 14–22, 2019. doi: 10.1109/MVT.2018.2879335.

[422] Z. El-Rewini, K. Sadatsharan, and B. Kantarci, "Cybersecurity Challenges in Vehicular Networks: A Survey," IEEE Commun. Surv. Tutor., vol. 22, no. 2, pp. 1271–1302, 2020. doi: 10.1109/COMST.2020.2982452.

[423] G. Bloom, J. A. Clark, and Á. MacDermott, "Security and Privacy Threats in Vehicular Ad Hoc Networks: A Comprehensive Survey," IEEE Access, vol. 9, pp. 121479–121505, 2021. doi: 10.1109/ACCESS.2021.3109422.

[424] E. Heard, J. Xie, S. Maharjan, and Y. Zhang, "Cybersecurity in Intelligent Transportation Systems: Threats, Challenges, and Solutions," IEEE Netw., vol. 35, no. 2, pp. 84–90, 2021. doi: 10.1109/MNET.011.2000522.

[425] O. Kaiwartya, Y. Cao, X. Li, and O. Carsten, "Internet of Vehicles for Smart and Safe Driving: A Comprehensive Review," IEEE Trans. Intell. Transp. Syst., vol. 21, no. 7, pp. 4613–4624, 2021. doi: 10.1109/TITS.2020.3044868.

[426] S. Parkinson, P. Ward, K. Wilson, and J. Miller, "Cyber Threats Facing Autonomous and Connected Vehicles: Future Challenges," IEEE Trans. Intell. Transp. Syst., vol. 21, no. 11, pp. 3417–3430, 2020. doi: 10.1109/TITS.2019.2926677.

[427] P. Dhal, S. N. Mohanty, and M. K. Khan, "Lightweight Cryptographic Solutions for the Internet of Vehicles (IoV)," IEEE Trans. Veh. Technol., vol. 69, no. 12, pp. 15909–15917, 2020. doi: 10.1109/TVT.2020.3032062.

[428] S. Feng, F. Yan, F. Sun, and Y. Liu, "Intrusion Detection in the Controller Area Network (CAN) Using Deep Neural Networks," IEEE Trans. Intell. Transp. Syst., vol. 22, no. 7, pp. 4667–4676, 2021. doi: 10.1109/TITS.2020.3025252.

[429] G. Rathee, K. Salah, R. Jayaraman, I. Yaqoob, and M. Omar, "A Blockchain Framework for Securing Connected and Autonomous Vehicles (CAVs): Architecture, Challenges, and Solutions," IEEE Access, vol. 9, pp. 111264–111288, 2021. doi: 10.1109/ACCESS.2021.3102500.

[430] C. Liu, C. Gu, and M. Guizani, "Formal Verification for Automotive Software: A Survey on Methods and Tools," ACM Comput. Surv., vol. 54, no. 8, pp. 1–30, 2022. doi: 10.1145/3469881.

[431] A. Weimerskirch and M. Wolf, "Automotive Cybersecurity: Review and Outlook on ISO/SAE 21434," SAE Int. J. Transp. Cybersecurity Privacy, vol. 4, no. 2, pp. 75–83, 2021. doi: 10.4271/12-04-02-0012.

[432] R. van der Heijden and E. Haber, "Policy and Regulation for Cyber-Physical Vehicle Systems: State of the Art and Future Directions," Transp. Res. Part A Policy Pract., vol. 136, pp. 252–266, 2020. doi: 10.1016/j.tra.2020.03.008.

[433] J. Joy, A. George, and M. Anjana, "Privacy Preservation for User Data in Intelligent Connected Vehicles: A Survey of Emerging Trends," IEEE Internet Things J., vol. 9, no. 16, pp. 14234–14250, 2022. doi: 10.1109/JIOT.2022.3145382.

[434] C. Nowakowski, S. E. Shladover, and H.-S. Tan, "Heavy vehicle automation: human factors lessons learned," Procedia Manuf., vol. 3, pp. 2945–2952, Jan. 2015. doi: 10.1016/j.promfg.2015.07.824.

[435] Union of Concerned Scientists, "Maximizing the benefits of self-driving vehicles," Policy Paper, Feb. 2017. [Online]. Available: https://www.jstor.org/stable/resrep17299.

[436] D. Lee and D. J. Hess, "Regulations for on-road testing of connected and automated vehicles: Assessing the potential for global safety harmonization," Transp. Res. Part A Policy Pract., vol. 136, pp. 85–98, Apr. 2020, doi: 10.1016/j.tra.2020.03.026.

[437] D. J. Fagnant and K. Kockelman, "Preparing a nation for autonomous vehicles: opportunities, barriers and policy recommendations," Transp. Res. Part A Policy Pract., vol. 77, pp. 167–181, May 2015, doi: 10.1016/j.tra.2015.04.003.

[438] E. Fraedrich, D. Heinrichs, F. J. Bahamonde-Birke, and R. Cyganski, "Autonomous driving, the built environment and policy implications," Transp. Res. Part A Policy Pract., vol. 122, pp. 162–172, Mar. 2018, doi: 10.1016/j.tra.2018.02.018.

[439] S. E. Shladover and C. Nowakowski, "Regulatory challenges for road vehicle automation: Lessons from the California experience," Transp. Res. Part A Policy Pract., vol. 122, pp. 125–133, Oct. 2017, doi: 10.1016/j.tra.2017.10.006.

[440] T. Sever and G. Contissa, "Automated driving regulations – where are we now?," Transp. Res. Interdiscip. Perspect., vol. 24, p. 101033, Feb. 2024, doi: 10.1016/j.trip.2024.101033.

[441] H. Surden and M. A. Williams, "Technological opacity, predictability, and self-driving cars," Cardozo Law Rev., vol. 38, p. 121, 2016.

[442] Public Policy Forum, "Artificial intelligence: Connected and automated vehicles," Report, 2018. [Online]. Available: https://ppforum.ca/wp-content/uploads/2018/12/ConnectedandAutomatedVehicles-PPF-NOV2018-EN.pdf.

[443] E. R. Straub and K. E. Schaefer, "It takes two to Tango: Automated vehicles and human beings do the dance of driving – Four social considerations for policy," Transp. Res. Part A Policy Pract., vol. 122, pp. 173–183, Mar. 2018, doi: 10.1016/j.tra.2018.03.005.

[444] V. Marchau, J. Zmud, and N. Kalra, "Editorial for the special issue – Autonomous vehicle policy," Transp. Res. Part A Policy Pract., vol. 122, pp. 120–124, May 2018, doi: 10.1016/j.tra.2018.04.017.

[445] A. Faisal, T. Yigitcanlar, Md. Kamruzzaman, and G. Currie, "Understanding autonomous vehicles: A systematic literature review on capability, impact, planning and policy," J. Transp. Land Use, vol. 12, no. 1, Jan. 2019, doi: 10.5198/jtlu.2019.1405.

[446] Z. Altunyaldiz, "Legal aspects of autonomous vehicles," Council of Europe, Committee on Legal Affairs and Human Rights, pp. 1–14, 2020. [Online]. Available: https://pace.coe.int/en/files/28721/html.

[447] N. Chase and U.S. Energy Information Administration, "Autonomous vehicles: uncertainties and energy implications," Conf. Proc., Jun. 2018. [Online]. Available: https://www.eia.gov/conference/2018/pdf/presentations/nicholas_chase.pdf.

[448] P. Koopman, "Testimony of Dr. Philip Koopman," [Online]. Available: https://d1dth6e84htgma.cloudfront.net/IDC_Philip_Koopman_Ph_D_Testimony_Self_Driving_Cars_AV_Hearing_2023_07_26_c2ebaa103f.pdf.

[449] E. R. Straub and K. E. Schaefer, "It takes two to Tango: Automated vehicles and human beings do the dance of driving – Four social considerations for policy," Transp. Res. Part A Policy Pract., vol. 122, pp. 173–183, Mar. 2018, doi: 10.1016/j.tra.2018.03.005.

[450] J. M. Anderson et al., "Autonomous vehicle technology: A guide for policymakers," RAND, Mar. 22, 2016. [Online]. Available: https://www.rand.org/pubs/research_reports/RR443-2.html.

[451] A. Taeihagh and H. S. M. Lim, "Governing autonomous vehicles: emerging responses for safety, liability, privacy, cybersecurity, and industry risks," Transp. Rev., vol. 39, no. 1, pp. 103–128, Jul. 2018, doi: 10.1080/01441647.2018.1494640.

[452] N. E. Vellinga, "From the testing to the deployment of self-driving cars: Legal challenges to policymakers on the road ahead," Comput. Law Security Rev., vol. 33, no. 6, pp. 847–863, Jun. 2017, doi: 10.1016/j.clsr.2017.05.006.

[453] "Global Guide to Autonomous Vehicles 2024," [Online]. Available: https://www.dentons.com/en/insights/guides-reports-and-whitepapers/2024/may/29/global-guide-to-autonomous-vehicles-2024.

[454] L. Hansson, "Regulatory governance in emerging technologies: The case of autonomous vehicles in Sweden and Norway," Res. Transp. Econ., vol. 83, p. 100967, Sep. 2020, doi: 10.1016/j.retrec.2020.100967.

[455] M. Wansley, "Regulating driving automation safety," LARC @ Cardozo Law, [Online]. Available: https://larc.cardozo.yu.edu/faculty-articles/807.

[456] World Economic Forum, "Autonomous Vehicle Policy Framework: Selected National and Jurisdictional Policy Efforts to Guide Safe AV Development," Insight Rep., Nov. 2020. [Online]. Available: www3.weforum.org/docs/WEF_C4IR_Israel_Autonomous_Vehicle_Policy_Framework_2020.pdf.

[457] S. Gless, E. Silverman, and T. Weigend, "If robots cause harm, who is to blame? Self-driving cars and criminal liability," New Crim. Law Rev., vol. 19, no. 3, pp. 412–436, Jan. 2016, doi: 10.1525/nclr.2016.19.3.412.

[458] G. Contissa, F. Lagioia, and G. Sartor, "The ethical knob: Ethically-customisable automated vehicles and the law," Artif. Intell. Law, vol. 25, no. 3, pp. 365–378, Sep. 2017, doi: 10.1007/s10506-017-9211-z.

[459] L. Collingwood, "Privacy implications and liability issues of autonomous vehicles," Inf. Commun. Technol. Law, vol. 26, no. 1, pp. 32–45, Jan. 2017, doi: 10.1080/13600834.2017.1269871.

[460] D. Lee and D. J. Hess, "Public concerns and connected and automated vehicles: Safety, privacy, and data security," Humanit. Soc. Sci. Commun., vol. 9, no. 1, Mar. 2022, doi: 10.1057/s41599-022-01110-x.

[461] T. de J. Mateo Sanguino, J. M. Lozano Dominguez, and P. de Carvalho Baptista, "Cybersecurity certification and auditing of the automotive industry," in Adv. Transp. Policy Plan.: Policy Implications of Autonomous Vehicles, D. Milakis, N. Thomopoulos, B. van Wee, Eds. London, U.K.: Elsevier, 2020, pp. 97–124.

[462] "Automated Vehicles Comprehensive Plan," U.S. Dep. Transp., [Online]. Available: https://www.transportation.gov/av/avcp.

[463] U.S. Dep. Transp., "Second amended Standing General Order 2021-01 Incident Reporting for Automated Driving Systems (ADS) and Level 2 Advanced Driver Assistance Systems (ADAS)," May 2023. [Online]. Available: https://www.nhtsa.gov/sites/nhtsa.gov/files/2023-04/Second-Amended-SGO-2021-01_2023-04-05_2.pdf.

[464] "Standing General Order on Crash Reporting | NHTSA," NHTSA, [Online]. Available: https://www.nhtsa.gov/laws-regulations/standing-general-order-crash-reporting.

[465] "中华人民共和国工业和信息化部," [Online]. Available: http://www.miit.gov.cn/.

[466] "深圳政府在线_深圳市人民政府门户网站," [Online]. Available: http://www.sz.gov.cn/.

[467] "EUR-LEX - 52021PC0206 - EN - EUR-LEX," [Online]. Available: https://eur-lex.europa.eu/legal-content/EN/TXT/?uri=CELEX:52021PC0206.

[468] "Implementing regulation - 2022/1426 - EN - EUR-Lex," [Online]. Available: https://eur-lex.europa.eu/eli/reg_impl/2022/1426/oj.

[469] "Automated Vehicles - Law Commission," Law Commission, Sep. 03, 2024. [Online]. Available: https://www.lawcom.gov.uk/project/automated-vehicles/.

[470] Dep. Transp., "Future of transport regulatory review: Modernising vehicle standards," GOV.UK, Nov. 12, 2021. [Online]. Available: https://www.gov.uk/government/consultations/future-of-transport-regulatory-review-modernising-vehicle-standards.

[471] "国土交通省," [Online]. Available: https://www.mlit.go.jp/.

[472] "日本自動車研究所｜日本自動車研究所," [Online]. Available: https://www.jari.or.jp/.

[473] "Home | NITI Aayog," Jan. 08, 2025. [Online]. Available: https://www.niti.gov.in/.

[474] "Ministry of Road Transport & Highways, Government of India," [Online]. Available: https://morth.nic.in/.

[475] "The regulatory framework for automated vehicles in Australia," Natl. Transp. Comm., Policy Paper, 2021. [Online]. Available: https://www.ntc.gov.au.

[476] Natl. Transp. Comm., "Regulating government access to C-ITS and automated vehicle data," 2019. [Online]. Available: https://www.ntc.gov.au/sites/default/files/assets/files/NTC-Policy-Paper-Regulating-government-access-to-C-ITS-and-AV-data.pdf.

[477] J. Fleetwood, "Public health, ethics, and autonomous vehicles," Am. J. Public Health, vol. 107, no. 4, pp. 532–537, Apr. 2017.

[478] Z. Assaad, "Driverless taxi chaos in San Francisco erodes public trust in autonomous technology," ABC News, Oct. 14, 2023. [Online]. Available: https://www.abc.net.au/news/science/2023-10-14/san-francisco-driverless-taxis-autonomous-cars-trial-trust/102914982

[479] Y. Lu, "Lost Time for No Reason': How Driverless Taxis Are Stressing Cities," The New York Times, Nov. 20, 2023. [Online]. Available: https://www.nytimes.com/2023/11/20/technology/driverless-taxis-cars-cities.html

[480] S. E. Shladover, Connected and automated vehicle systems: Introduction and overview. Journal of Intelligent Transportation Systems, 22(3), 190–200, https://doi.org/10.1080/15472450.2017.1336053

[481] https://group.mercedes-benz.com/innovations/product-innovation/autonomous-driving/drive-pilot-95-kmh.html

[482] www.motor1.com/news/563945/level3-autonomous-tech-real-world/

[483] https://www.nhtsa.gov/laws-regulations/standing-general-order-crash-reporting/ads

[484] https://www.nature.com/articles/s41467-024-48526-4

[485] https://ascelibrary.org/doi/abs/10.1061/9780784484876.005

[486] https://www.tandfonline.com/doi/full/10.1080/19439962.2023.2284175

[487] https://www.aimsun.com/

[488] https://www.cs.utexas.edu/~aim/

[489] https://eclipse.dev/sumo/

[490] https://carla.org/